\documentclass{article}
\usepackage{frExamplee}
\usepackage{graphicx}

\usepackage{natbib}
\let\bibhang\relax
\usepackage{apalike}

\usepackage{setspace}
\usepackage{appendix}

\usepackage{xcolor}
\usepackage[english]{babel}
\usepackage{amsmath}
\usepackage{amsfonts}

\usepackage{csquotes}% Recommended
\usepackage[parfill]{parskip}
\usepackage{color}
\usepackage{placeins}
\usepackage{subcaption}
\usepackage{flushend}
\usepackage{algorithm}
\usepackage{algpseudocode}
\usepackage{multirow}
\usepackage{graphicx}
\usepackage{float}
\usepackage{colortbl}
\usepackage{bm}

\usepackage{todonotes}

\DeclareMathOperator*{\argmax}{arg\,max}
\DeclareMathOperator*{\argmin}{arg\,min}

\title{Feature Space Exploration For Planning Initial Benthic AUV Surveys}

\author{
Jackson Shields \\
Australian Centre for Field Robotics\\
University of Sydney \\
NSW, Australia 2006 \\
\texttt{j.shields@acfr.usyd.edu.au} \\
\And
Oscar Pizarro \\
Australian Centre for Field Robotics \\
University of Sydney \\
\texttt{o.pizarro@acfr.usyd.edu.au} \\
\AND
Stefan B. Williams \\
Australian Centre for Field Robotics \\
University of Sydney \\
\texttt{stefan.williams@sydney.edu.au}
}

\begin{document}

\maketitle

\begin{abstract}

Special-purpose Autonomous Underwater Vehicles (AUVs) are utilised for benthic (seafloor) surveys, where the vehicle collects optical imagery of the seafloor. Due to the small-sensor footprint of the cameras and the vast areas to be surveyed, these AUVs can not feasibly collect full coverage imagery of areas larger than a few tens of thousands of square meters. Therefore it is necessary for AUV paths to sample the surveys areas sparsely, yet effectively. Broad-scale acoustic bathymetric data is readily available over large areas, and is often a useful prior of seafloor cover. As such, prior bathymetry can be used to guide AUV data collection. This research proposes methods for planning initial AUV surveys that efficiently explore a feature space representation of the bathymetry, in order to sample from a diverse set of bathymetric terrain. This will enable the AUV to visit areas that likely contain unique habitats and are representative of the entire survey site. We propose several information gathering planners that utilise a feature space exploration reward, to plan freeform paths or to optimise the placement of a survey template. The suitability of these methods to plan AUV surveys is evaluated based on the coverage of the feature space and also the ability to visit all classes of benthic habitat on the initial dive. Informative planners based on Rapidly-expanding Random Trees (RRT) and Monte-Carlo Tree Search (MCTS) were found to be the most effective. This is a valuable tool for AUV surveys as it increases the utility of initial dives. It also delivers a comprehensive training set to learn a relationship between acoustic bathymetry and visually-derived seafloor classifications.

% \todo{Mention MCTS, RRT, Templates; Mention benchmark methods; Mention field trials}

\end{abstract}

%% ----- INTRODUCTION -----

\section{Introduction}\label{sec:intro}

Engaging Autonomous Underwater Vehicles (AUVs) for marine scientific surveys has several benefits: their ability to collect data in areas inaccessible to humans; the greater endurance; and their enhanced data capture ability. Benthic AUV surveys observe and monitor the seafloor habitat by using AUVs to collect imagery along preplanned paths. Generally the areas to be characterised are vast (measured in millions to billions of square meters), and the image sensor footprint of the AUV is small (~1-10 $m^2$), therefore full-coverage can not feasibly be collected at these scales. In order to collect data which is representative of the entire survey site, while satisfying the constraints of the vehicle capabilities, consideration must be given to the planned survey path. Ideally, these benthic surveys should sample from all the habitat types present in the survey area. 

Remotely-sensed data, such as bathymetry, backscatter, aerial and satellite imagery can be collected over broad areas at a much larger scale than is feasible with AUVs, yet at a relatively low resolution. This data can provide useful insights into the habitats likely to be found \citep{Brown2011}. Habitat modelling relates the low-resolution, full-coverage, remotely-sensed data to the habitat labels, which are derived from high-resolution AUV imagery in order to infer the benthic habitat beyond where the AUV has sampled. The degree to which the remotely-sensed data can predict the benthic habitat is variable and depends on the data modality and habitats of interest. Benthic habitat mapping often utilises acoustically derived remotely-sensed data such as bathymetry and backscatter, where the structure and acoustic reflectance of the seafloor is indicative of the benthic habitat.  In shallow water, LiDAR can be used to obtain the bathymetry. Alternatively for shallow water, hyperspectral imagery can also be used to determine the benthic habitat \citep{Thompson2017}, where the spectral response of the seafloor can be used to classify the benthic habitat with higher accuracy and spatial resolution than acoustic methods.  In this research, we focus on building habitat models using acoustic data to identify broad habitat labels such as sand, reef, kelp, coral. The habitat models can be used to direct the AUV to sample specific habitats, as well as areas that are likely to most improve the model \citep{Bender2013a, Rao2017a, Shields2020}. For these models to be created, they need a balanced and representative seed dataset consisting of remotely-sensed data and in-situ, geo-referenced imagery. This motivates the need to capture all the habitat types and thoroughly explore the feature space within the initial survey.

Currently, AUV surveys are mostly planned manually, with the survey planner identifying areas of interest while inspecting the remotely-sensed data. While expert knowledge can provide valuable insights when designing a survey, automatically generating a survey can provide a reliable and principled method for survey design \citep{Foster2020}. This paper explores methods for autonomously planning a representative and comprehensive initial AUV survey by utilising prior bathymetry. Features are extracted from the bathymetry that represent the structure of the seafloor, which is indicative of the benthic habitat \citep{Brown2011}. These features come from either established geometric processes or encoding bathymetric patches with an autoencoder. Geometric features represent the structure of the seafloor through a combination of depth, slope, aspect and rugosity. Encoded features are formed by compressing small patches of bathymetry with an autoencoder, with the latent space of the autoencoder being a compact representation of the input patch. For this application, these features are extracted at a single scale, determined by the resolution of the bathymetry. Extracting these scales at several resolutions can be used to extend the feature vector and can improve the performance when performing habitat modelling \citep{Wilson2007}. The feature space is the collection of these features and is a representation of the range of bathymetric terrain found in the survey area. AUV paths are subsequently planned to maximise coverage of this feature space to ensure the AUV observes the complete range of bathymetric terrain and benthic habitats found in the survey area.

\begin{figure}[!htbp]  % old col width = 0.57
    \centering
    \includegraphics[width=0.99\columnwidth]{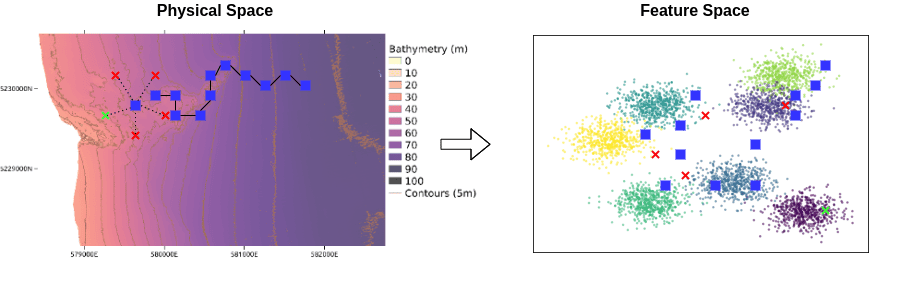}
    \caption{An overview of linked feature space and physical space exploration, where the objective is to comprehensively explore the feature space given a budget in physical space. Features extracted from along the survey path are projected into the feature space. In this demonstration, the blue squares indicate features already on the path, with the planner making a decision about where to move physically (indicated by crosses). The planner chooses the green cross over the red cross as it occupies a more unexplored area of the feature space.}
    \label{fig:intro:concept}
\end{figure}

We introduce the problem of exploring the feature space of remotely-sensed data by planning in physical space. This concept is demonstrated in Figure \ref{fig:intro:concept}. This paper serves as an investigation into various approaches to address this problem. While we focus on the problem of visiting diverse bathymetric terrain with an AUV, this approach is applicable to any application where in-situ sampling is planned from remotely sensed data. Analogues to this problem exist in space exploration \citep{Li2005} and agriculture monitoring \citep{RaminShamshiri2018a}.

In this paper, we propose an information gathering framework that consists of an objective function that rewards collecting features that are different to features already collected, and an evaluation function that scores paths based on the coverage of the feature space. We integrate this framework into information gathering planners based on MCTS, RRT and placing survey templates. This is compared against a more conventional planning approach, that first proposes a set of points that represent the feature space and then plans a spatial path between them using set-TSP (Travelling Salesman Problem). These methods are evaluated based on how much of the feature space they explore given a set distance budget, with the results showing a significant increase in information collected compared to random sampling.

%Two planning methods are proposed to efficiently traverse the feature space. The first method proposes a set of points that represent the feature space and a spatial path between them is solved as a set-TSP (Travelling Salesman Problem) problem. The second method uses a RIG-Tree (Rapidly-exploring Information Gathering) \cite{Hollinger2014} inspired exploratory planner, which is rewarded according to a feature space distance metric to areas which are not yet explored. These methods are evaluated based on how much of the feature space they explore given a set distance budget.

The primary contributions of this paper are:
\begin{itemize}
    \item The formalisation of the problem of exploring a feature space representation of remotely-sensed data for informative path planning of a mobile robot. We present the case of exploring the bathymetric feature space for benthic survey planning. Following this, we propose a novel approach to this problem that links feature-based active learning to robotic information gathering in order to plan AUV surveys that comprehensively explore the feature space. We develop a suite of planners based on RRT, MCTS and setting survey templates that utilise a feature-based reward function to plan informative paths.
	\item Demonstration of this approach on several survey areas, each with different sources of bathymetry, highlighting the versatility and value of this approach.
	\item Investigate planning with different feature representations, to show the methods proposed are suitable for various feature spaces and can be applied to analogue problems with different remotely-sensed data modalities. 
	\item Development of evaluation criteria for initial surveys that do no not rely on visual habitat labels and use these criteria to evaluate each method's ability to plan initial benthic surveys.
	\item Integration of AUV motion and terrain constraints into the planning process to allow for safe operation of the AUV.
\end{itemize}

The remainder of this paper is organised as follows: Section \ref{sec:related_work} provides an overview of the literature in this field; Section \ref{sec:feature_extraction} outlines the methods for bathymetric feature extraction; Section \ref{sec:info} proposes a reward function for exploring the feature space and proposes a set of information gathering planners; Section \ref{sec:results} presents results for each of the proposed planners and Section \ref{sec:field_trials} details the field trials conducted. Finally, Section \ref{sec:conclusion} concludes the paper and identifies avenues of future research.

\section{Related Work}\label{sec:related_work}

% \begin{itemize}
%     \item Robotic Information gathering
%     \item Robotic path planning
%     \begin{itemize}
%         \item Traditional cell-based planning - A* etc.
%         \item Sampling-based robotic planning - PRM, RRT, MCTS
%         \item Reinforcement learning
%     \end{itemize}
%     Robotic information gathering
%     \begin{itemize}
%         \item \citet{Chen2019}
%         \item \citet{Patten2018}
%         \item \citet{Best2016}
%         \item \citet{Kodgule2019}
%         \item Ayrton - Vulcan paper - adding risk constraint
%         \item GP-regression RRT \citet{Viseras2019a}
%         \item Q-learning reinforcement learning \citet{Viseras2019}
%     \end{itemize}
%     Benthic Survey planning
%     \begin{itemize}
%         \item Bender - placement of survey grids
%         \item Foster 2014
%         \item Foster 2020
%     \end{itemize}
%     Active Learning
%     \begin{itemize}
%         \item Feature-based vs Model-based planning
%         \item Need to seed the model
%         \item \citet{Nguyen2004}
%         \item \citet{Fujii1998}
%         \item \citet{Fujii1998}
%         \item \citet{Geifman2017}
%     \end{itemize}
% \end{itemize}

% robot info gathering
Robotic information gathering is path planning where the goal is to maximise an information reward. The need to autonomously explore an environment is multidisciplinary while promoting many unique applications, including tracking biological hotspots with AUVs \citep{Das2013}, space exploration \citep{Arora2019} and active perception and illumination \citep{Sheinin2016}.

% Path planners
Traditionally, most robotic path planners would operate on discrete spaces, with the robot's environment decomposed into a grid. However, these approaches do not scale well to large or highly-dimensional environments, as the search space becomes too large. Sampling-based planners such as Probabalistic Road Maps (PRM) \citep{Kavraki1996} and Rapidly-exploring Random Trees \citep{Karaman2011} are able to operate in large and highly-dimensional environments while generating fast and usable solutions. More recently, Monte-Carlo Tree Search planners have become popular, buoyed by the success of the go-playing algorithm, AlphaGO \citep{Silver2016a}.

% RIG tree + Derivatives
\citet{Hollinger2014} propose three information gathering algorithms; Rapidly-exploring Information Gathering (RIG)-Roadmap, RIG-Tree and RIG-Graph. These algorithms incrementally plan paths to maximise information given a budget constraint. RIG-Tree, which is based on RRT*, was shown to be the most effective. These algorithms were designed to be adaptable for different information goals by changing the reward function. \citet{Jadidi2019} use a RIG-Tree based planner to guide a robot to efficiently map a spatial field using a GP (Gaussian Process), where the reward function is derived from mutual information. They developed stopping criteria to allow planning to finish when the map is deemed to be sufficiently explored. This highlights the adaptability of RIG-Tree based planners to different information objectives.

% \citet{Viseras2019a} - Robotic Active Information Gathering for Spatial Field Reconstruction with Rapidly-Exploring Random Trees and Online Learning of Gaussian Processes. \textcolor{red}{TAKEAWAY}

% MCTS
MCTS-based information gathering planners provide an effective balance between exploration and exploitation, by using the UCT (Upper Confidence Bound 1 applied to Trees)\citep{Kocsis2006} to select promising nodes for expansion. \citet{Chen2019} propose a multi-objective extension of the MCTS search process to optimise over multiple, possibly competing, objectives. This planner is utilised to plan AUV paths that maximise visiting environmental hotspots, such as algal blooms. To improve the operating utility in risk-prone environments, \citet{Ayton2019} extends MCTS by adding a constraint to minimise risk in the formulated plan. These examples show the versatility of MCTS-based informative path planners.

% Mapping spatial fields
A key objective in the adaptive sampling literature is to characterise an environmental phenomenon with minimal samples. \citet{Krause2008} investigated this problem, focusing on approximating a spatial field (such as temperature) with a given number of observations. They model the spatial field using GPs and propose both maximum entropy and mutual information criteria that can be used to optimally place sensors such that the GP uncertainty over the spatial field is minimised. This approach has been integrated into information gathering planners \citep{Singh2009,Hitz2017,Jadidi2019,Candela2021}. There has been research into modelling the bathymetry with GPs to enable efficient navigation or mapping, for example \citep{Wilson2018}. 
However, these examples operate under the assumption that points sampled close together are similar, which is appropriate when modelling spatial fields such as temperature or wind. In this research, we do not make that assumption, but instead rely solely on the remotely-sensed data to inform the informative path planning, as the seafloor structure is a strong indicator of the benthic habitat. An avenue of research could be to model the feature space with GPs, but that can be complex given the high dimensionality of the feature space.

Many of the information gathering algorithms presented here are used to map a spatial field using a minimal trajectory. \citet{Kodgule2019} takes a different direction, by planning a ground robot's in-situ spectroscopy measurements to aid in spectral unmixing of remotely-sensed hyperspectral data. MCTS with a differential entropy reward is used to plan the robot's path, with the results indicating this planner improves the accuracy of the unmixing process for a given budget. Another objective for information gathering planners is active classification, where the planner aims to collect information that will most improve an object classifier. \citet{Candela2021} uses hyperspectral satellite imagery for benthic habitat mapping in shallow water. Their approach leverages a Variational Autoencoder (VAE) to extract features from the hyperspectral data, with these features used as input into a neural network that performs benthic habitat classification. They compare several planning approaches to collect in-situ observations to ground truth the remotely-sensed data, including MCTS, Bayesian Optimisation (BO) and Ergodic Optimal Control. They find that although computationally intensive, MCTS plans the most informative sampling trajectory. \citet{Patten2018} use a MCTS-based planner to plan a ground robot's path that will most improve its classification of objects with an onboard laser-scanner. A GP classifier is used to classify the laser scans and the information reward is predicted mutual information from the simulated path. For an Unmanned Aerial Vehicle (UAV) tasked with collecting imagery that will be used for terrain classification, \citet{Ruckin2021} links active learning with informative path planning. A probabilistic segmentation network is used to classify each pixel of the image, while the uncertainty estimate from this network is used to reward the planner for visiting areas that will most improve the underlying model. This method is effective for improving a model with minimal samples, however it needs a model to be trained on some initial data before this strategy can be used. This highlights the need for an additional method to collect informative data before any samples have been collected. Overall, this collection of research highlights the utility of information gathering planners for finding informative paths for a wide variety of objectives.

% include???
% \textcolor{orange}{
Recent advances have been made around using Reinforcement Learning (RL) to plan informative trajectories. \citet{Viseras2019} use deep RL to guide multiple robots to collect informative samples of a spatial field. The reward function is designed to both maximise information gain and avoid collisions between agents. Results demonstrate improved performance using RL over GP-driven informative sampling. However RL models can be excessively noisy and difficult to debug, as the planner is a black-box model. RL methods can be useful when there is a complex and or abstract relationship between the robot and the environment, however the added complexity of these approaches is unnecessary when a more direct solution is possible.
%} 
%\textcolor{red}{(TAKEAWAY. WHY NOT USE RL.)} 
% Downsides: %https://towardsdatascience.com/why-you-shouldnt-use-reinforcement-learning-163bae193da8
% \begin{itemize}
%     \item excessively noisy
%     \item hard to debug, explain
%     \item more exact methods possible
% \end{itemize}

Active learning involves proposing training samples that will most improve the underlying model. It is primarily focussed on identifying which data points from a large data pool should be labelled \citep{Settles2009}. There are two classes of active learning algorithms; model predictive and feature based (or density based) approaches. Model predictive techniques involve training the model on an initial seed dataset, and evaluating the model's uncertainty when querying unlabelled points. Alternatively, feature based approaches identify a set of samples that most comprehensively represents the distribution of features in the dataset. Model predictive techniques rely on having an initial seed dataset on which to train the initial model, which is often randomly selected from the data pool. For this habitat mapping application, randomly sampling the environment is inefficient due to the difficulty and cost of deploying an AUV and the spatial correlation of benthic habitats, meaning that randomly-placed transects are unlikely to sufficiently represent the survey area. Furthermore, bathymetry can be a useful prior for explaining benthic habitats and should be utilised for planning. Collecting the initial dataset is known as the \lq cold start' problem \citep{Gong2019}. As there is no need to query a trained model, feature based approaches can be used for cold start active learning. \citet{Fujii1998} postulate that queried samples should be both similar to unlaballed data points and dissimilar to collected samples. \citet{Nguyen2004} use clustering of the feature space and querying from each cluster to ensure the collected samples match the distribution of features in the unlabelled dataset. \citet{Geifman2017} propose Farthest-First Active (FF-Active), which uses the intermediate output of a pre-trained neural network as the feature space and the next data point to collect should be farthest in feature space from the collected samples, in order to cover the distribution of features with minimal samples. Similarly, \citet{Sener2017} propose a core-set method, that solving the k-centre problem to comprehensively cover the feature space of the pool data using a minimal number of samples. By using a feature-based active learning approach to select where to sample, the AUV can collect informative data while minimising distance travelled. To classify benthic images captured by an AUV, \citet{Yamada2021} observes that using training samples that are distributed across the feature space results in higher accuracy than class-balancing the training samples, highlighting the importance of comprehensively sampling from across the feature space. A limitation of feature-based active learning is that it distributes features uniformly across the feature space, whereas to most improve the model it can be more effective to focus sampling in specific areas, for example around the decision boundaries. However this application focuses on the \lq cold start' problem, where sampling is being allocated before any model has been trained. For this case, using feature-based active learning to collect samples that cover the feature space is the most effective strategy given the available data.

Using bathymetry as a prior, \citet{Bender2013a} optimises the placement of an AUV survey template in the environment. 
He proposes modelling the distribution of features using a Gaussian Mixture Model (GMM) and then minimises the Kullback–Leibler (KL) divergence between the overall distribution of features in the environment and the distribution of samples in selected survey areas. For this application, aiming to replicate the overall distribution of features could result in suboptimal surveys, as there is often an overabundance of bare habitats. Furthermore, following a set survey template limits the habitat types that can be visited in a single dive.

\citet{Foster2014} analyse the impact of different AUV survey designs on the statistical bias of habitat models, when using these models to predict the percentage cover of species observed through the AUV imagery. From this, a set of guidelines is developed for designing AUV surveys that will likely lead to a representative survey of the area of interest. % todo - why planned surveys naturally follow these guidelines
Following from this, \citet{Foster2020} develop a method for placing AUV surveys, so that they are both spatially balanced and visit specific environmental features. These surveys take into account scientists' specifications to focus on the features of interest.

There is a research opportunity for creating free-form AUV trajectories that will collect a comprehensive seed dataset for a predictive habitat model (for example \citep{Shields2020}). Sampling-based information gathering planners provide an  adaptable and exploratory framework for robotic path planning. By taking a feature-based approach to active learning, the planner can be rewarded for exploring the feature space, to generate, prior to the first deployment, a plan that is expected to collect informative samples.

\section{Feature Extraction}\label{sec:feature_extraction}
Bathymetry can be a strong indicator of the benthic habitat \citep{Friedman2012}. The strength of this relationship is variable, dependent on the area, resolution of the bathymetry, and the habitats present. Bathymetric features based on basic geometric descriptions such as depth, slope, aspect and rugosity are known to be typically correlated to the corresponding benthic habitats \citep{Wilson2007}.

\subsection{Geometric Features}\label{sec:feature_extraction:geometric}

With the bathymetry raster as input, we compose a geometric feature vector composed of the following attributes; depth, slope, aspect and ruggedness. The depth attribute is simply the depth of the seafloor at the given location. The slope and aspect are calculated using Horn's algorithm \citep{Horn1981}, that uses the surrounding 8 raster cells to calculate the gradient, with each cell weighted according to its distance to the centre pixel. The Terrain Ruggedness Index (TRI) \citep{Wilson2007} is used to estimate the ruggedness, which is the mean difference between a cell and its surrounding eight cells.

Using Horn's method, the slope in the x-direction ($p_x$) and y-direction ($p_y$) can be calculated as: % https://desktop.arcgis.com/en/arcmap/10.3/tools/spatial-analyst-toolbox/how-slope-works.htm
\begin{equation*}
    p_x =  \frac{(z_{-+} + 2 z_{-0} + z_{--}) - (z_{++} + 2 z_{+0} + z_{+-})}{8d},
\end{equation*}
\begin{equation*}
    p_y = \frac{(z_{--} + 2z_{-0} + z_{-+}) - (z_{++} + 2z_{+0} + z_{+-})}{8d},
\end{equation*}
where $z_{00}$ corresponds to the centre cell, $z_{-+}$ corresponds to the top left cell and $z_{+-}$ corresponds to the top right cell in a 9-square grid. $d$ is size of each raster cell. Following this, the magnitude of the slope ($s$)  and aspect ($a$) can be calculated as:
\begin{equation}
    s = \sqrt{p_x^2 + p_y^2},
\end{equation}
\begin{equation}
    a = \arctan \frac{p_x}{p_y},
\end{equation}

where the aspect is decomposed into the northing and easting components to avoid the angle discontinuity.

Using this notation, the ruggedness as calculated by TRI ($r_{TRI}$) can be written as:
\begin{equation}
r_{TRI} = \frac{1}{8}\sum_{i \in \{-,0,+\}}\sum_{j \in \{-,0,+\}} z_{00} - z_{ij}
\end{equation}

The final feature vector at a given location is composed as follows:
\begin{equation}
    \mathbf{z}_{\text{geometric}} = \begin{pmatrix} d & s & a_N & a_E & r_{TRI} \end{pmatrix},
\end{equation}
where $d$ is the depth, $s$ is the slope, $a_N$ and $a_E$ are the aspects in the northerly and easterly directions respectively and $r_{TRI}$ is the ruggedness.

\subsection{Learnt Features}\label{sec:feature_extraction:learnt}

The geometric features, while intuitive,  can be too coarse to represent some benthic habitats, not necessarily representing the bathymetry efficiently and comprehensively. Alternatively, neural networks can be used to extract useful features from the bathymetry. 

As there are no habitat labels, these features need to be learnt in an unsupervised manner. An autoencoder is an effective way to conduct unsupervised feature learning \citep{Rumelhart1986}. There are two components in an autoencoder, an encoder and a decoder. The encoder compresses the input into a latent space, while the decoder uses this latent space representation as its input and attempts to reconstruct the original input.  An autoencoder is trained to minimise a reconstruction loss such as the mean-squared error loss function: 
\begin{equation}
L_{MSE} = \frac{1}{n} \sum_{i=1}^n (\mathbf{x}_i - \hat{\mathbf{x}}_i)^2
\end{equation}
Convolutional neural networks are capable of extracting rich features where there are local patterns to the data \citep{Lecun2015}. This makes them effective for bathymetric data. In this paper, the autoencoder is trained on patches randomly sampled from the entire bathymetric area. The bathymetric patches used as input to the encoder are centred around a given coordinate and are typically a square of width 21 pixels. The patches are pre-processed by subtracting the mean-value of the patch, in order to give the input a mean value of zero. The encoder consists of two convolutional layers with 512 filters in each, followed by one fully connected layer of 256 neurons with a latent dimension of 32. The decoder is the reverse of the encoder. The autoencoder network is identical to that used for bathymetric feature extraction in the downstream task of habitat modelling \citep{Shields2020}, where these parameters have been empirically selected to optimise that task.

The final feature vector appends the mean depth value to the encoded features:
\begin{equation}
	\mathbf{z}_{encoded} = \begin{pmatrix}
		\mathbf{z}_{latent}  & \bar{d}
	\end{pmatrix}
\end{equation}

\begin{figure}[!ht]  % old col width = 0.57
    \centering
    \includegraphics[width=0.6\columnwidth]{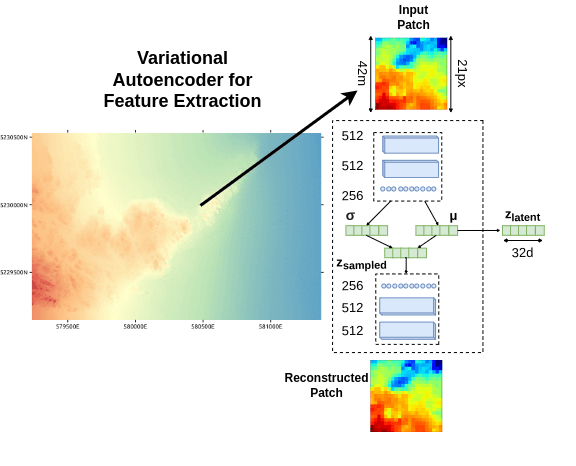}
    \caption{The VAE used for feature extraction. Square ($21\times21$ pixels) patches of bathymetry are extracted from random locations of the bathymetry. These patches are compressed into a representative feature space by the autoencoder, before being reconstructed by the decoder aiming to recreate the sample patch. Using helps generate a continuous feature space, which is useful for exploring.}
    \label{fig:feat:vae_diagram}
\end{figure}

A Variational Autoencoder (VAE) conditions the latent space by imposing a multivariate Gaussian distribution on the latent space. A diagram of the bathymetric VAE used is included in Figure \ref{fig:feat:vae_diagram}.  The conditioning of the latent space leads to a smooth and continuous feature space, enabling easier feature exploration. A VAE is trained using the Evidence Lower Bound (ELBO) loss function, as displayed in the equation below. The left term is equal to the mean squared error, while the right term is the  Kullback-Leibler (KL) divergence between the predicted distribution and a standard normal distribution \citep{Kingma2019}.
\begin{equation}
    L_{ELBO} = \mathbb{E}_{q(z|x)} [\ln p(x|z)] - D_{KL}[q(z|x)||p(z)]
\end{equation}

This paper compares using both geometric and learn features, however the methods explored here are applicable to any feature representation. Section \ref{sec:results} presents results for planning with both geometric features and encoded features.

\section{Informative Path Planning by Exploring the Feature Space}\label{sec:info}

\begin{figure}[!ht]  % old col width = 0.57
    \centering
    \includegraphics[width=0.9\columnwidth]{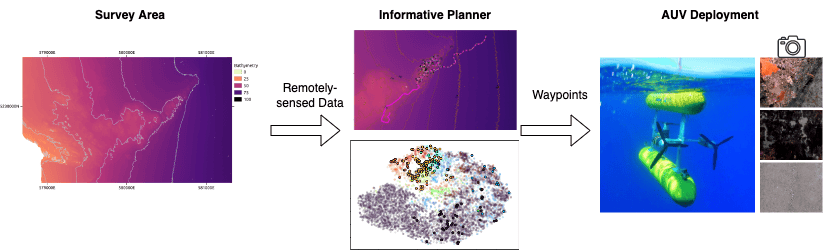}
    \caption{An overview of the process for planning an informative initial survey. The inputs to this process are the survey area and corresponding remotely-sensed data. Using the remotely-sensed data, the planner designs a survey plan that uniformly samples from the feature space. The output of this process is a set of waypoints that the AUV is tasked to follow. Finally, the AUV is deployed and captures benthic seafloor imagery.}
    \label{fig:feat:info_overview}
\end{figure}

This research proposes a framework for informative sampling of a given survey area, which is summarised in Figure \ref{fig:feat:info_overview}. The inputs to this process are the geographical bounds of the area of interest and the corresponding remotely-sensed data (such as bathymetry) covering the area. The proposed informative planner then jointly explores the physical and feature space in order to comprehensively explore the feature space. This planner operates offline, before the AUV is scheduled to be deployed. Having the planner operate offline enables this framework to plan paths for different AUV systems. The output of the planner is a set of waypoints that the AUV is tasked to follow. The AUV is then deployed and captures imagery of the seafloor. The georeferenced imagery can then be used for downstream tasks such as habitat modelling.

\subsection{Problem Definition}
The overarching goal of this research is to explore the feature space, such that the AUV visits the full range of bathymetry terrain which in turn makes it more likely to observe each habitat present in the survey area. \citet{Bender2013a} approaches the initial sampling by planning a \lq representative survey', where the distribution of bathymetric features of the survey should match the distribution of features in the environment. This approach is successful in planning initial benthic surveys, however it can lead to over-sampling of areas that are abundant in the environment. For example, for an area that is primarily composed of flat, relatively featureless terrain with smaller areas of reef, a survey planned to match this distribution would allocate the bulk of its samples to the featureless terrain.

We take a different approach, by proposing that an initial survey should visit the full range of bathymetric features present in the environment regardless of the density of these features. This is equivalent to uniformly sampling across the distribution of features. This is inspired by feature-based active learning, where the objective is to sample from the entire feature space \citep{Sener2017}. A criticism of uniform sampling is that it samples from sparse areas of the feature space, however this is a desirable quality in this application, as it avoids over-sampling of abundant areas (such as the bare, sand habitats common in marine environments) and focuses sampling on areas of diverse terrain that are likely to contain differing habitats.

%As outlined in Section \ref{sec:related_work}, GPs can be used to model a spatial field, which can be combined with a maximum entropy or mutual information criteria to create an effective sampling strategy \citep{Krause2008}. These strategies operate under the assumption that points sampled close together are similar, which is appropriate when modelling spatial fields such as temperature. Here we are not making that assumption, but instead relying solely on the remotely-sensed data to inform our informative path planning. A different avenue of research could be to model the feature space with the GP, but that can be complex given the high dimensionality of the feature space. 

The problem can be defined as selecting a subset of locations to sample ($\mathbf{X}_{path}$) from the survey area ($\mathbf{X}_{area}$), such that the bathymetric features extracted at these locations ($\mathbf{Z}_{path}$) thoroughly explore the overall bathymetric feature space ($\mathbf{Z}_{area}$). Here, $l$ represents a loss function that estimates how well the feature space is explored.
\begin{equation}
	\min_{\mathbf{X}_{path} \subset \mathbf{X}_{area}} \mathbb{E}_{\mathbf{X}_{path}} [ l(\mathbf{Z}_{path} ; \mathbf{Z}_{area}) ]
	\label{eq:problem_def}
\end{equation}

\subsection{Information Reward}\label{sec:infoplan:reward}

For the planner to sample informative areas, the reward function needs to encourage visits to features that are different to already collected features. A reward function based on the Mahalanobis distance between features is used. The Mahalanobis distance \citep{Mahalanobis1936} measures the distance between two points in multivariate space. If the variables are correlated and scaled (as is the case for an autoencoder), the Euclidean distance does not produce meaningful distances, whereas the Mahalanobis distance accounts for possible correlations and scaling differences among the dimensions of the feature vectors. This method is adaptable to any distance measure.

The Mahalanobis distance is given by:
\begin{equation}
D_{m}=\sqrt{(\mathbf{z}_a - \mathbf{z}_b)\text{Cov}[Z]^{-1}(\mathbf{z}_a-\mathbf{z}_b)^T},
\label{eq:info:maha_dist}
\end{equation}
where $\mathbf{z}_a, \mathbf{z}_b \in \mathbb{R}^n$ are the two feature points and $n$ is the number of dimensions in the feature space. $\text{Cov}[Z]^{-1}$ is the inverse covariance matrix of the data. As all the features can be extracted from the environment, the covariance can be estimated for the entire feature space.

Given that the goal is to explore the feature space, the reward function should prioritise features which are distinct from the samples collected so far \citep{Geifman2017}. This is known as Farthest First active learning. Therefore, the reward is calculated as the minimum feature distance to any collected feature:
\begin{equation}
R_{m} = \min_{\forall \mathbf{z}_i \in \mathbf{Z}} D_m(\mathbf{z}_a,\mathbf{z}_i),
\label{eq:info:maha_reward}
\end{equation}
where $R_m$ is the reward, $\mathbf{z}_a$ is the feature currently being evaluated,  $Z$ is the set of features being compared against and  $D_m$ is the Mahalanobis distance.

\subsection{Evaluation of a path }\label{sec:info:eval}

To evaluate the informativeness of a path, the coverage of the feature space should be considered. Using the feature reward (Equation \ref{eq:info:maha_reward}) on the path is not sufficient, as this reward evaluates the diversity of features found on the path, but does not evaluate how this path approximates the survey area. Here we propose multiple metrics to evaluate the informativeness of a path, each with a different perspective.

To assess the diversity of samples on a path, the mean of the paired feature distance between samples is used.
\begin{equation}
    M_{PD} = \frac{1}{N^2}\sum_{\mathbf{z}_i \in \mathbf{Z}_{path}}\sum_{\mathbf{z}_j \in \mathbf{Z}_{path}} D_m (\mathbf{z}_i , \mathbf{z}_j),
    \label{eq:info:eval_pd}
\end{equation}
where $\mathbf{z}_i$ and $\mathbf{z}_j$ are a pair of features collected along the candidate path ($\mathbf{Z}_{path}$), $N$ is the number of features on the path and $D_m$ is the Mahalanobis distance outlined in Equation \ref{eq:info:maha_dist}. Although this metric captures the diversity of samples, it does not capture how well the path characterises the environment. For example, a path that visits a diverse range of terrain but does not visit a large area of similar terrain (e.g. a bare, sandy habitat) will likely score highly using this method, even though it fails to characterise a large area of the environment. 

To ensure the evaluation function is capturing how well the path characterises an environment, the path should minimise the feature distance from a large set of points extracted from the environment and one of the points on the path. As there is a very large number of pixels (on the scale of millions) in the remotely-sensed data, to reduce computational burden, the environment's features approximated by spatially sampling a large set of points and extracting the feature vector at each of these points. The following metric finds the closest feature distance from features randomly extracted from the environment, to one of the features collected on the path.
\begin{equation}
    M_{SD} = \sum_{\mathbf{z}_i \in \mathbf{Z}_{space}} R_{m} (\mathbf{z}_i; \mathbf{Z}_{path}),
    \label{eq:info:eval_sd}
\end{equation}
where $\mathbf{z}_i$ is one of the features randomly extracted from the environment ($\mathbf{Z}_{space})$ and $\mathbf{Z}_{path}$ is the set of features extracted from the path. $R_m$ (Equation \ref{eq:info:maha_reward}) finds the closest feature distance between the given feature ($\mathbf{z}_i$) and the set of reference features ($\mathbf{Z}_{path}$). While this method maximises the area that is characterised by the path, it is biased towards large spatial areas. Using this metric, a path that only visits a large area of similar terrain will score higher than a path that visits a range of terrain but does not explore as much of the similar terrain. Therefore this metric is not used in the evaluation of the surveys.

\begin{figure}[!ht]  % old col width = 0.57
     \centering
     \includegraphics[width=0.6\columnwidth]{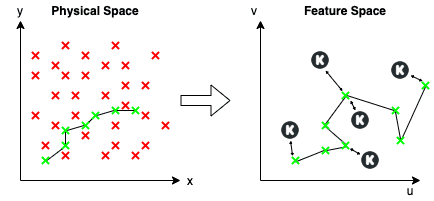}
     \caption{Visualising the process for evaluating a path. On the left it shows the physical (spatial) space and the robot's path. The green crosses are points sampled on the robots path, while the red crosses are points randomly sampled from the survey area. The right plot shows the robot's path in feature space. The red crosses are clustered in feature space to form the $k$ centres. For each of the $k$ centres, the distance to the closest feature from the path is found. The total evaluation score is the sum of all these distances.}
     \label{fig:info:eval_k}
 \end{figure}

To reduce the bias towards large spatial regions in Equation \ref{eq:info:eval_sd}, we propose the following evaluation function $M_{K}$. First, the feature space is clustered using the K-means algorithm to select $K$ points that represent the feature space. The evaluation metric is the sum of all the closest feature distances from each of the $K$ points to any feature extracted from the path.
\begin{equation}
    M_{K} = \sum_{\mathbf{z}_k \in \mathbf{Z}_{K}} R_{m} (\mathbf{z}_k; \mathbf{Z}_{path}),
    \label{eq:info:eval_k}
\end{equation}
where $\mathbf{z}_k$ is one of the centres ($\mathbf{Z}_K$) that represent the latent space, $\mathbf{Z}_{path}$ represents the features extracted from the path being evaluated and $R_m$ is the reward function presented in Equation \ref{eq:info:maha_reward}. This metric and the transition from physical to feature space is displayed in Figure \ref{fig:info:eval_k}.

\subsection{RRT-based Informative Path Planner}\label{sec:info:rrt}

\begin{figure}[!ht]  % old col width = 0.57
    \centering
    \includegraphics[width=0.99\columnwidth]{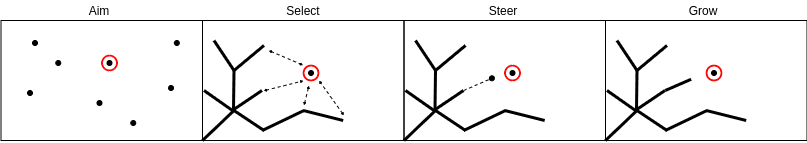}
    \caption{InfoRRT Process. First, the planner selects a target point from the larger survey area. This target point is selected either at random or is the highest reward point from many candidate target points. Next, nodes on the search tree that are closest to the target point are evaluated for expansion. This node is selected such that it will be the highest value, when combined with the target point. A new node is created by steering the node in the direction of the target point, and finally this node is added to the search tree.}
    \label{fig:continous:rrt_diagram}
\end{figure}

\begin{figure}[!ht]
    \centering
    \begin{subfigure}[t]{0.48\columnwidth}
        \centering
        \includegraphics[width=\textwidth]{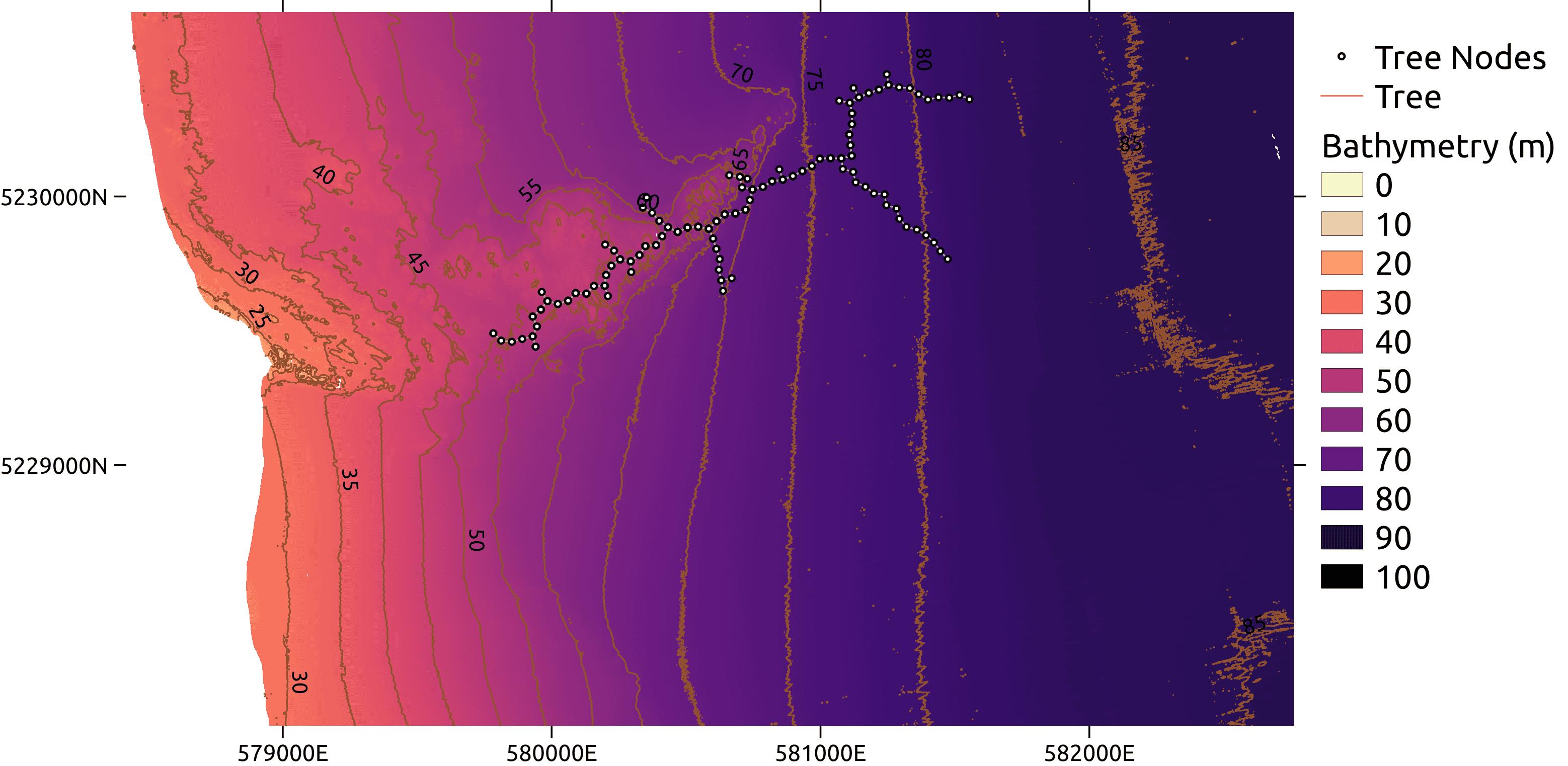}
        \caption{}
        \label{info:rrt:real:tree_only}
    \end{subfigure}
    ~ 
    \begin{subfigure}[t]{0.48\columnwidth}
        \centering
        \includegraphics[width=\textwidth]{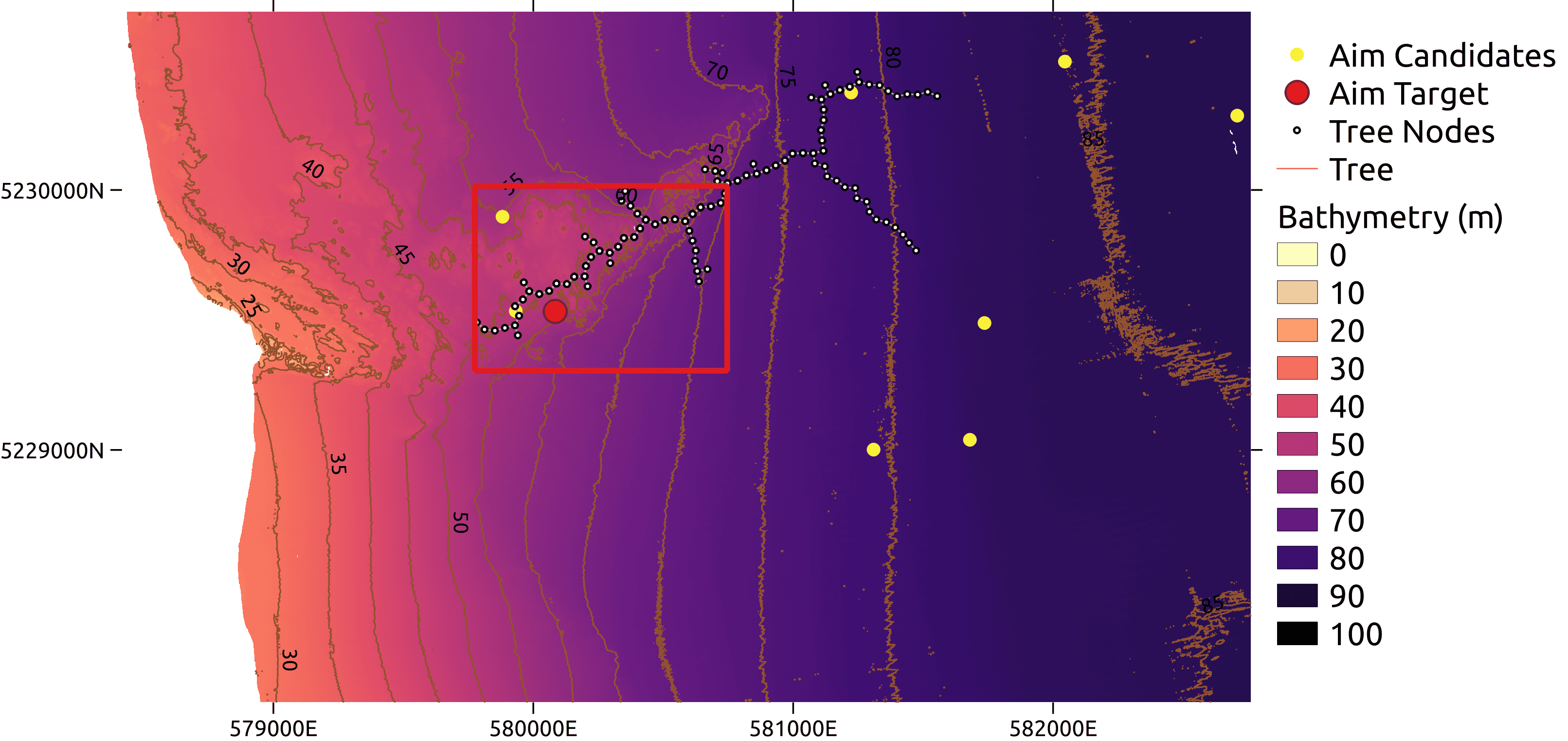}
        \caption{}
        \label{info:rrt:real:aim_candidates}
    \end{subfigure}
    ~ 
    \begin{subfigure}[t]{0.48\columnwidth}
        \centering
        \includegraphics[width=\textwidth]{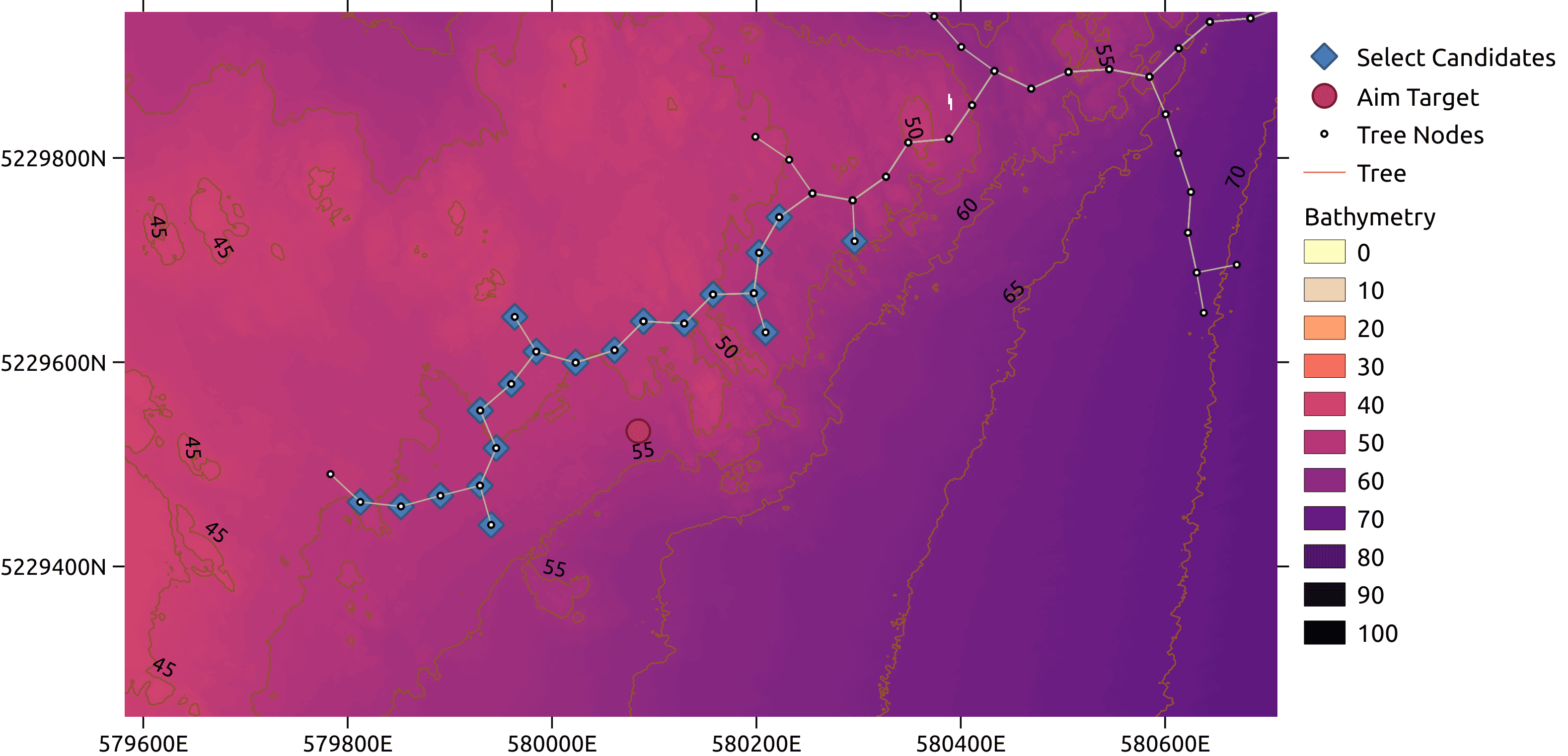}
        \caption{}
        \label{info:rrt:real:select}
    \end{subfigure}
    ~ 
    \begin{subfigure}[t]{0.48\columnwidth}
        \centering
        \includegraphics[width=\textwidth]{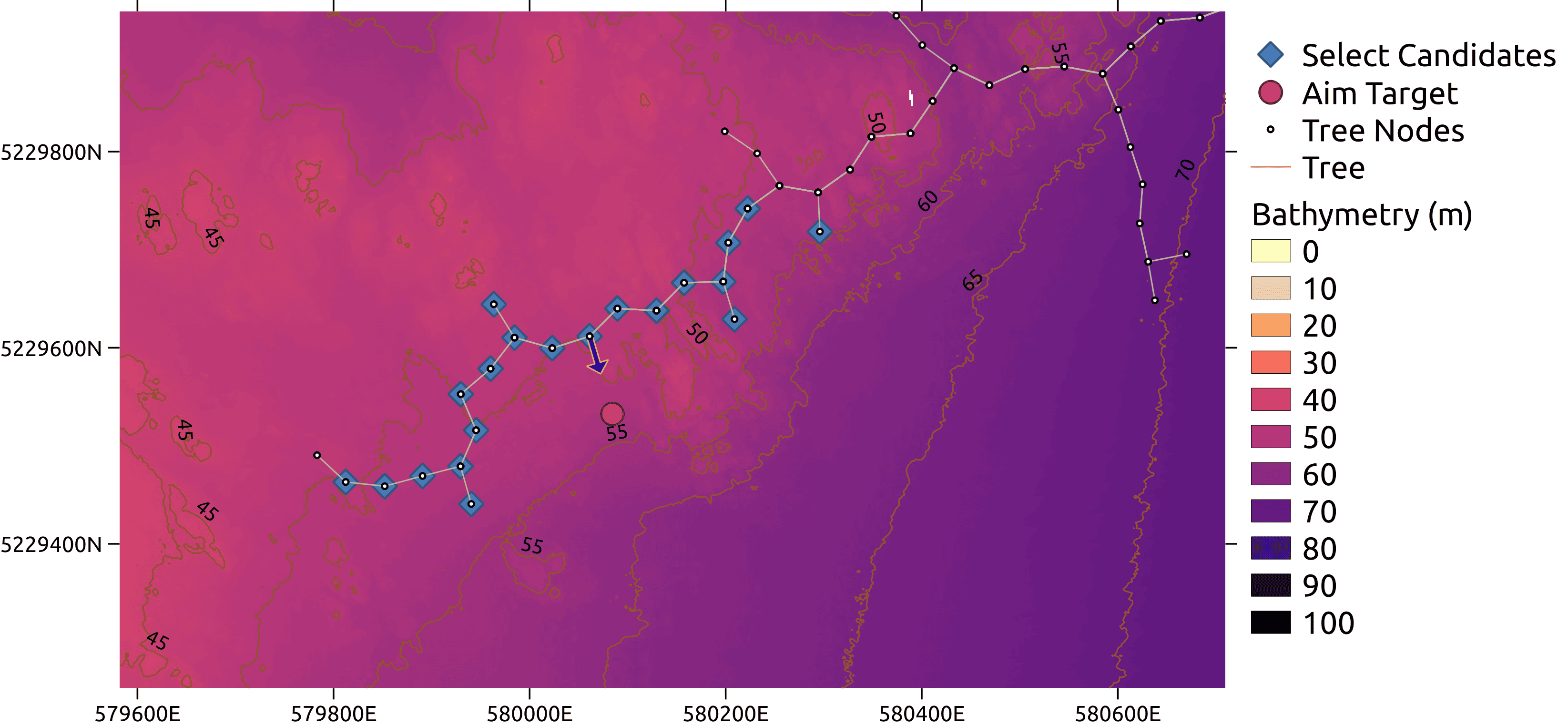}
        \caption{}
        \label{info:rrt:real:expand}
    \end{subfigure}
    ~ 
    \begin{subfigure}[t]{0.48\columnwidth}
        \centering
        \includegraphics[width=\textwidth]{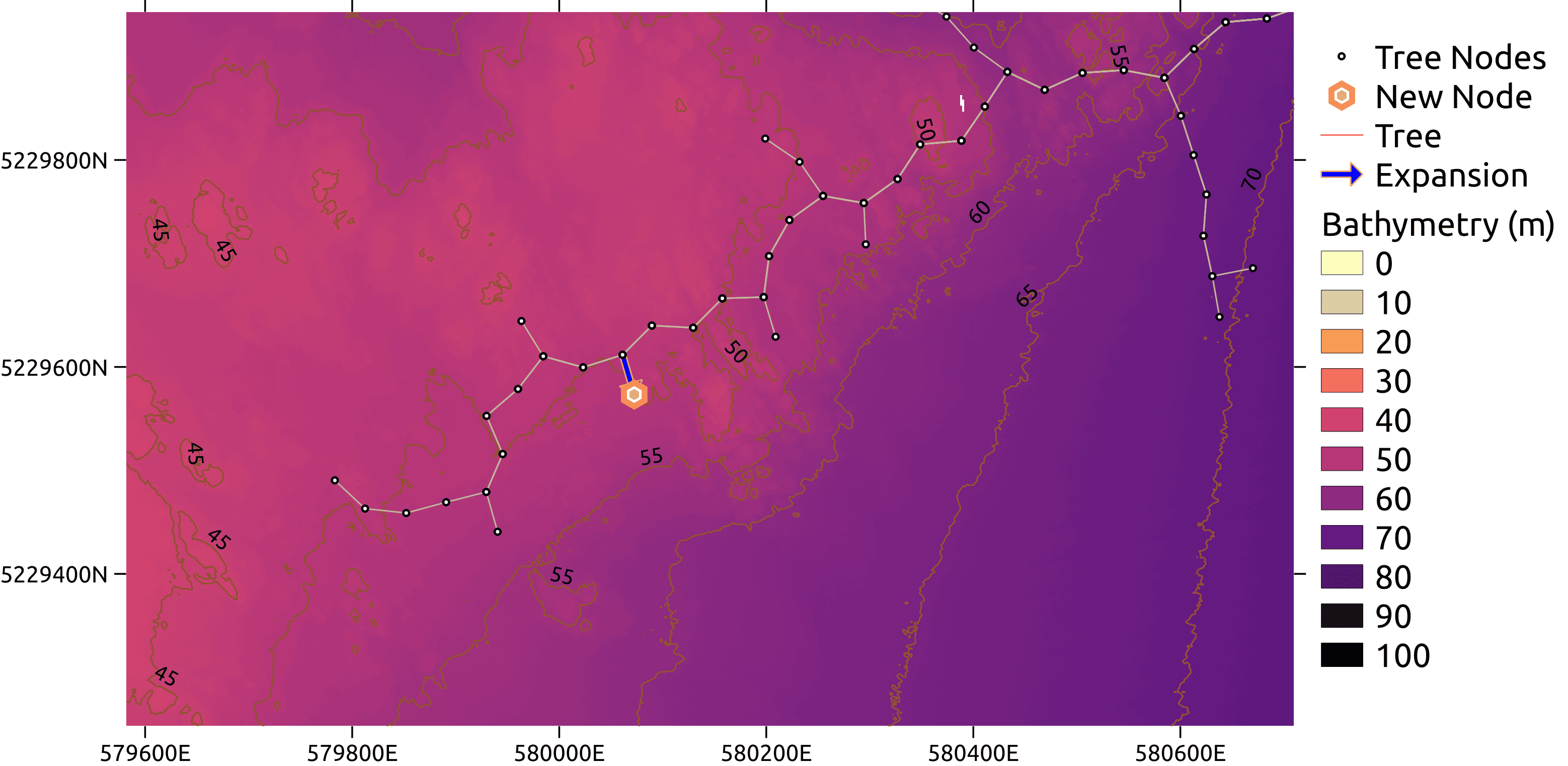}
        \caption{}
        \label{info:rrt:real:new_node}
    \end{subfigure}
    ~ 
    \begin{subfigure}[t]{0.48\columnwidth}
        \centering
        \includegraphics[width=\textwidth]{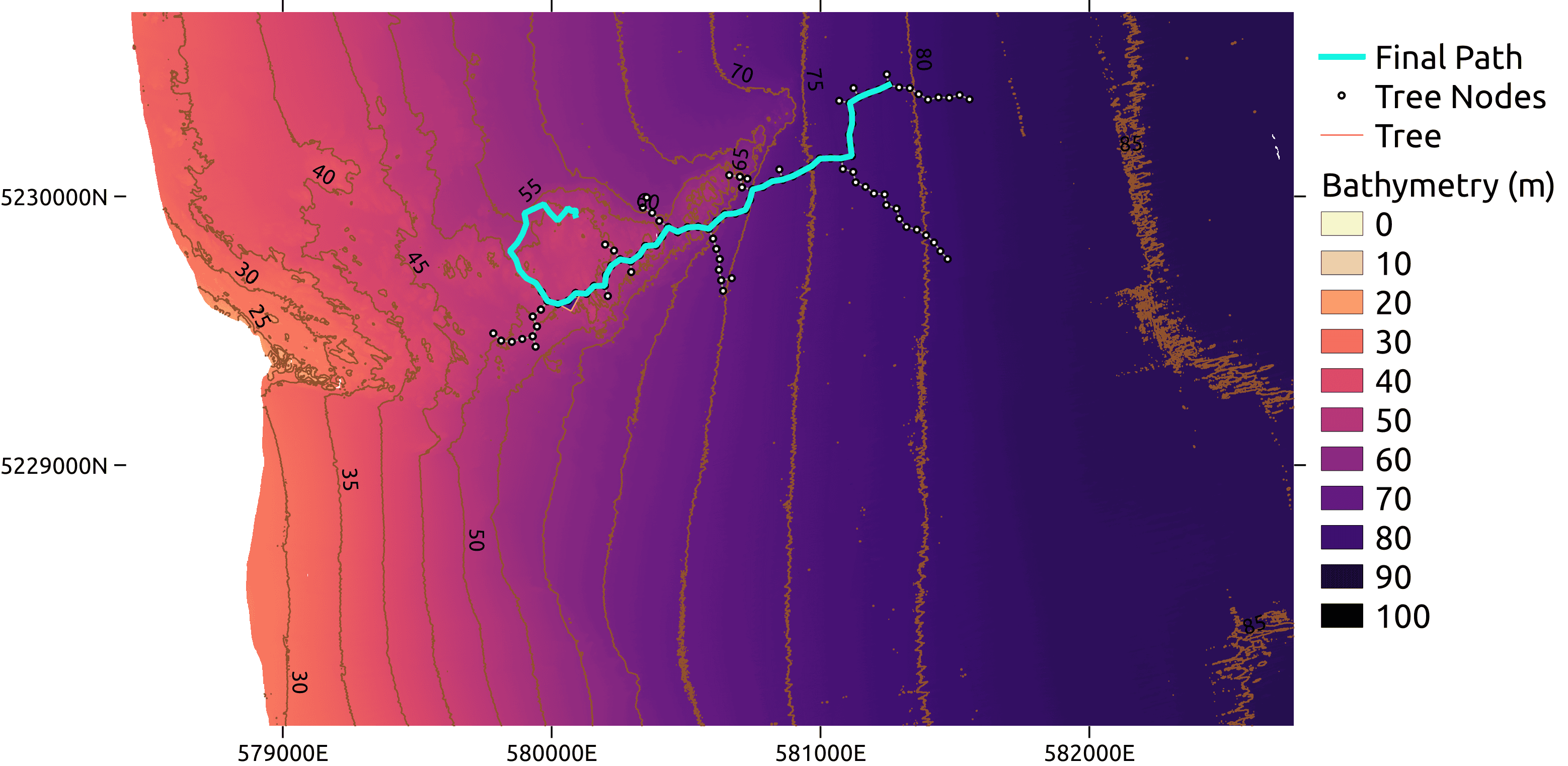}
        \caption{}
        \label{info:rrt:real:final}
    \end{subfigure}
    \caption{An iteration of informative planning using \textit{RRT}. (a) shows the tree before expansion. (b) details the \textit{aim} process, where several candidates from the search space are selected. A target node is selected that has the maximum feature wise distance to nodes already in the tree. (c) details the \textit{select} process, where candidate nodes from the tree are evaluated for expansion. (d) outlines the \textit{expand} process, where the tree is expanded from the selected node towards the target node. (e) shows the new node as part of the expanded tree. (f) shows the final path.}
    \label{info:rrt:real}
\end{figure}

The algorithm presented here is a planner designed to use the information reward (proposed in Section \ref{sec:infoplan:reward}) to explore the feature space. It is primarily based on RIG-Tree \citep{Hollinger2014}, which is in turn based on RRT*, which promotes exploration of the entire environment. This algorithm incrementally grows a tree by expanding branches towards high value areas. For this application, the ability to query all of the robot's environment rather than just within the sensing radius makes the \textit{aim} (outlined below) functionality very effective. The algorithm is summarised in Algorithm \ref{alg:info_rrt} and illustrated in Figure \ref{fig:continous:rrt_diagram}. An iteration of this planner is also displayed in Figure \ref{info:rrt:real}. The key extensions to RIG-Tree / RRT* are the use of the novel reward function outlined in Equation \ref{eq:info:maha_reward}, that allows for incremental building of informative trajectories and consideration of the entire trajectory in the information reward. We also develop a novel interpretation of the \textit{aim} function to direct the tree towards areas that have unique features compared to those already collected by the tree. Furthermore, we propose a method to establish the starting position of the tree that considers the informativeness of the region.

\textit{Find Starting Position}\\
The starting position has a significant impact on the overall path, especially if the path budget is small compared to the size of the survey area. The AUV will not be sampling efficiently if it has to traverse a significant distance to get to informative areas. Therefore, the starting location should be chosen so it places the AUV in an informative region. To estimate the informative starting region, random points are sampled from the environment and features are extracted at each of these points. A subset of these points is selected as potential starting positions. 

\begin{figure}[!ht]  % old col width = 0.57
    \centering
    \includegraphics[width=0.6\columnwidth]{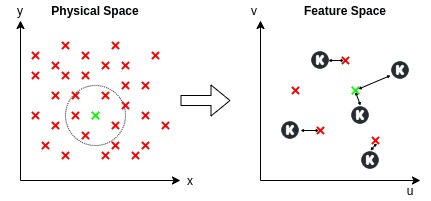}
    \caption{Selecting informative starting positions. Features are extracted from points within a search radius of each candidate starting position (green). The value of the starting position is the sum of the feature distance between all K-centres that approximate the latent space and the closest feature.}
    \label{fig:info:starting_position}
\end{figure}

For each of these potential starting positions, all the points in its vicinity are used to estimate the region's value. The $M_K$ metric is used to estimate the informativeness, which is calculated as the sum of the minimum feature distances from any one of the points in the vicinity, to K-centres distributed throughout the feature space. This method is successful in selecting informative starting positions and typically selects the locations around the interface between bathymetric terrain, for example between reef and sand. A diagram of this process is displayed in Figure \ref{fig:info:starting_position}.

% \todo{algorithm}

% \todo{equation}

\textit{Aim} \\
Select a target point either at random or towards a goal point, as is the case in RRT. For moving towards a goal point, spatial points are randomly sampled from the environment. For each of these points, the reward is calculated between the point and all the other points currently in the tree. The point with the maximum reward is selected. The objective is to direct the tree towards informative areas.

\begin{equation}
    \mathbf{x}_{target} = \argmax_{\mathbf{x}_i \in \mathbf{X}_{space}} [ R_m (\mathbf{z}_i; \mathbf{Z}_{tree})],
\end{equation}
where $\mathbf{X}_{space}$ are locations randomly sampled from the search space, $\mathbf{z}_i$ are the features sampled from the location $\mathbf{x}_i$ and $\mathbf{Z}_{tree}$ is the set of features already collected from the tree.

\textit{Select} \\
Given the target point, find the nearest (spatially) $k$ nodes on the tree. For each of these nodes, assess the reward for moving to this target point. The planner selects the node that has the maximum reward and that will not exceed the maximum distance budget.

\begin{equation}
    v = \argmax_{v_i \in V_{nearest}} [R_m (\mathbf{z}_{target} ; \mathbf{Z}_{path} )],
\end{equation}
where $V_{nearest}$ is the $k$ nearest nodes to the target point, $\mathbf{x}_{target}$. $\mathbf{z}_{target}$ are the features extracted from the point $\mathbf{x}_{target}$ and $\mathbf{Z}_{path}$ are the features collected on the path to node $v_i$.

\textit{Steer} \\

Extend the selected node towards the target point, while checking for collision with any obstacles. For this application, these obstacles are usually areas that are either too shallow or deep for the AUV to operate.

\textit{Grow} \\
Create a new node at this new point and add it to the tree.

\begin{algorithm}[!ht]
\caption{InfoRRT}\label{alg:inforrt}
\begin{algorithmic}[1]

\Function{Aim}{}
     \If{ Aim Towards Informative Area}
        \State $x_{target} \gets \argmax_{x_i \in X_{space}} [ R_m (z_i; \bar{Z}_{tree})]$ \Comment{Select the location that maximises the feature distance to points already in the tree}
    \Else
        \State $x_{target} \gets RandomLocation()$
    \EndIf
    \State \Return $x_{target}$
\EndFunction
\\

\Function{Select}{$x_{target}$}
	\State $V_{nearest} \gets \text{NearestKNodes}(x_{target}, k)$
	\State $v_{selected} \gets \argmax_{v_c \in V_{nearest}}[Rm(z_{target}, Z_{path})]$ \Comment{Select the node to expand that maximises the information reward}
    \State \Return $v_{selected}$
\EndFunction
\\
\Function{Steer}{$v$, $x_{target}$}
\State $x_{new} \gets ExtendTowards(x_v, x_{target},d_{expand})$ \Comment{Extend towards the target location by the expansion distance}
\If { CheckConstraints }
	\State \Return True, $x_{new}$
\Else
	\State \Return False, $x_{new}$
\EndIf
\EndFunction
\\
\Function{Grow}{$x_{new}$}
	\State $R \gets R_m ( z_i, Z_{path})$
	\State $c \gets c + d_{expand}$
	\State $v_{new} \gets \text{Node}(x_{new},R,c)$
\EndFunction

\Procedure{InfoRRT}{}
    \State FindStartingPosition() \Comment{Selects a starting position}
    \While{Computating Budget Remains}
        \State $x_{target} \gets \text{Aim()}$ \Comment{Selects a target point to aim towards}
        \State $v \gets \text{Select}(x_{target})$ \Comment{Selects a node to expand towards the target point}
        \State $\text{valid}, x_{new} \gets \text{Steer}(v, x_{target})$ \Comment{Expands the selected node towards the target point}
        \If{ valid }
        \State $\text{Grow}(x_{new})$ \Comment{Creates a new node and adds it to the tree}
        \EndIf
    \EndWhile
\EndProcedure
\end{algorithmic}
\label{alg:info_rrt}
\end{algorithm}

\FloatBarrier

\subsection{MCTS-based Informative Path Planner}
MCTS is a planning method that incrementally builds a search tree by taking random samples in the decision space \citep{Browne2012}. At each iteration, MCTS attempts to select the optimal node for expansion, based on either exploration (searching in areas that are undersampled) or exploitation (searching in areas that have a higher reward). The value of an action is estimated by performing simulations of further trajectories given this action, using a default policy. 

\begin{figure}[!ht]  % old col width = 0.57
    \centering
    \includegraphics[width=0.9\columnwidth]{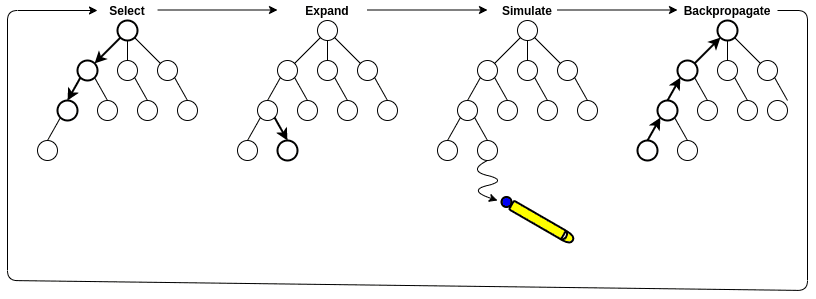}
    \caption{The four stages of MCTS. First, a node is selected for expansion. Next, the selected node is expanded according to the expansion policy, which is to aim towards high value areas. From this expanded node, further actions are simulated to provide an estimate of the node's expected future value. Finally, this value estimate is backpropagated up the tree, updating the value of each node that led to this action.}
    \label{fig:info:mcts_process}
\end{figure}

\begin{figure}[!ht]
    \centering
    \begin{subfigure}[t]{0.48\columnwidth}
        \centering
        \includegraphics[width=\textwidth]{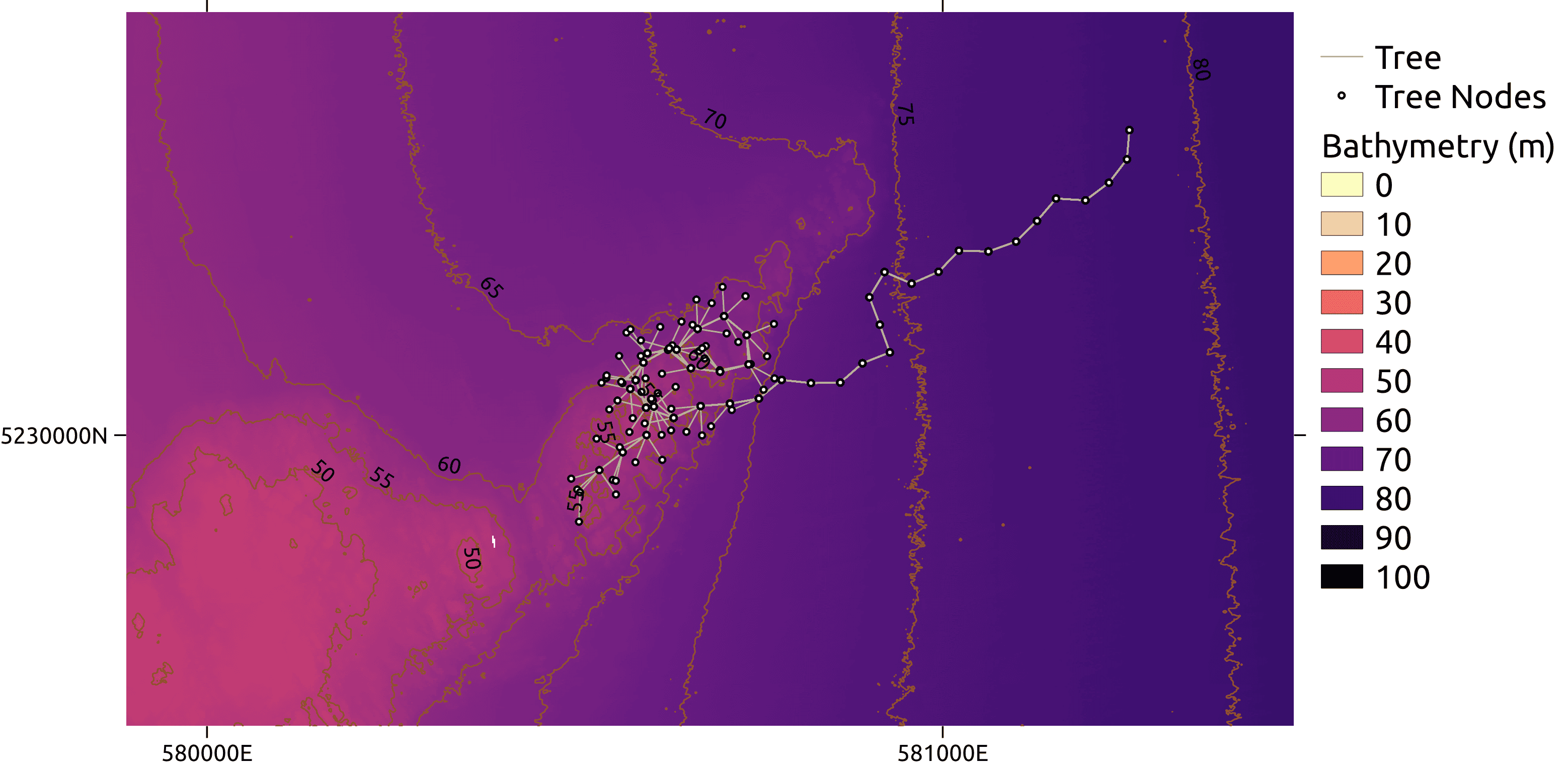}
        \caption{}
        \label{info:mcts:real:tree_only}
    \end{subfigure}
    ~ 
    \begin{subfigure}[t]{0.48\columnwidth}
        \centering
        \includegraphics[width=\textwidth]{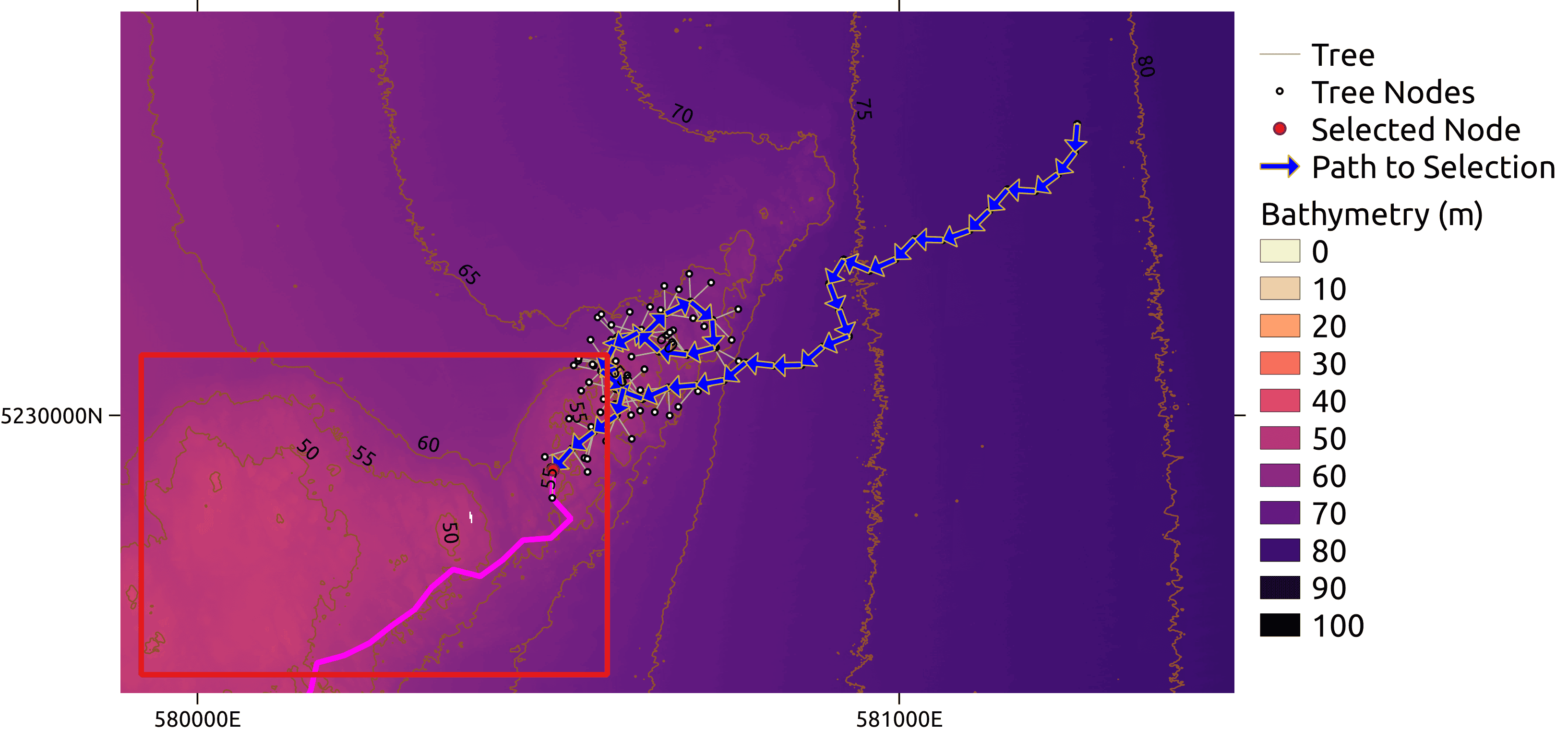}
        \caption{}
        \label{info:mcts:real:select}
    \end{subfigure}
    ~ 
    \begin{subfigure}[t]{0.48\columnwidth}
        \centering
        \includegraphics[width=\textwidth]{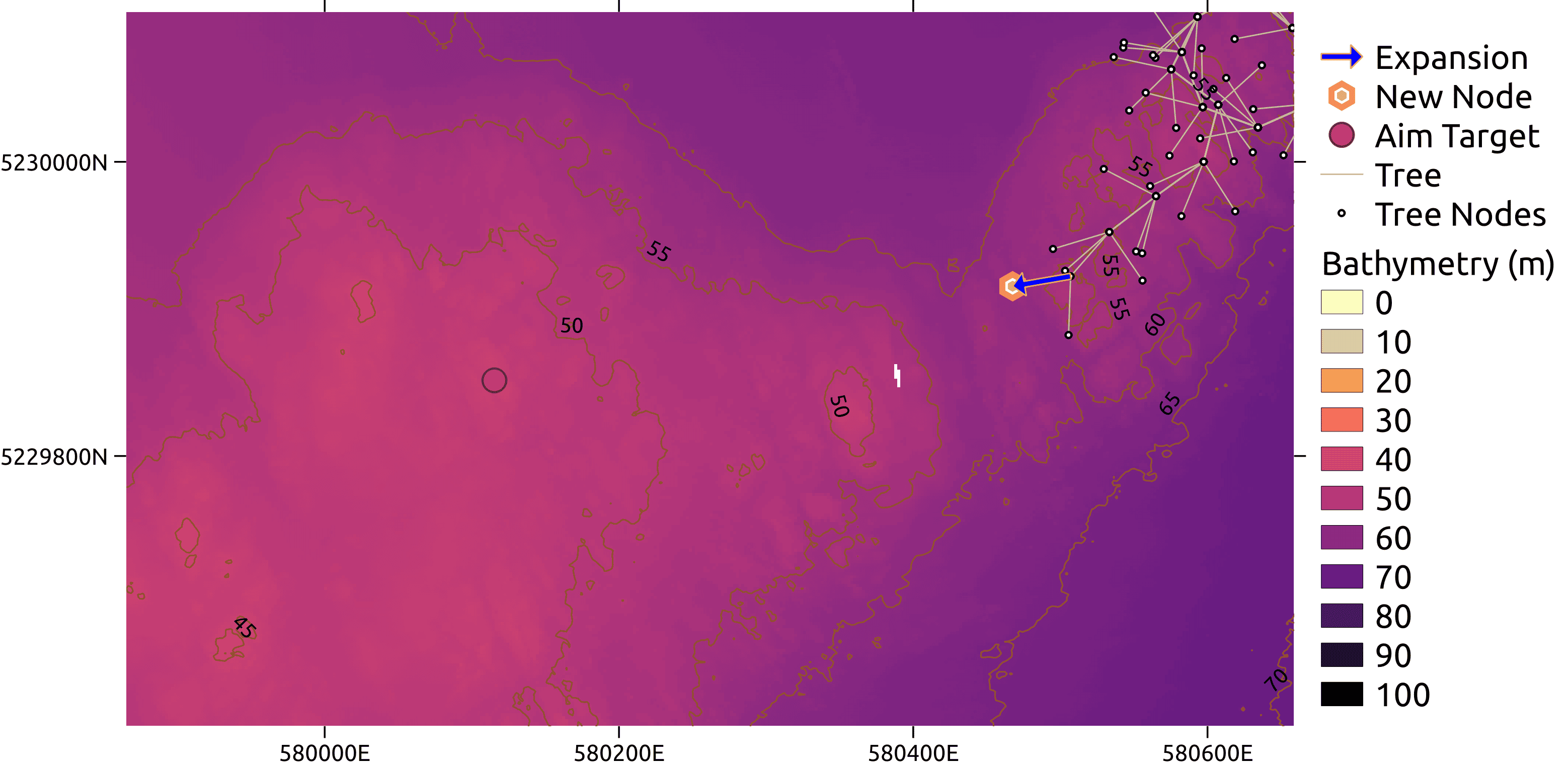}
        \caption{}
        \label{info:mcts:real:expand}
    \end{subfigure}
    ~ 
    \begin{subfigure}[t]{0.48\columnwidth}
        \centering
        \includegraphics[width=\textwidth]{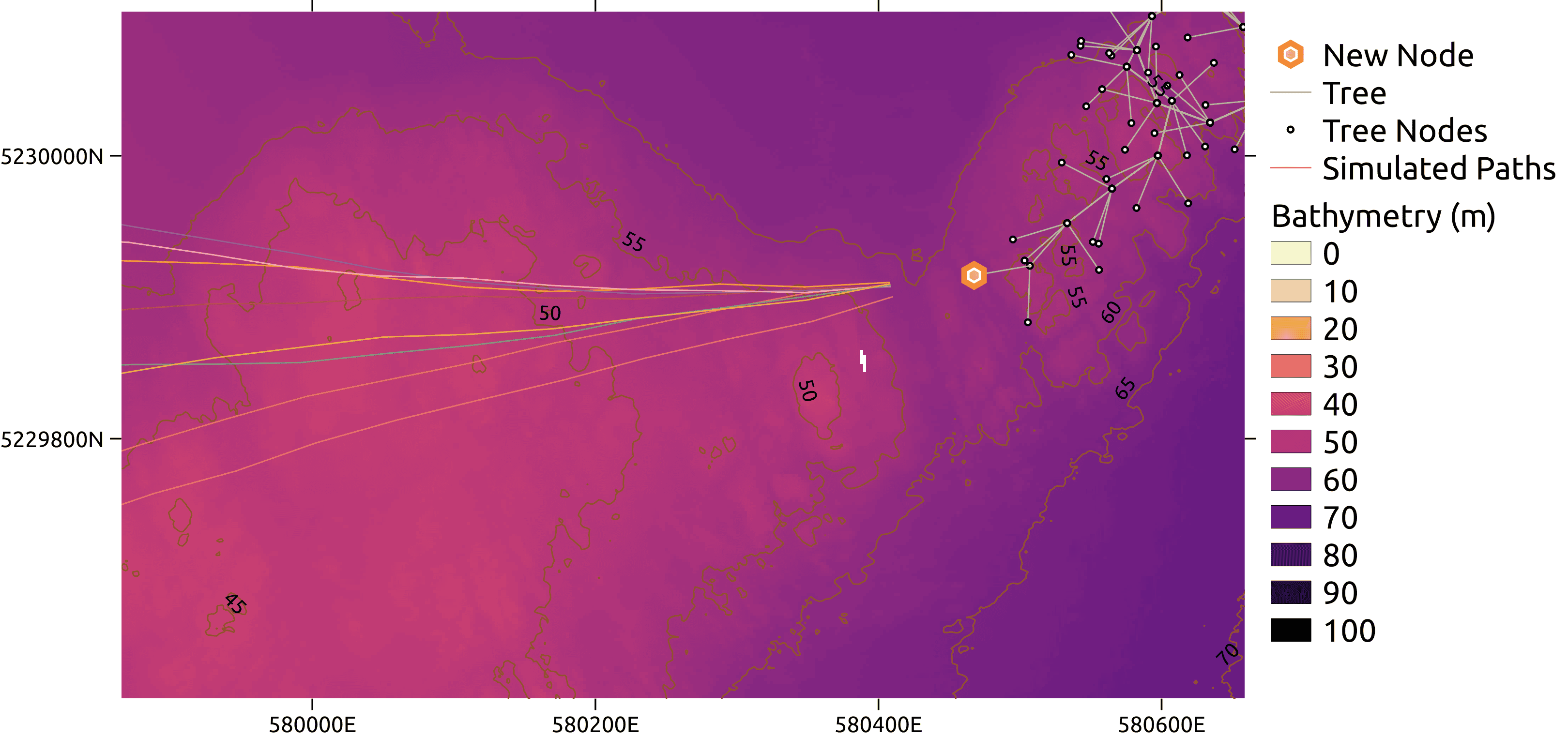}
        \caption{}
        \label{info:mcts:real:simulate}
    \end{subfigure}
    ~ 
    \begin{subfigure}[t]{0.48\columnwidth}
        \centering
        \includegraphics[width=\textwidth]{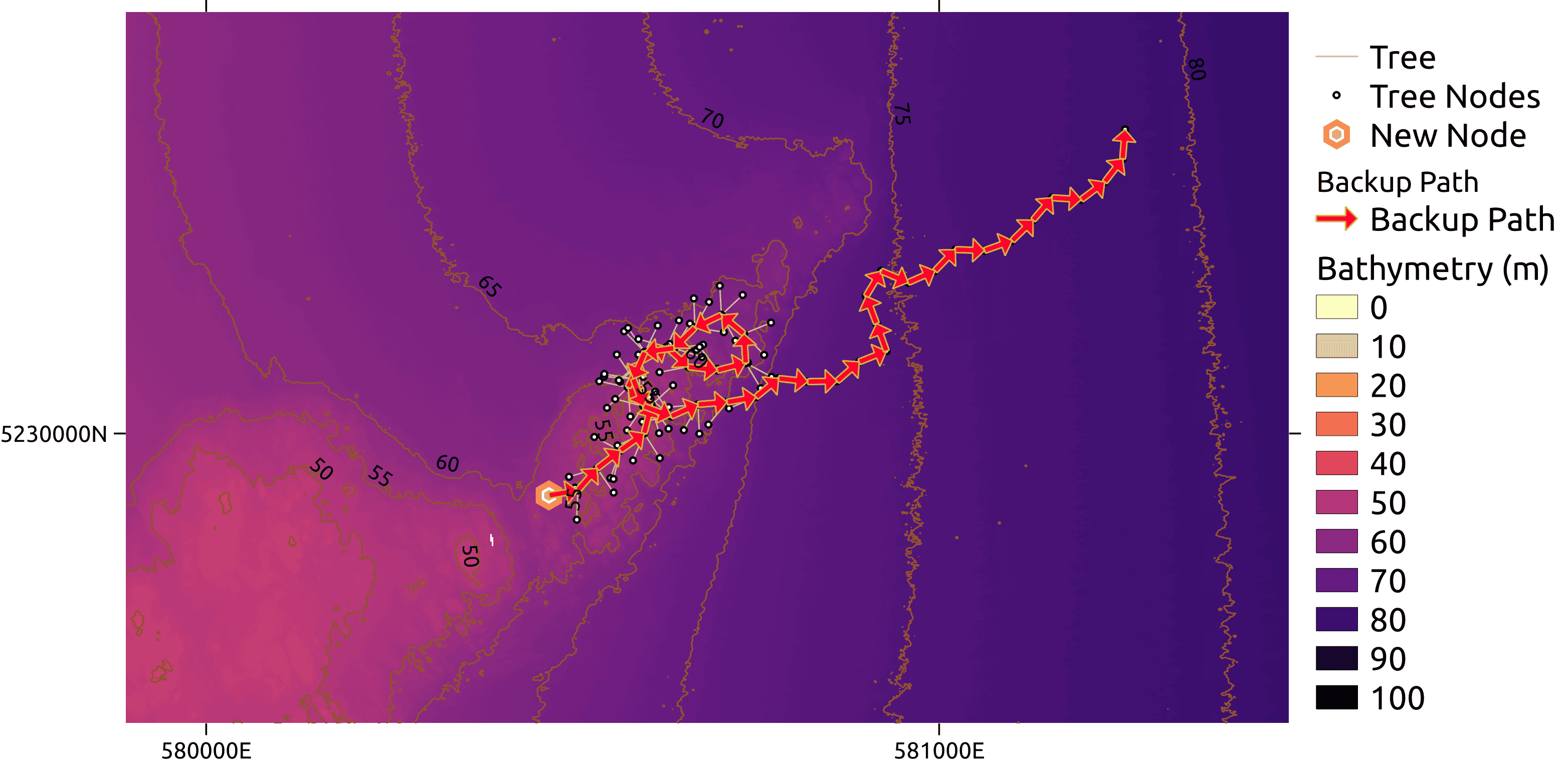}
        \caption{}
        \label{info:mcts:real:backup}
    \end{subfigure}
    ~ 
    \begin{subfigure}[t]{0.48\columnwidth}
        \centering
        \includegraphics[width=\textwidth]{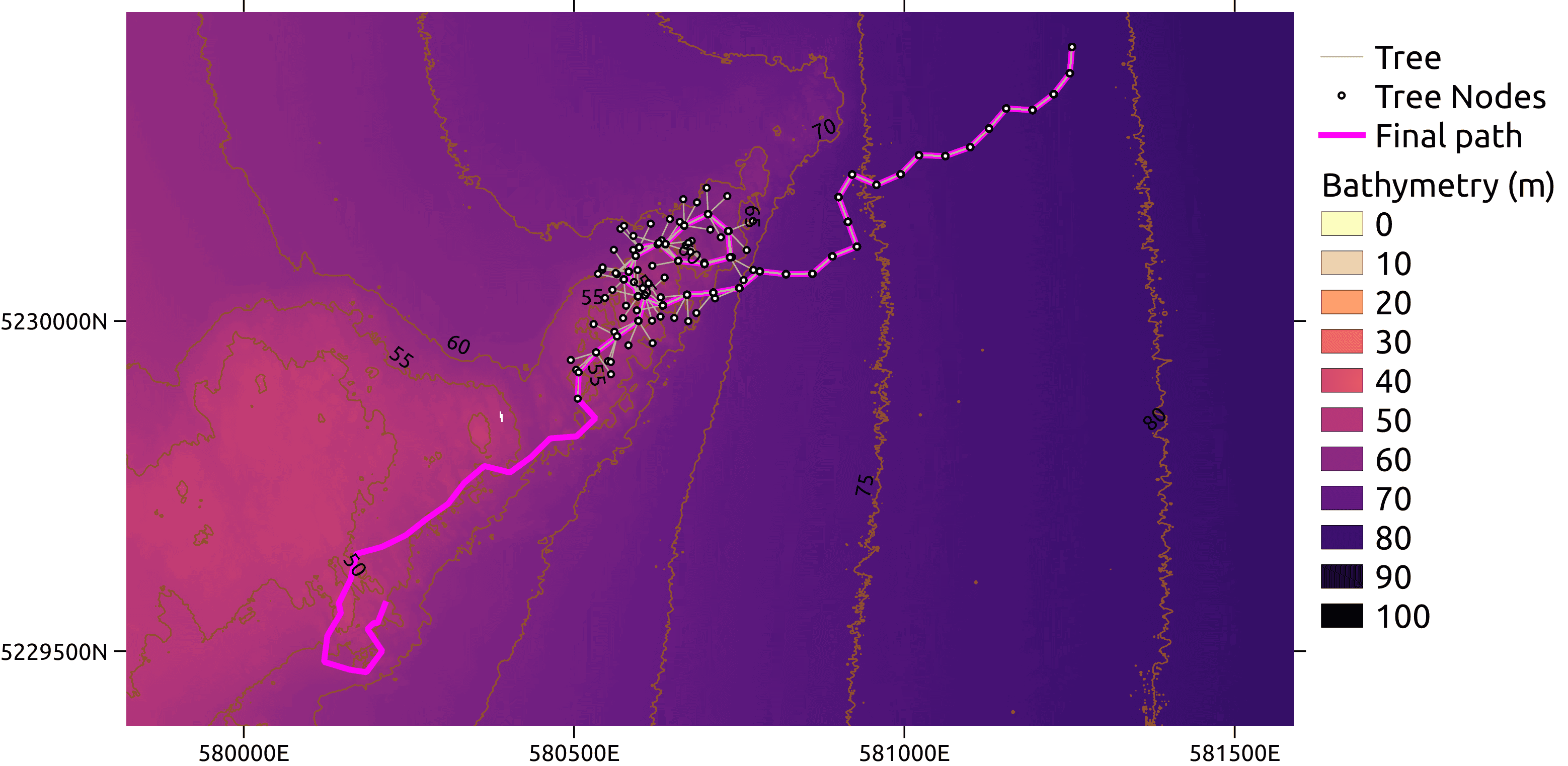}
        \caption{}
        \label{info:mcts:real:final}
    \end{subfigure}
    \caption{An iteration of informative planning using \textit{MCTS}. (a) shows the tree before expansion. (b) highlights the selection process, where a node is selected for expansion based on the UCB-1 algorithm \citep{Browne2012} (c) details the expansion process, where the tree is expanded from the selected node, to an informative target (d) shows the simulated future rollouts, (e) shows the backpropagation process, where the value of the tree is updated from the simulation result. (f) shows the final path.}
    \label{info:mcts:real}
\end{figure}

MCTS provides a solid base for an information gathering planner as it is capable in large decision spaces and effectively balances exploration and exploitation.  This research extends MCTS to explore the feature space by guiding the planner using the reward function outlined in Equation \ref{eq:info:maha_reward}. Furthermore, it promotes natural exploration by occasionally aiming towards high value areas, by incorporating the \textit{aim} functionality from RRT*. Many MCTS implementations operate with discrete actions where the possible actions from a specific node are predefined, such as in game-playing (Go, Chess etc.) or where a continuous space is divided into a grid. This implementation of MCTS uses continuous spaces to build its search tree. It manages the increased complexity by limiting the number of expansions per node and also preventing a random action from a specific node being too similar to a previous action. Operating on continuous spaces allows the planner to better fit the environment while also allowing easier integration of robot constraints. A diagram of the MCTS process is included in Figure \ref{fig:info:mcts_process}, while a visualisation of a single iteration of the planner is illustrated in Figure \ref{info:mcts:real}.

\subsection{MCTS Process}

\textit{Find Starting Position} \\
The method for finding the start position is identical to that use in the \textit{RRT} method, which finds an informative region in which to expand the tree.

\textit{Selection}\\
MCTS aims to expand nodes based on a \lq best first' policy. The node to expand is based on the UCB-1 algorithm \citep{Kocsis2006} that is found to optimally balance exploration and exploitation.

\begin{equation}
    v = \argmax_{v_c \in v_{children}} \left[ Q_v + C_{uct} \sqrt{\frac{\ln N_v}{N_{v_c}}}\, \right],
    \label{eq:mcts:select_uct}
\end{equation} 

where $v$ is the node evaluated, $Q_v$ is the expected future reward given from visiting node $v$, $C_{uct}$ is the exploration vs exploitation constant, here with a value of $\sqrt{2}$. $N$ is the number of times the parent node has been visited and $N_{v_c}$ is the number of times the child node has been visited.

\textit{Expansion}\\
Inspired from our RRT approach, the selected node is expanded either randomly or towards high value areas. Both node expansions have to satisfy constraints, including the direction change being within an acceptable range to stop excessive jitter in the path.

The expand function can be used to direct the AUV towards informative areas. To encourage exploration, the informative areas are identified as areas that are different to features already in the tree. This selects the location that maximises the feature distance to all nodes currently in the tree.
\begin{equation}
    \mathbf{x}_{new} = \argmax_{\mathbf{x}_i \in \mathbf{X}_{space}} [ R_m (\mathbf{z}_i; \mathbf{Z}_{tree})],
\end{equation}
where $\mathbf{x}_{new}$ is the new location of the expanded node, $\mathbf{x}_i$ is a location from $\mathbf{X}_{space}$ that has been randomly sampled from the environment, $\mathbf{z}_i$ is the feature extracted from the remotely-sensed data at point $\mathbf{x}_i$ and $\mathbf{Z}_{tree}$ is the set of features collected in the entire search tree. 

During this process, the path is checked to ensure there are no obstacles. If an obstacle does exist, the expansion process is skipped and the planner begins the selection process again.

\textit{Simulation} \\
In the simulation phase, the value of an action is estimated by performing multiple simulations of future actions. For this application, AUV paths are generated until the distance budget is exceeded or until the simulated path hits an obstacle. The simulation follows the same action policy as the expansion step. The simulation can be run multiple times to provide a better estimate of the expected future rewards. The value for the simulation becomes:
\begin{equation}
	r_{sim} = \frac{1}{N_{s}}\sum_{i=0}^{N_{s}} \sum_{\mathbf{z}_j \in \mathbf{Z}_{path}} R_m (\mathbf{z}_j; \mathbf{Z}_{path}),
\end{equation}
where $N_s$ is the number of simulations run, $\mathbf{Z}_{path}$ is the set of features collected during each simulation run.

\textit{Backpropagation} \\
After a simulation is performed, this evaluation is then used to update the search tree. Starting from the recently expanded node and moving up the tree until it reaches the root node, each node value is re-evaluated. As there are incremental rewards for subsequent nodes, we use the valuation function from \citet{Patten2018}:

\begin{equation}
    Q_v = \eta^\tau R_v + \frac{1}{N_v} \left( r_v + \sum_{v_c \in v_{children}} N_{v_c} Q_{v_c} \right),
\end{equation}
where $Q_v$ is the estimated value of the node, $\eta$ is the discount factor, $\tau$ is the depth of the current node, $R_V$ is the immediate reward for visiting the node, $N_v$ is the number of visits to the current node, $r_v$ is the simulation reward for the current node, $v_c$ is a child node of the node currently being evaluated, $N_{v_c}$ is the number of visits to the child node and $Q_{v_c}$ is the estimated value of the child node.

\begin{algorithm}[ht]
\caption{InfoMCTS}\label{alg:mcts}
\begin{algorithmic}[1]

\Function{Select}{}\label{mcts:line:function_select}
    \State $v \gets RootNode() $
    \While{NotTerminal} \Comment{Starting at the root node, select a node for expansion}
        \State $v \gets \argmax_{v_c \in v_{children}} [ Q_v + C_{uct} \sqrt{\frac{log N}{N_{v_c}}}]$
    \EndWhile
    \State \Return $v$
\EndFunction
\\
\Function{Act}{$v$}\label{mcts:line:function_act}
    \If{ Aim Towards Informative Area}
        \State $X_{space} \gets ManyRandomLocations()$
        \State $x_{target} \gets \text{argmax}_{x_i \in X_{space}} [ R_m (z_i; Z_{tree})]$ \Comment{Select a target location that maximises the feature-wise distance compared to features already in the search tree}
        \State $x_{new} \gets ExtendTowards(x_v, x_{target},d_{expand})$ \Comment{Extend towards the target location by the expansion distance}
    \Else
        \State $x_{new} \gets RandomlyExtendPath()$ \Comment{Extend the path by the expansion distance}
    \EndIf
    \State \Return $x_{new}$
\EndFunction

\\
\Function{Expand}{$v$}\label{mcts:line:function_expand}
    \State $x_{new} \gets \text{Act}(v)$
    \State $R \gets R_m(z_i ; Z_{path})$
    \State $c \gets c + d_{expand}$
    \State $v_{new} \gets Node(x_{new}, R, c)$ \Comment{Create a new node, with a new state, reward and cost}
    \State \Return $v_{new}$
%    \textcolor{red}{Continue}
\EndFunction

\\
\Function{Simulate}{$v$}\label{mcts:line:function_simulate}
    \For{$s=1,2,\ldots,\text{Simulations}$}
        \State $c \gets Cost(v)$
        \State $x \gets State(x)$ 
        \While{$c < Distance Budget$} \Comment{Simulate until budget is exhausted}
            \State $x \gets Act(v)$ 
            \State $c \gets c + d_{expand}$
            \State $r \gets r + R_m (z; Z_{path})$ 
        \EndWhile
    \EndFor
\EndFunction

\\
\Function{Backpropagation}{$v$}\label{mcts:line:function_backup}
    \While{$v$ \text{is not} $v_{root}$} \Comment{Move back up the tree, updating the expected reward}
    	\State $Q_v \gets \eta^\tau R_v + \frac{1}{N_v} ( r_v + \sum_{v_c \in v_{children}} W_{v_c} Q_{v_c} )$
        \State $v \gets v_p$
    \EndWhile
\EndFunction
\\
\algstore{mctsalg}
\end{algorithmic}
\end{algorithm}
\FloatBarrier

\begin{algorithm}
\begin{algorithmic} [1]                   % enter the algorithmic environment
\algrestore{mctsalg}
\Procedure{InfoMCTS}{}
    \State $x_{start} \gets FindStartingPosition()$ \Comment{Selects a starting position}
    \State $v \gets Node(x_{start}, 0, 0)$ \Comment{Creates the starting node, with zero reward and cost}
    \While{}
        \State $v \gets Select()$ \Comment{Select a node for expansion}
        \State $v \gets Expand(v)$ \Comment{Expand the selected node}
        \State $Simulate(v)$ \Comment{Simulate further actions from this node}
        \State $Backpropagation(v)$ \Comment{Adjust the tree values}
    \EndWhile
\EndProcedure
\end{algorithmic}
\end{algorithm}
\FloatBarrier

% \FloatBarrier

\subsection{Informative Survey Template Placement}

Survey templates are routinely used for planning benthic surveys. A survey template is beneficial as it is easily interpreted by human operators and guarantees a spatially balanced survey. However, the set template may not correspond with the interesting areas of the terrain. Survey templates are traditionally planned manually to cover areas of interest. Alternatively, these templates can be informatively placed to ensure the template maximises its exploration of the feature space. This approach has been pioneered by \citet{Bender2013a}, who placed an informative survey so the distribution of features on the survey matched the distribution of features found in the overall survey area. We propose an informative placement method that evaluates many different candidate placements with the evaluation function outlined in Equation \ref{eq:info:eval_k}, in order to uniformly sample the feature space of the survey area. This can be defined as:

\begin{equation}
	\mathbf{X}_{path} = \argmin_{\mathbf{X}_{path} \in \text{RandomPlacements}}[M_K (\mathbf{Z}_{path}; \mathbf{Z}_{area})],
\end{equation}

where $\mathbf{X}_{path}$ is a candidate survey with features $\mathbf{Z}_{path}$, that is evaluated with the metric $M_K$ using the reference features for the entire survey area ($\mathbf{Z}_{area}$).

\begin{figure}[!ht]  % old col width = 0.57
    \centering
    \includegraphics[width=0.8\columnwidth]{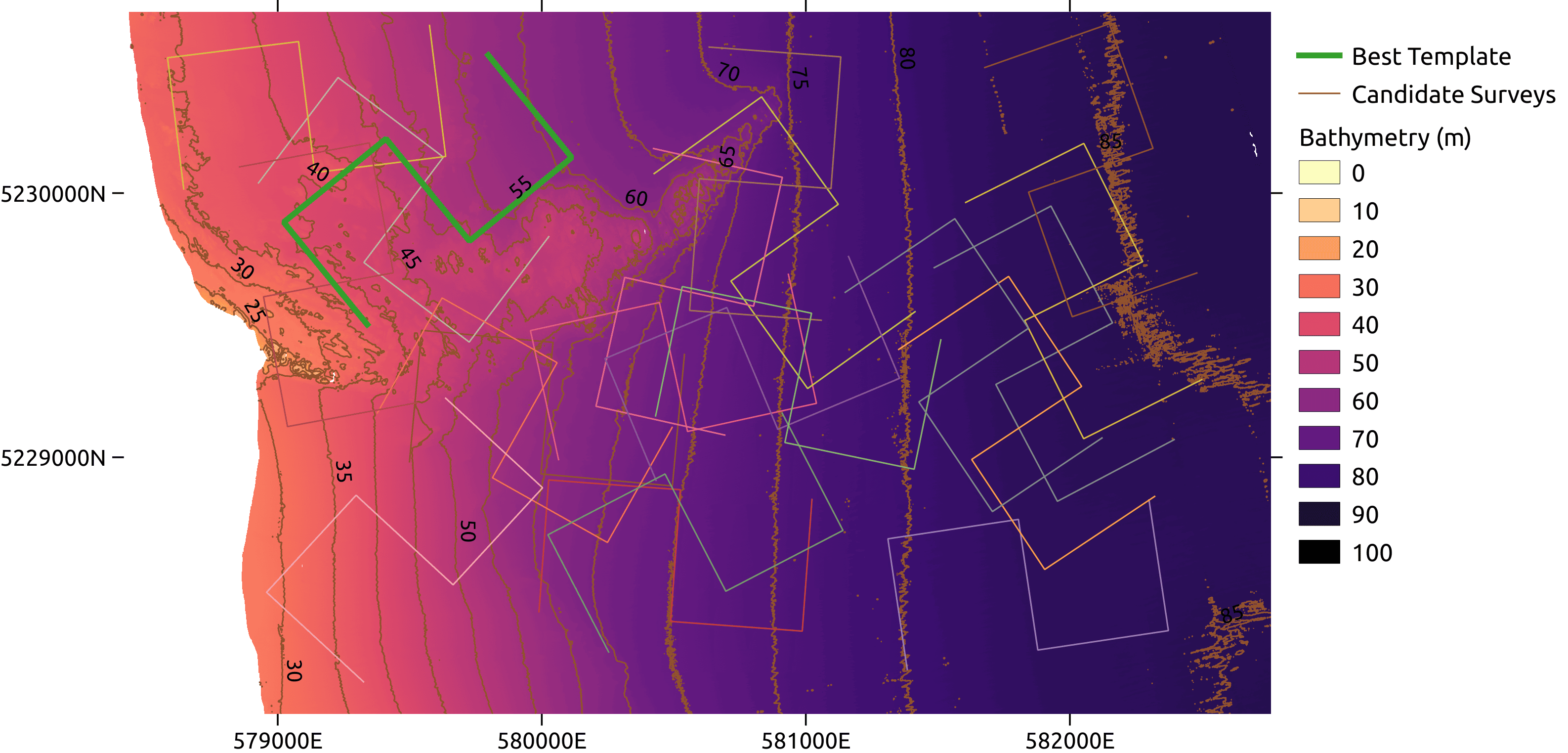}
    \caption{Shows the informative placement of survey templates. Many surveys are randomly placed in the survey area, by altering the location and orientation. The thin lines are a subselection of the candidate surveys. Here 20 surveys are shown, while in the planning process at least 1000 surveys are used. The green indicates the most informative survey.}
    \label{fig:templates:candidates}
\end{figure}

The starting location, orientation and type of survey template are the degrees of freedom when placing a survey. There is an endless array of survey templates to choose from. In this paper, we consider a broad-grid consisting of five equal segments, with each of the sides being one-fifth of the total distance budget in length. This is designed to satisfy the guidelines for spatially-balanced surveys presented in \citet{Foster2014}. The informative placements of surveys is displayed in Figure \ref{fig:templates:candidates}, showing a selection of randomly placed surveys and the most informative survey presented in green.

% \begin{figure}[H]  % old col width = 0.57
%     \centering
%     \includegraphics[width=0.3\columnwidth]{Figures/Methods/Templates/Templates.png}
%     \caption{Informative survey template placement. Location, orientation and template type are randomly varied. The red surveys are the candidate survey, while the green indicates the most informative survey.}
%     \label{fig:templates:candidates}
% \end{figure}

\FloatBarrier

\subsection{Cluster-TSP}\label{sec:info:cluster_tsp}

This method proposes a set of points that together represent the feature space and plans a path that visits these points. The bathymetry for the entire planning space is first projected into the latent space by the bathymetry encoder. The corresponding features (i.e. latent space coordinates)  are clustered using the Gaussian Mixture Models (GMM) clustering algorithm \citep{Reynolds2009}. GMMs fit a given number of Gaussians to the training data using the Expectation-Maximisation algorithm. The GMM is used to assign a cluster label to each of the bathymetry patches, which can then be visualised is the spatial domain, where the pixel value is the cluster assigned at the spatial coordinates for that bathymetric patch. The clusters in feature space tend to form several disjoint patches in the spatial domain. Each spatial cluster grouping in the spatial domain is represented as a polygon.

\begin{figure}[!ht]  % old col width = 0.57
    \centering
    \includegraphics[width=0.7\columnwidth]{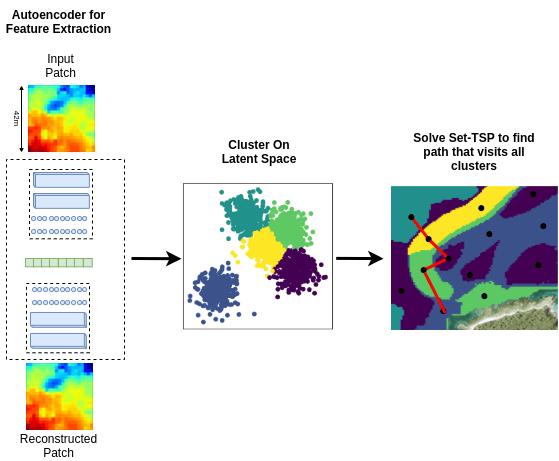}
    \caption{Process diagram for the Cluster-TSP method. Bathymetric patches of the entire survey area are extracted from the raster and projected into the latent space of the autoencoder. Clustering is then run on this feature space. Finally, a set-TSP problem is solved to find an AUV path that visits each cluster type and hence explores the feature space.}
    \label{fig:info:cluster:process}
\end{figure}

\begin{figure}[!ht]
    \centering
    \begin{subfigure}[t]{0.48\columnwidth}
        \centering
        \includegraphics[width=\textwidth]{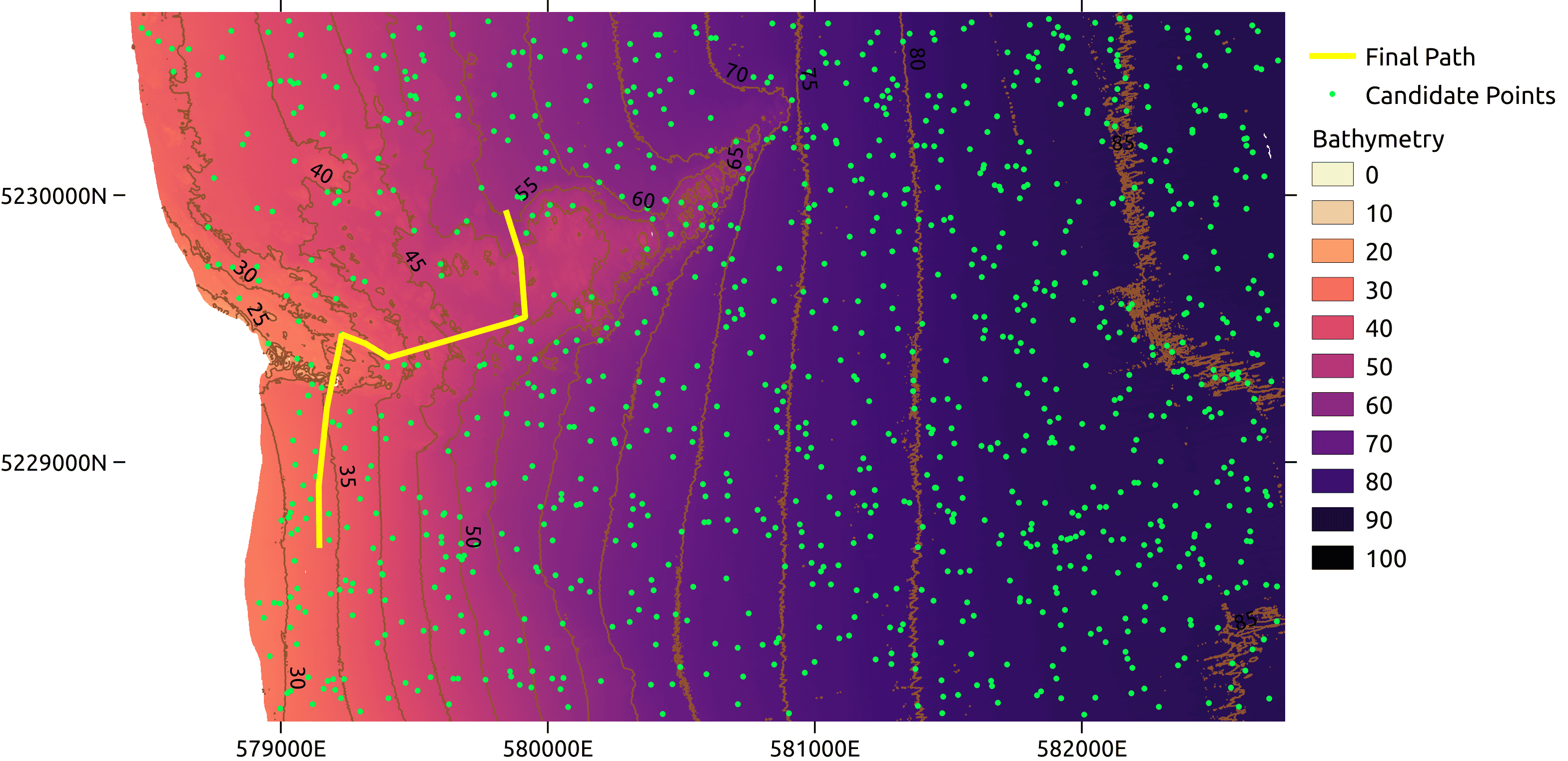}
        \caption{}
        \label{fig:info:tsp:demo_bathy}
    \end{subfigure}
    ~ 
    \begin{subfigure}[t]{0.48\columnwidth}
        \centering
        \includegraphics[width=\textwidth]{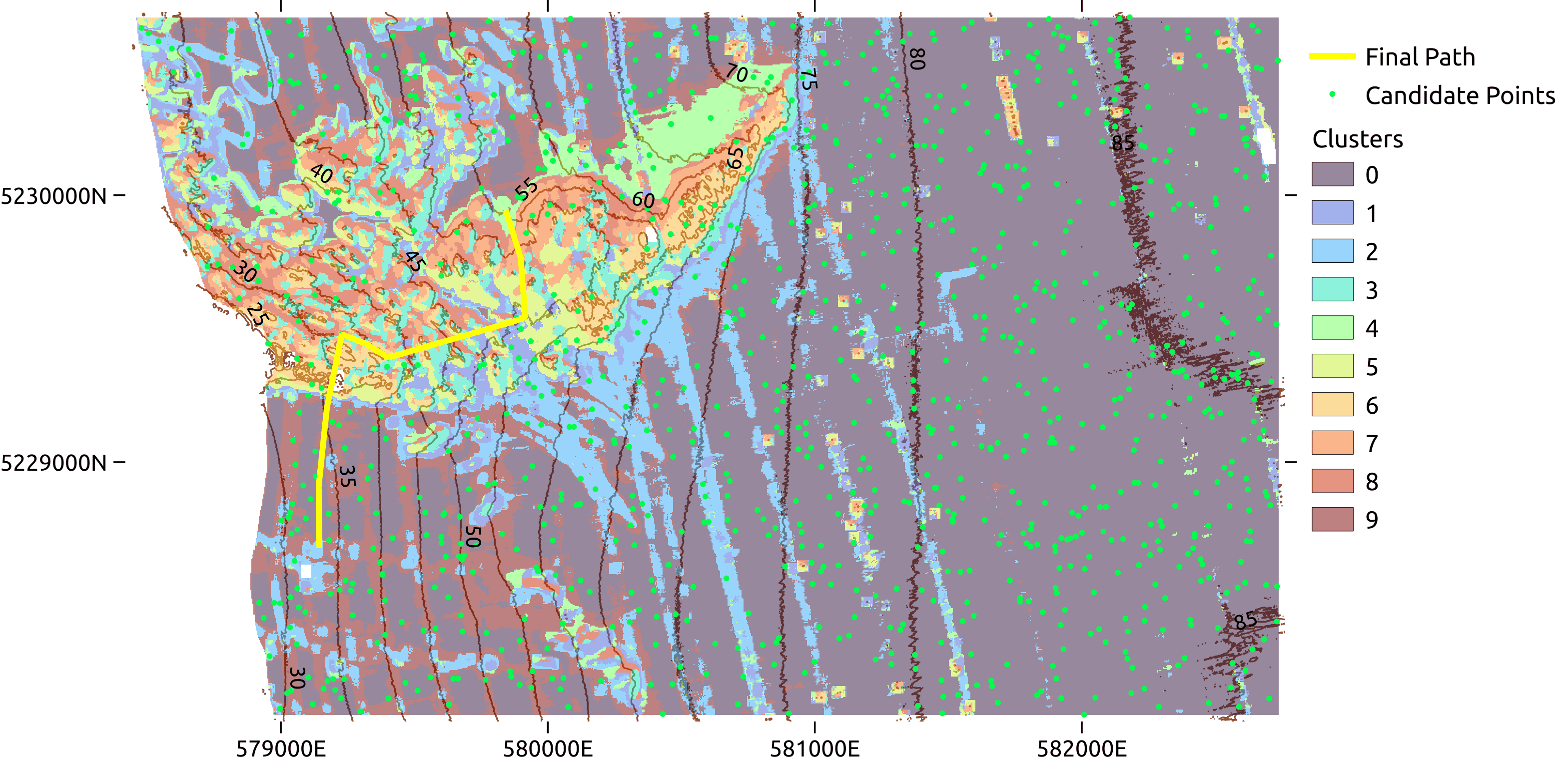}
        \caption{}
        \label{fig:info:tsp:demo_clusters}
    \end{subfigure}
    \caption{Placement of nodes and the final path on the O'Hara dataset (see Section \ref{sec:results}). Nodes are distributed uniformly across the survey area. The final path focusses on cluster groups in and around O'Hara reef to minimise the path length.}
    \label{fig:info:tsp:demo}
\end{figure}

Representative points for each cluster in the latent space are randomly sampled from each corresponding cluster polygon in the spatial domain. To explore the feature space, at least one of each cluster type should be visited in the spatial domain. Typically there will be several cluster polygon choices to visit for each cluster type. This problem can be framed as a set-TSP (Travelling Salesman Problem). Set-TSP is a generalisation of the travelling salesman problem, where the goal is to visit one of each type of node. To solve the Set-TSP problem the OR Tools library \citep{Perron2019} is leveraged, which uses hill-climbing or simulated annealing to optimise the routing between the nodes, where the nodes are points selected from each spatial cluster grouping. \textcolor{blue}{This process is visualised in Figure \ref{fig:info:cluster:process}, while Figure \ref{fig:info:tsp:demo} shows a demonstration of a path planned from the candidate nodes}. 

Both geometric and encoded features can be used with this method, however the clusters generated when using geometric features are excessively noisy in the spatial domain and do not form well defined spatial clusters. Hence only the encoded features are used for this method.

\FloatBarrier

\section{Results}\label{sec:results}

\begin{figure}[!ht]
    \centering
    \begin{subfigure}[t]{0.48\columnwidth}
        \centering
        \includegraphics[width=\textwidth]{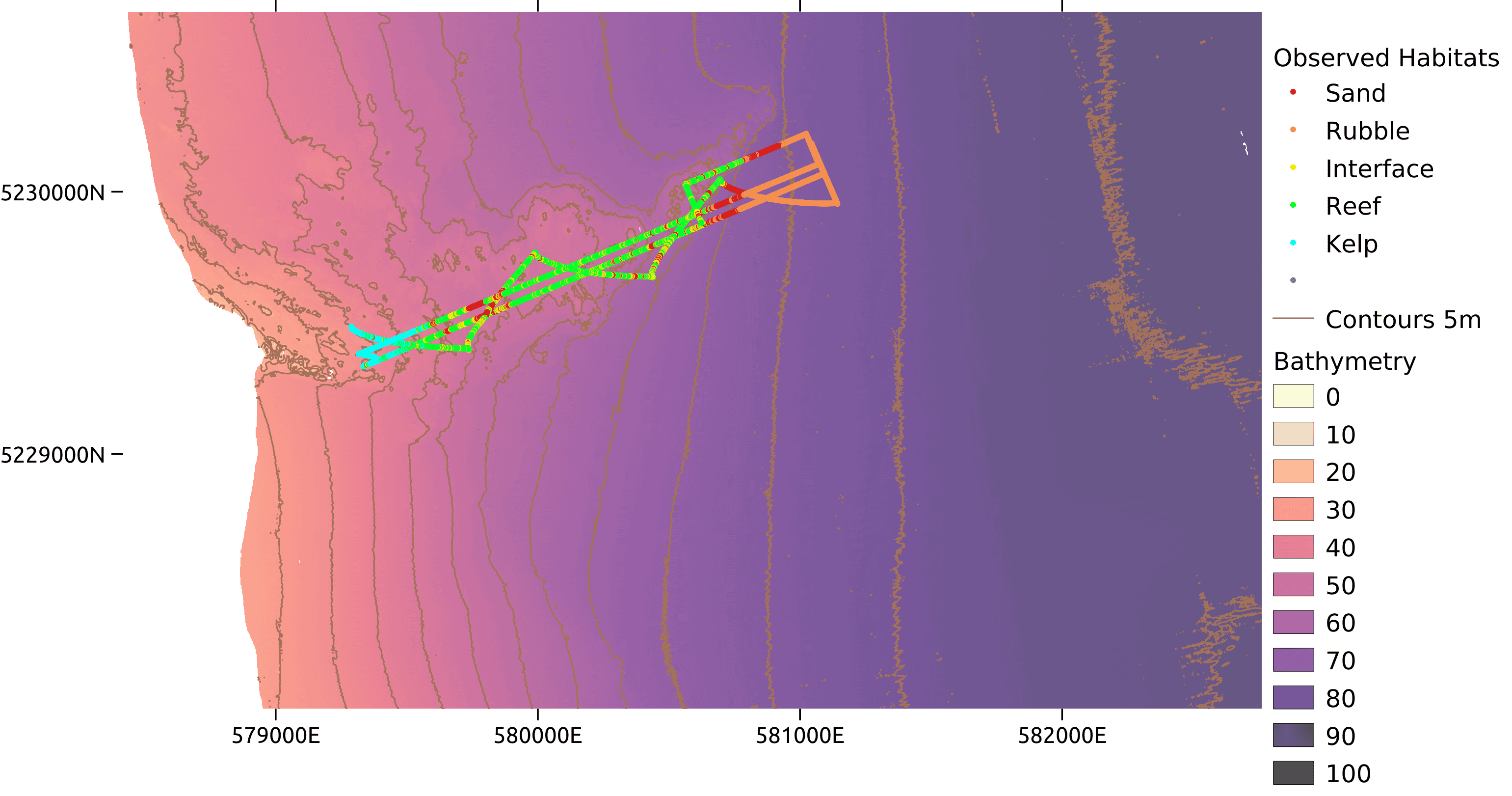}
        \caption{}
        \label{results:example:ohara_classes:bathy}
    \end{subfigure}
    ~ 
    \begin{subfigure}[t]{0.48\columnwidth}
        \centering
        \includegraphics[width=\textwidth]{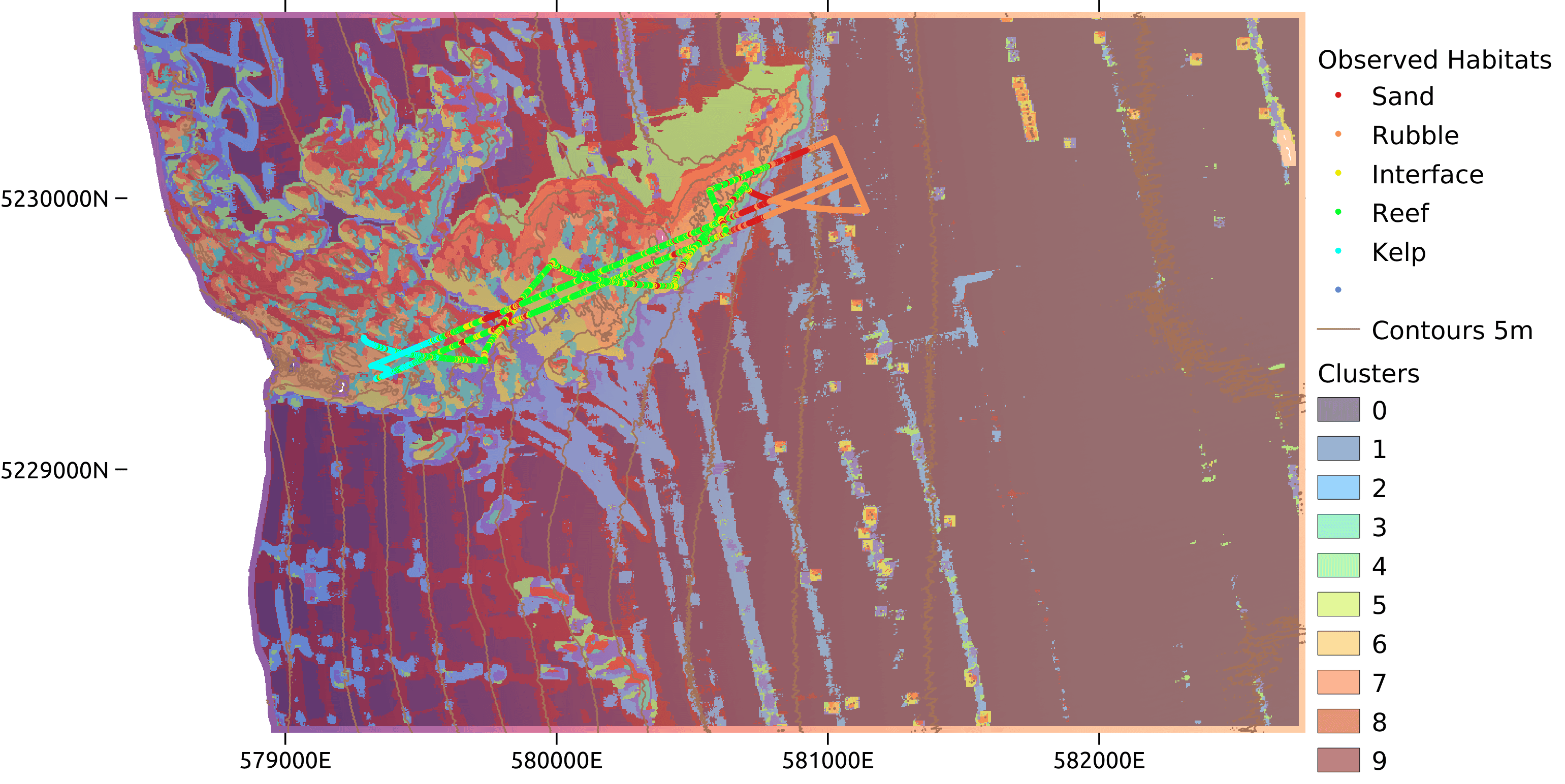}
        \caption{}
        \label{results:example:ohara_classes:clusters}
    \end{subfigure}
    \begin{subfigure}[t]{0.48\columnwidth}
        \centering
        \includegraphics[width=\textwidth]{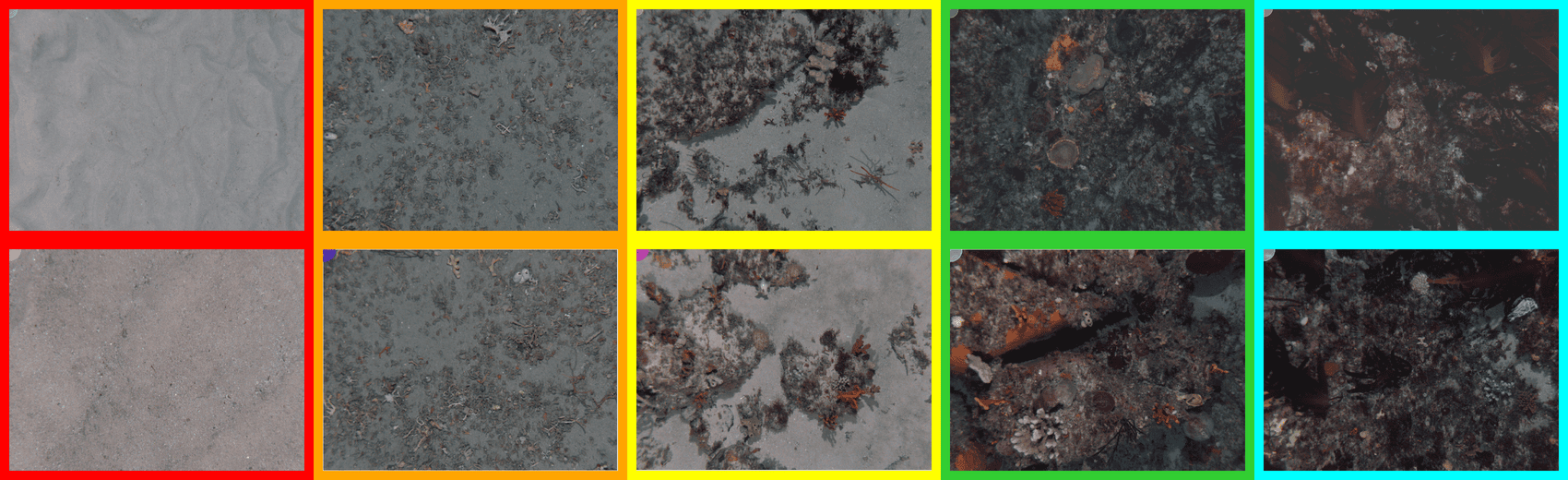}
        \caption{}
        \label{results:example:ohara_classes:images}
    \end{subfigure}
    
    \begin{subfigure}[t]{0.48\columnwidth}
        \centering
        \includegraphics[width=\textwidth]{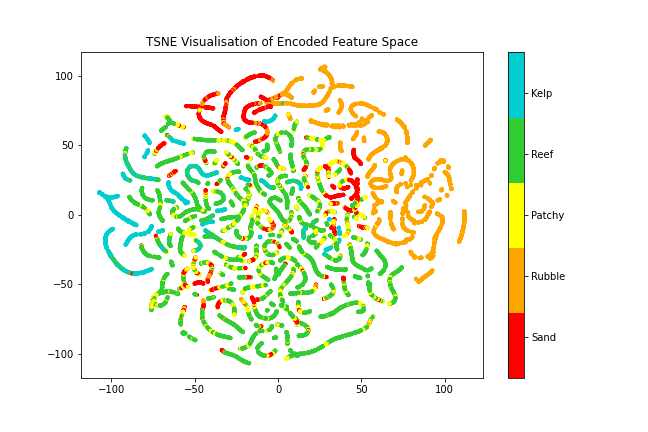}
        \caption{}
        \label{results:example:ohara_classes:tsne}
    \end{subfigure}
    ~
    \begin{subfigure}[t]{0.4\columnwidth}
        \centering
        \includegraphics[width=\textwidth]{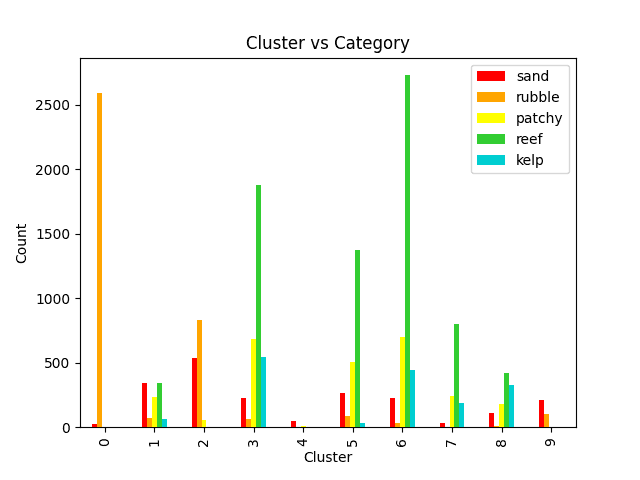}
        \caption{}
        \label{results:example:ohara_classes:catvscluster}
    \end{subfigure}
    \caption{Habitat labels from AUV imagery collected at O'Hara reef, Tasmania in 2008. The colours on the map correlate with the example images in (c). This shows the distribution of benthic habitats in the area. (d) displays a 2D t-SNE \citep{Maaten2008} projection of the bathymetric feature space, color coded with the habitat labels, which shows that similar areas of bathymetric terrain are likely to be similar benthic habitats. (e) shows the counts of each habitat class within each bathymetric cluster. Coordinates are displayed in UTM zone 55S.}
    \label{results:example:ohara_classes}
\end{figure}

\begin{figure}[!ht]
    \centering
    \begin{subfigure}[t]{0.48\columnwidth}
        \centering
        \includegraphics[width=\textwidth]{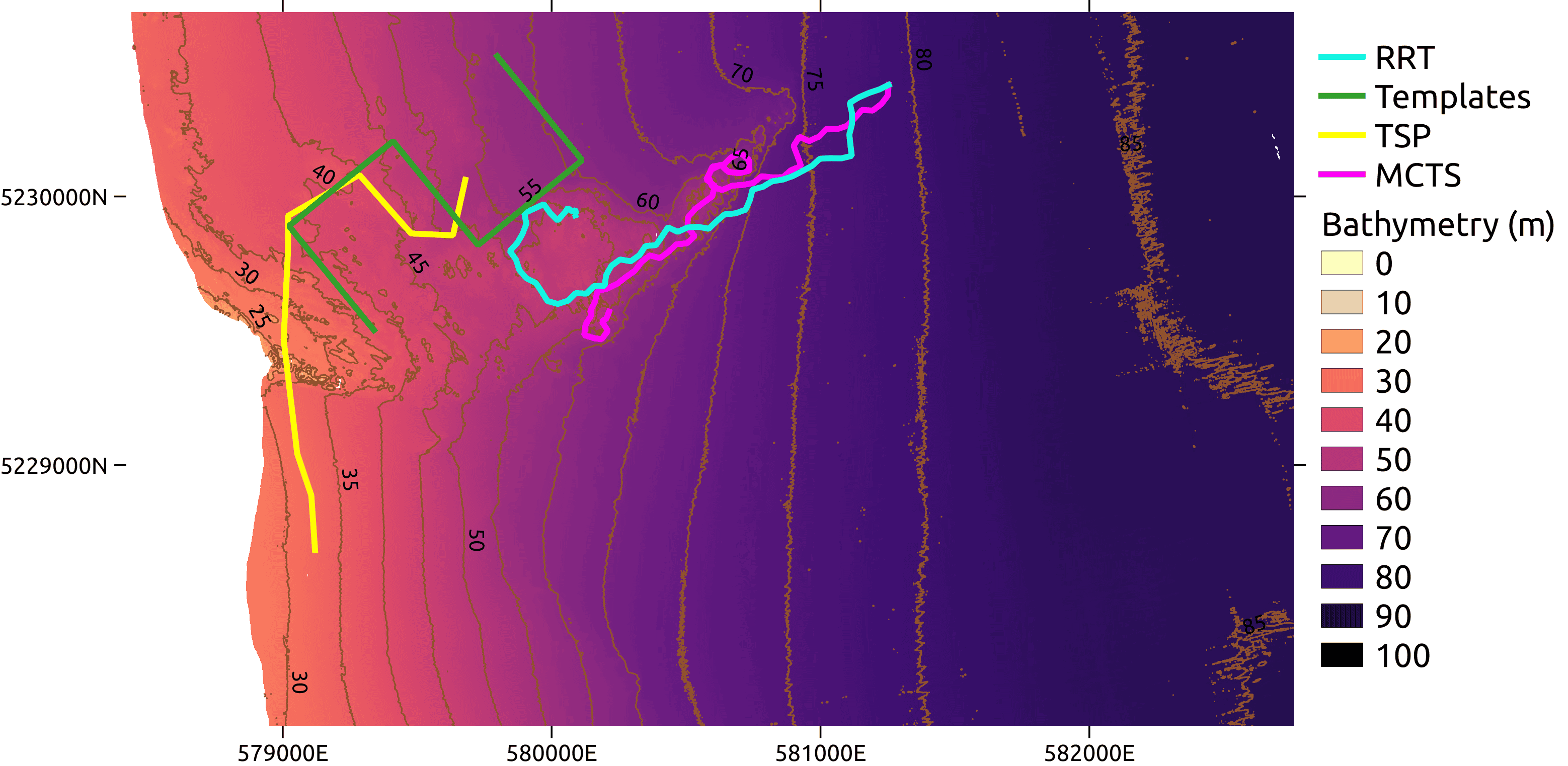}
        \caption{}
        \label{results:ohara:combined:bathy}
    \end{subfigure}
    ~ 
    \begin{subfigure}[t]{0.48\columnwidth}
        \centering
        \includegraphics[width=\textwidth]{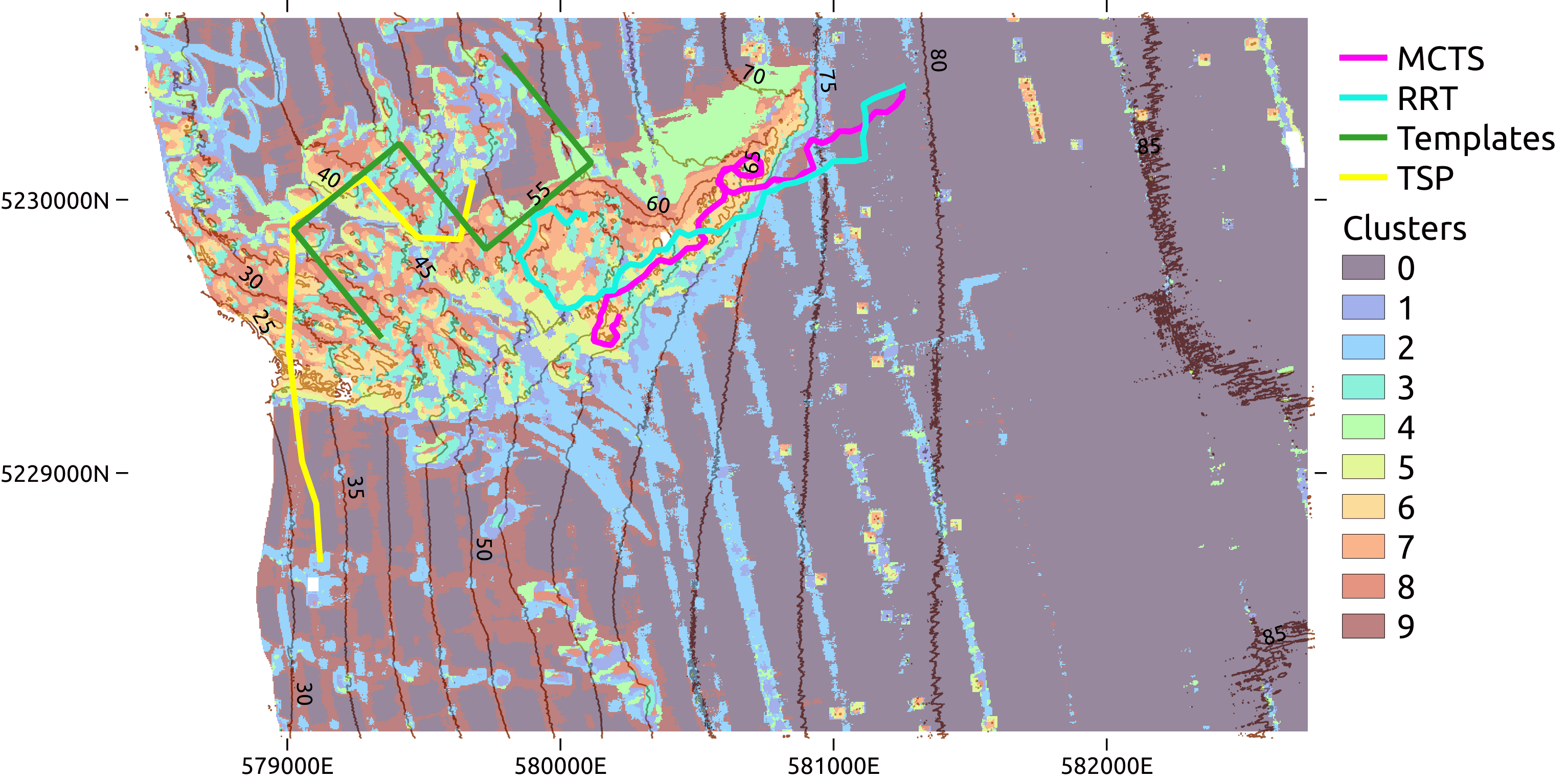}
        \caption{}
        \label{results:ohara:combined:cluster}
    \end{subfigure}
    \caption{Planned paths using encoded features for O'Hara Reef, Fortesque, Tasmania, Australia overlaid onto the bathymetry (a) and clusters (b). The budget for each path is 2500m.  The informed paths are able to collect features representative of the entire area by focusing on the areas around the O'Hara Reef. Coordinates are displayed in UTM zone 55S.}
    \label{results:ohara:combined}
\end{figure}

\begin{figure}[!ht]
    \centering
    \begin{subfigure}[t]{0.48\columnwidth}
        \centering
        \includegraphics[width=\textwidth]{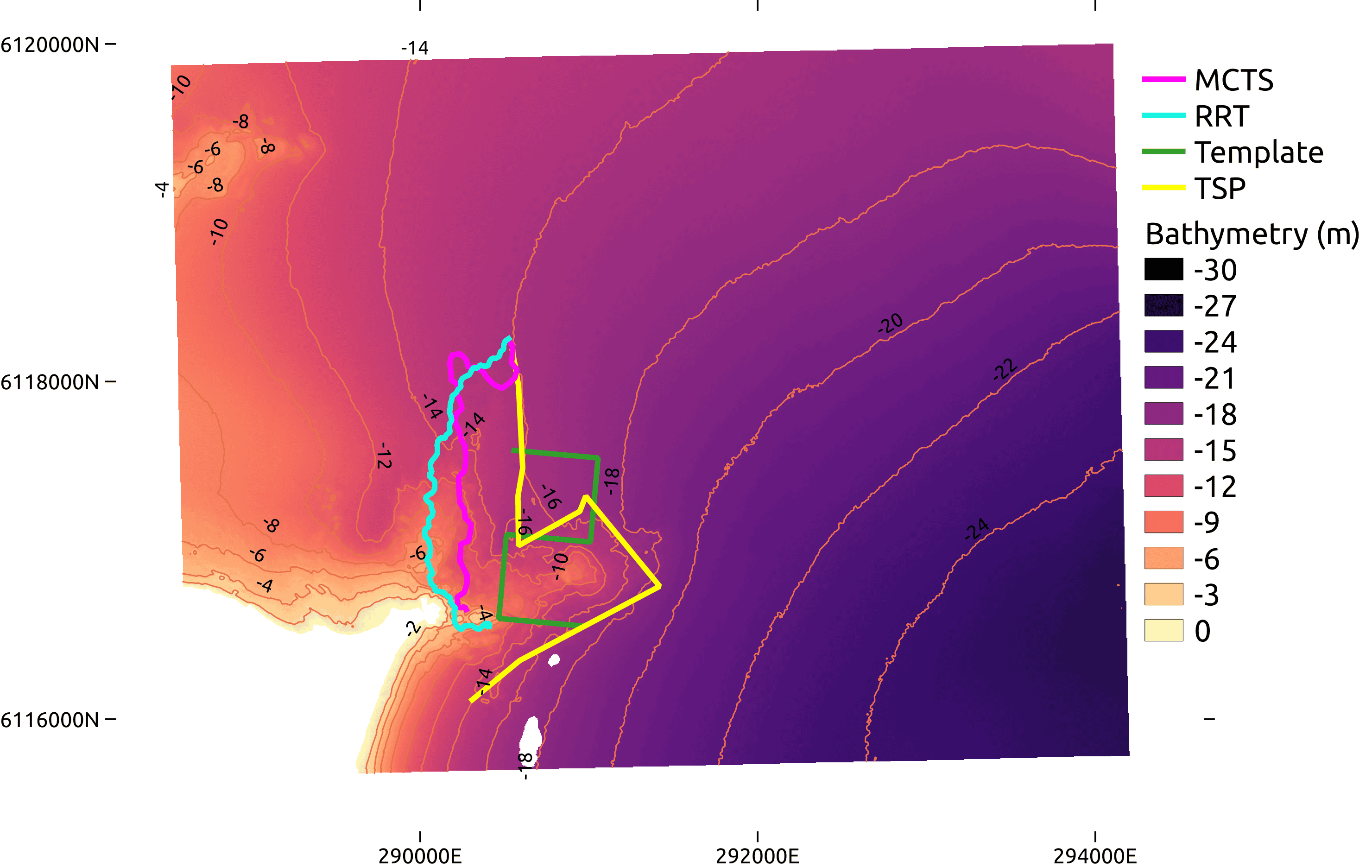}
        \caption{}
        \label{results:vincentia:combined:bathy}
    \end{subfigure}
    ~ 
    \begin{subfigure}[t]{0.48\columnwidth}
        \centering
        \includegraphics[width=\textwidth]{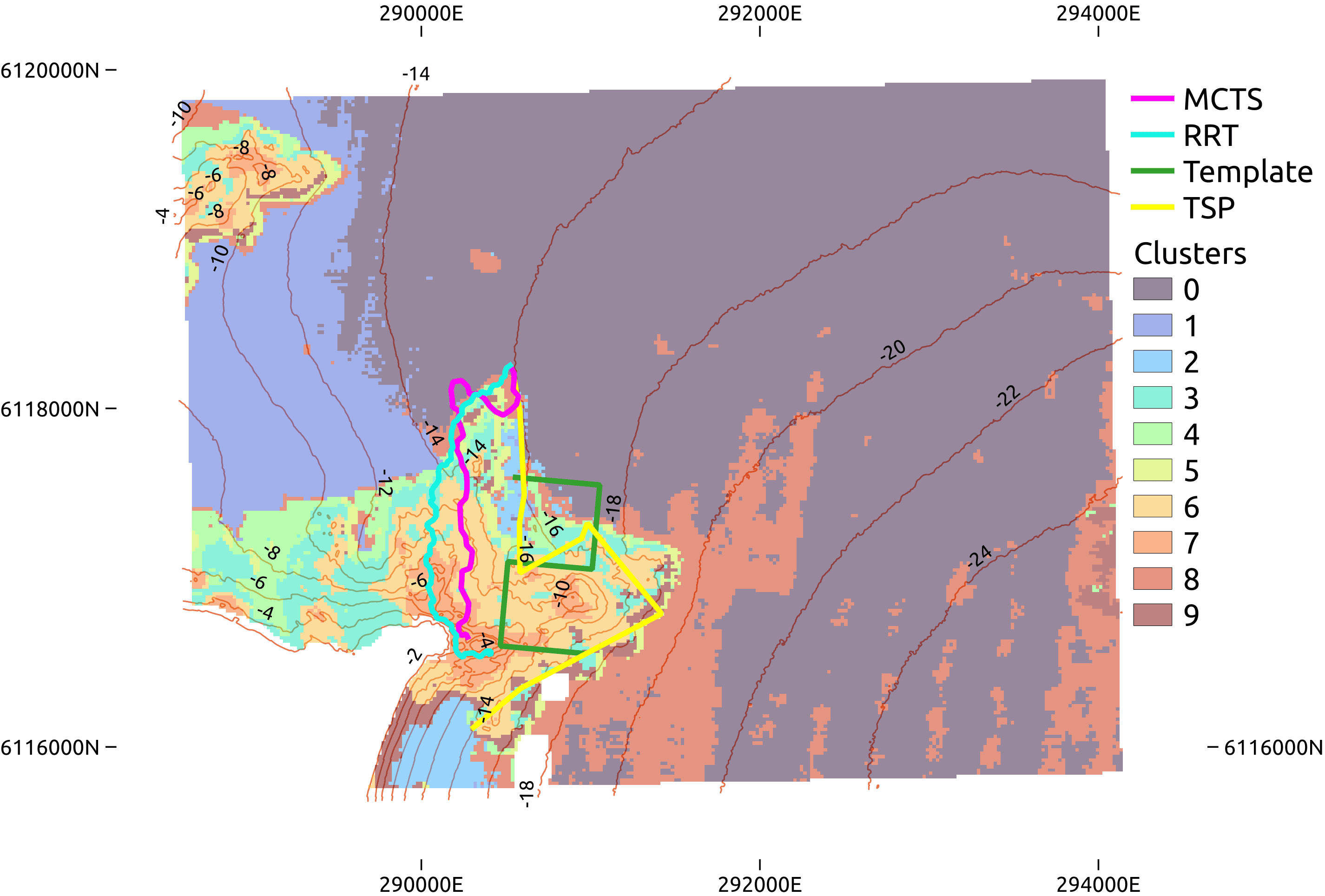}
        \caption{}
        \label{results:vincentia:combined:cluster}
    \end{subfigure}
    \caption{Planned paths using encoded features for Vincentia, Jervis Bay, NSW, Australia overlaid onto the bathymetry (a) and clusters (b). The budget for each path is 2500m. Coordinates are displayed in UTM zone 56S.}
    \label{results:vincentia:combined}
\end{figure}

\begin{figure}[!ht]
    \centering
    \begin{subfigure}[t]{0.48\columnwidth}
        \centering
        \includegraphics[width=\textwidth]{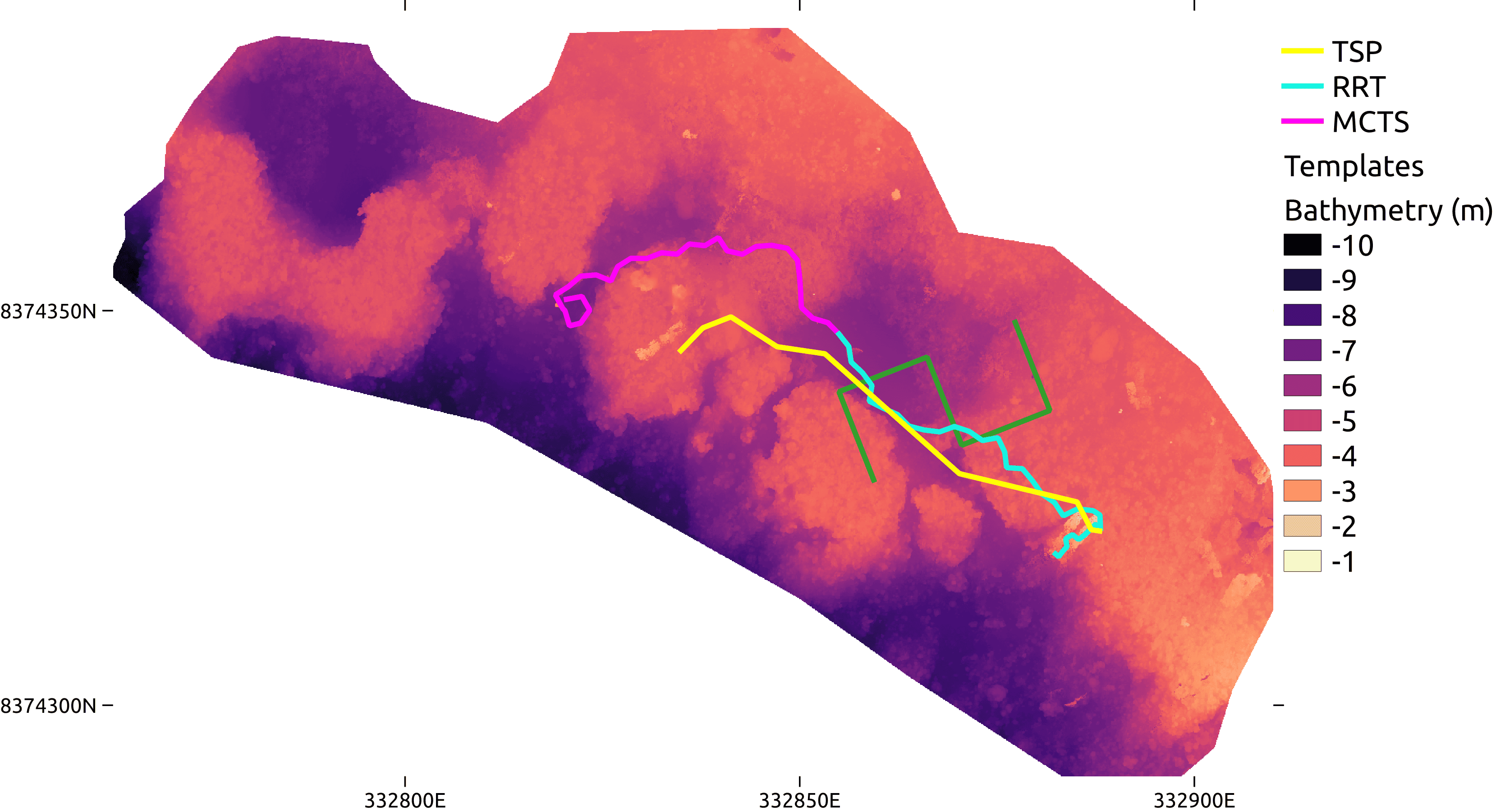}
        \caption{}
        \label{results:trimodal:combined:bathy}
    \end{subfigure}
    ~ 
    \begin{subfigure}[t]{0.48\columnwidth}
        \centering
        \includegraphics[width=\textwidth]{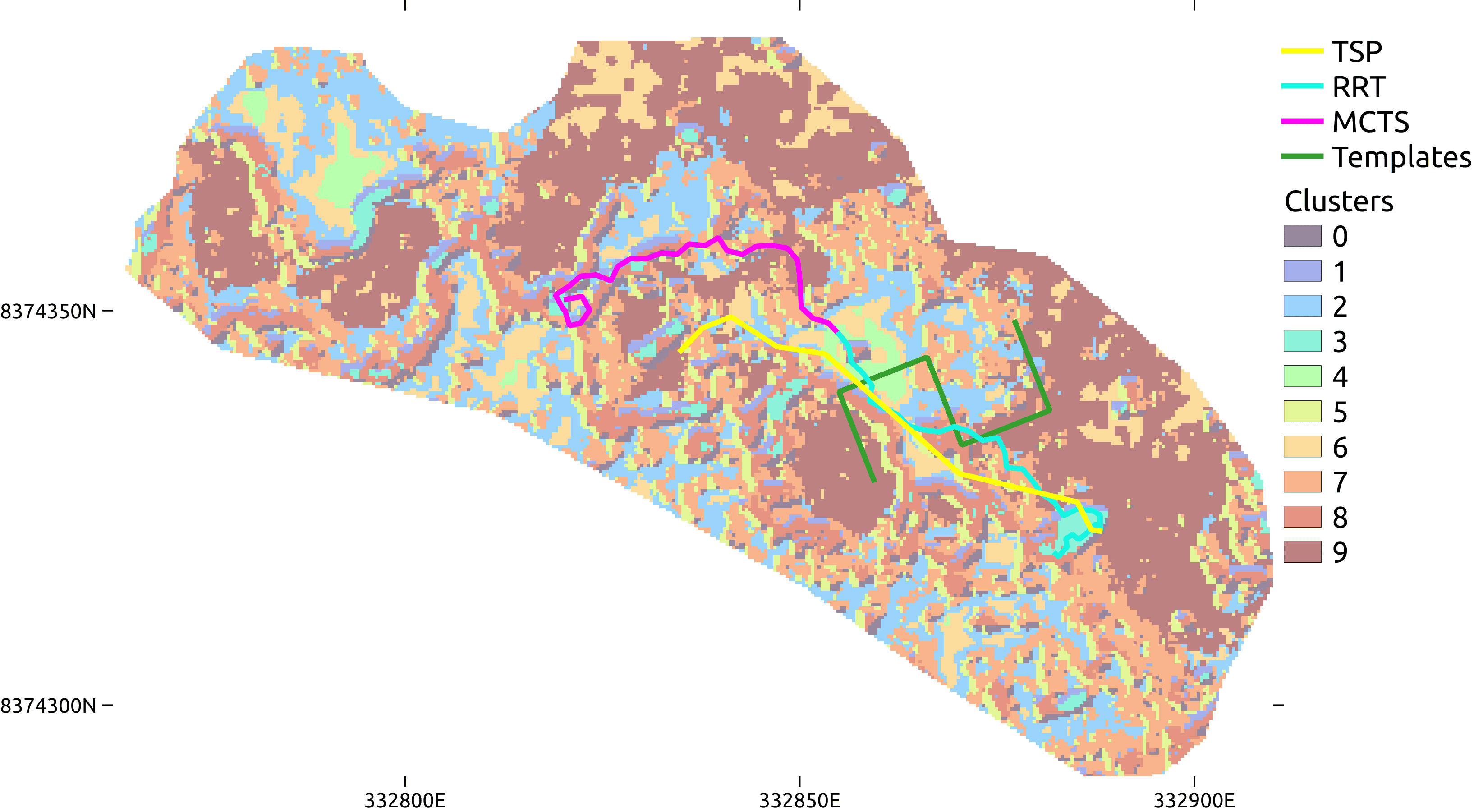}
        \caption{}
        \label{results:trimodal:combined:cluster}
    \end{subfigure}
    ~ 
    \begin{subfigure}[t]{0.48\columnwidth}
        \centering
        \includegraphics[width=\textwidth]{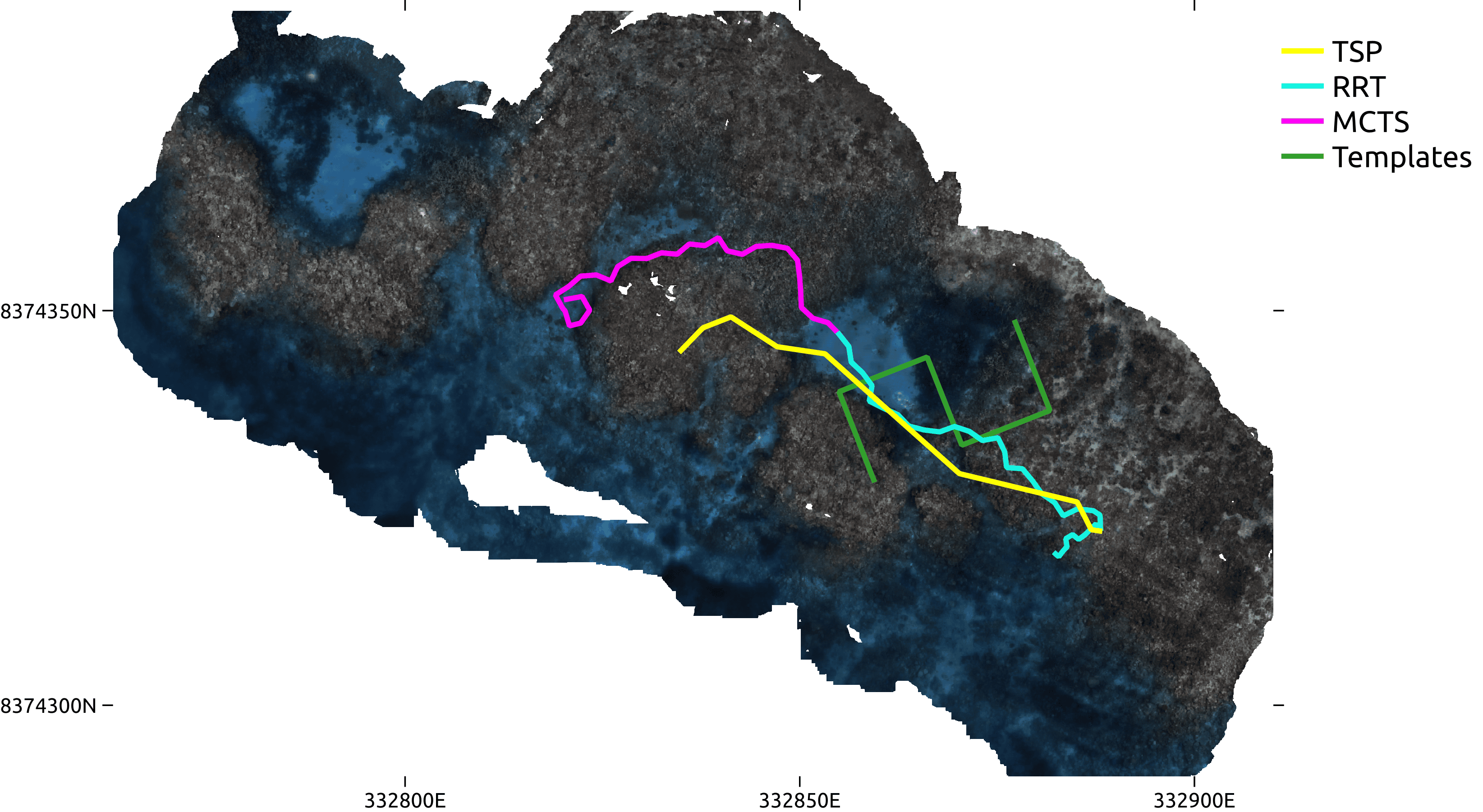}
        \caption{}
        \label{results:trimodal:combined:mosaic}
    \end{subfigure}
    ~ 
    \begin{subfigure}[t]{0.48\columnwidth}
        \centering
        \includegraphics[width=\textwidth]{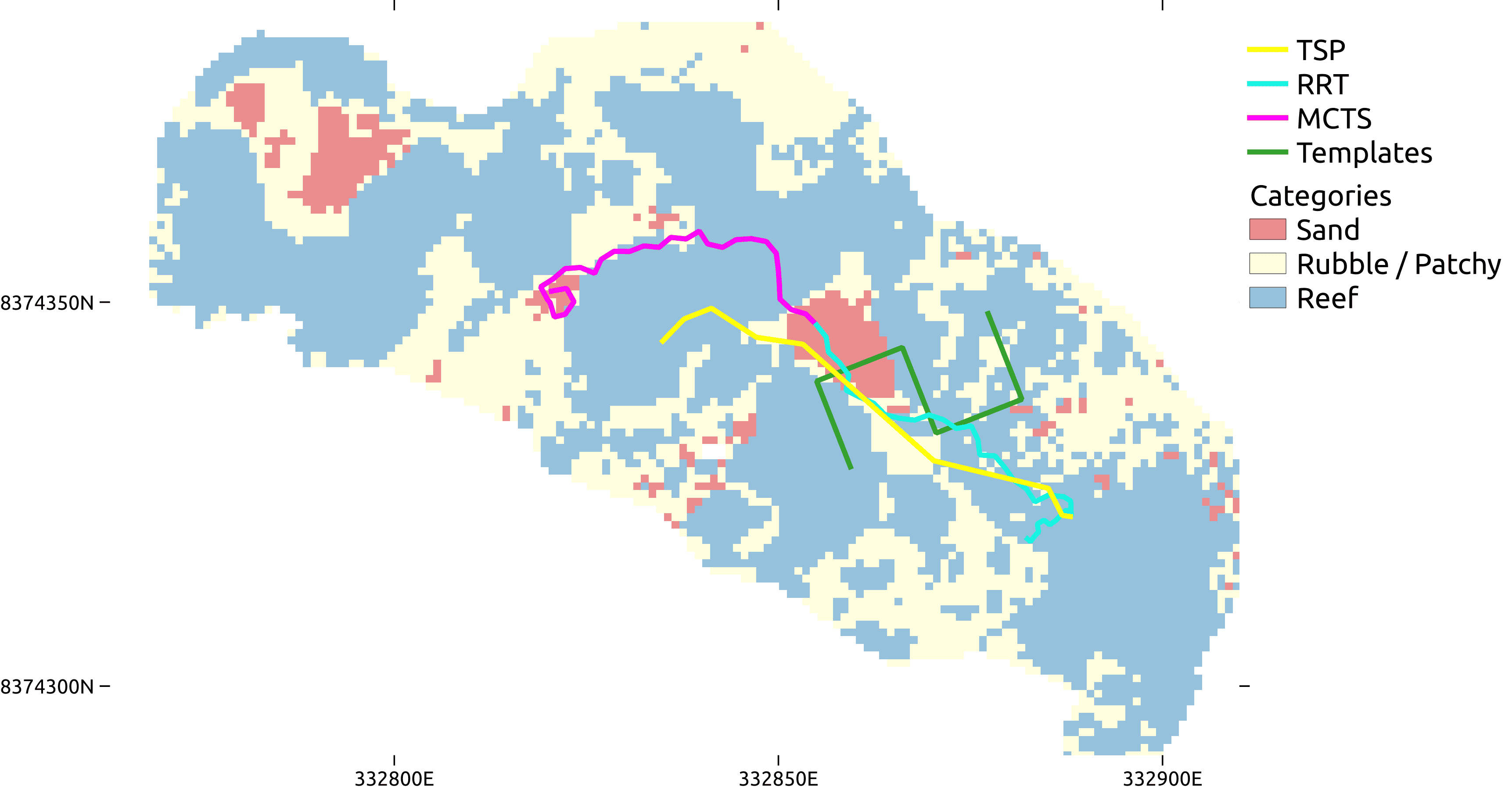}
        \caption{}
        \label{results:trimodal:combined:categories}
    \end{subfigure}
    \caption{Planned paths using encoded features for Trimodal Reef, Lizard Island, Queensland, Australia overlaid onto the bathymetry (a), clusters (b), image mosaic (c) and habitat labels (d). The budget for each path is 60m.  The habitat labels are sand (red), rubble / patchy reef (yellow) and reef (blue). Coordinates are displayed in UTM zone 55S.}
    \label{results:trimodal:combined}
\end{figure}

% Overview of datasets
This section presents results for the proposed feature exploration framework. It is demonstrated on three datasets, all with unique sources of bathymetry and scales, highlighting the adaptability of this approach. The \textit{O'Hara} dataset (Figure \ref{results:ohara:combined}) focuses on O'Hara Reef near Fortesque in Tasmania, Australia. It consists of ship-borne bathymetry gridded at 2m collected by GeoScience Australia \citep{Spinoccia2011}.  Figure \ref{results:example:ohara_classes} shows AUV imagery collected in this region in 2008 and the habitat classes overlaid onto the bathymetry and clusters. This provides context into the likely habitats found in this region. A well planned survey should visit all these habitat classes, or at least as many as can be differentiated using the bathymetry. The \textit{Vincentia} dataset (Figure \ref{results:vincentia:combined}) is focused on the area off Vincentia, Jervis Bay, Australia, with the bathymetry derived from LiDAR and gridded at 5m \citep{NSWMarineLidar2018}. For the \textit{Trimodal} dataset (Figure \ref{results:trimodal:combined}), an AUV was used to densely collect benthic imagery of Trimodal Reef, Lizard Island, Australia. The bathymetry is extracted from the photogrammetry at 0.1m. Furthermore, three classes of benthic habitats; sand, rubble and reef, have been coarsely labelled on the mosaic, to demonstrate how the proposed planners visit all the habitat classes.

% Benchmark method
To benchmark these proposed planners, random transects are placed across the environment, with the length of these transects being equal to the budget set for the planners. The number of transects is set to 100 to ensure the environment is adequately sampled. The benchmark, $L_B$ is the mean value of each of the validation criteria across all the transects. This benchmark is expected to perform well on the datasets selected, as the bathymetry is cropped to the area of interest so that any transect placed across the environment is likely to visit an interesting area.

% Evaluation criteria
Each path is evaluated using the criteria established in Section \ref{sec:info:eval}. $M_{PD}$ captures the diversity of features of the path, which should be maximised. $M_K$ uses features randomly sampled from the environment to measure how well the path characterises the environment. It summarises the feature space using clustering to prevent bias towards large spatial regions. Finally, $M_C$ measures the ratio of bathymetric cluster types visited, compared to the total number of cluster types:
\begin{equation}
    M_C = \frac{\text{Clusters Visited}}{\text{Total Clusters}}
    \label{eq:info:eval_mc}
\end{equation}

% Budget
The budget for each dataset is set according to an initial pass by the Cluster-TSP method. This is used to set the budget as we regard the feature-space as sufficiently explored when all the clusters have been visited. The budget is set at 2500m for the \textit{O'Hara} dataset,  2500m for the \textit{Vincentia} dataset and 60m for the \textit{Trimodal} dataset, given its smaller size.

% Parameters
The parameters used for each planner were empirically selected to achieve the best performance. For each run, the \textit{MCTS} and \textit{RRT} planners start three times, each for 10,000 iterations, with the best path automatically selected. The multiple starts avoids the planners proposing a poor path. The expansion distance for these methods was set relative to the resolution of the bathymetry and to the overall budget. For the \textit{Templates} method, 1000 candidate paths were evaluated.

We present results for planning using the geometric features in Table \ref{table:results:geometric} and for encoded features in Table \ref{table:results:encoded}. As each method is stochastic, each experiment is run ten times. The starting positions for the \textit{MCTS} and \textit{RRT} are kept the same for each run, and are selected using the method for starting in informative regions, outlined in Section \ref{sec:info:rrt}. 
 
The survey planners that utilise the remotely-sensed data (\textit{MCTS}, \textit{RRT}, \textit{TSP}, \textit{Templates}) significantly outperform random transects over the area of interest ($L_B$). As expected, $L_B$ performs relatively well as the operational area is cropped around the areas of interest. As the operational area becomes larger, randomly placing transects ($L_B$) is expected to perform poorly as a random transect will be less likely to intersect interesting areas of the feature space. 

The ability for the set of planners to propose informative paths for both the geometric and encoded feature representations highlights the generalisation of the approach to different feature representations. Both feature representations direct the AUV to similar informative areas; relatively flat areas that are likely to be sand, more rugose areas that are commonly reef, and interface areas that exist between the habitat types. This is expected as both the representations are capturing the geometric features of the bathymetry, albeit at different resolutions. However the performance of this approach for higher-dimensional feature representations may suffer as it relies upon the feature  distance, that is less reliable in higher dimensions due to sparsity. Similar samples can still have large feature distances between them which may not represent large underlying differences in benthic habitat. This can be observed in the paths planned with the encoded features, which focus too much on the highly rugose areas of the reef, where there can be large feature distances between similar and co-located samples. This leads to oversampling of these areas at the expense of others areas.  Therefore, care should be taken to construct a feature space that is both compact and descriptive in representing the important features of the underlying data. This may limit this approach from being used for planning informative surveys using highly-dimensional remote data such as satellite imagery.

Both the freeform planners, \textit{MCTS} and \textit{RRT} are able to comprehensively explore the feature space. These planners choose where next to sample in order to maximise the distance from previously collected samples, which effectively maximises the metric $M_{PD}$. This does not directly optimise approximating the feature space of the target area. It is too computationally intensive to directly optimise the $M_{K}$ method for these planners. However when planning with geometric features, the strong performance on the $M_{K}$ metric indicates that the approach of selecting samples that maximise the distance from existing samples leads to a thorough coverage of the feature space. The performance on the $M_K$ metric is lessened when planning with encoded features, due to the higher dimensions of the encoded feature space. This increased sparsity means that moderately different samples can have large feature distances between them, and the freeform planners will focus sampling on these areas. On the \textit{O'Hara} and \textit{Trimodal} datasets, the freeform planners consistently visit most or all clusters present in the dataset with $M_C$ over $0.96$ (an average 9.6 out of 10 clusters visited on each run), which is a key indication of an informative initial survey. Example paths can be seen in Figures \ref{results:ohara:combined}, \ref{results:vincentia:combined}, \ref{results:trimodal:combined}.

% Templates - in depth
The \textit{Templates} method, which involves placing a set survey template using the information metrics, scores highly across all the datasets presented, highlighting the benefits of utilising the bathymetry for positioning the survey template. For these experiments, we consider a 2D broad-grid that is shown to be effective using the guidelines developed in \citet{Foster2014}. In practice,  using a set survey template can limit performance, if the template cannot be aligned to interesting features in the environment. For example a square-grid would not be well suited to the long reef in the \textit{O'Hara} dataset. Evaluating multiple candidate surveys could improve results, providing a greater chance a template will be well matched to the target area. The current method is a brute-force method, where many randomly-placed candidate surveys are evaluated. Dynamically fitting survey templates to informative points could lead to surveys that provide more complete coverage of the feature space. Similarly, a secondary local optimisation of the survey template position could further increase the information collected.

%The freeform planning methods perform strongly across all three datasets, demonstrating the benefits of using the proposed information metrics in incremental information gathering planners. The methods perform similarly on the \textit{O'Hara} and \textit{Trimodal} datasets, with similar scores for $M_{K}$ and $M_{C}$ metrics. On the \textit{Vincentia} dataset, \textit{RRT} fails to consistently visit the range of bathymetric clusters, on average visiting 6.7 out of 10 clusters. The selection of a starting position has a major influence on the subsequent performance of the planned paths, hence the starting positions were placed in informative regions using the method outlined in Section \textcolor{red}{SEC STARTING} and the same starting positions were used for each run of both the \textit{RRT} and \textit{MCTS} methods. 

% TSP - in depth
The \textit{TSP} method is designed to visit all of the clusters, hence scoring highly on the $M_C$ metric. This method also performs well on the $M_{K}$ metric, however the feature distance methods (\textit{RRT},\textit{MCTS},\textit{Templates}) generally score better, highlighting the importance of exploring the feature space within a cluster. \textit{TSP} is an effective method for initial feature space exploration, however its reliance on clustering and lack of flexibility hinder its ability for planning benthic surveys. For some datasets, clustering can lead to suboptimal clusters, with clusters forming that are either too large or small, which will impact the ability of the planner to explore the latent space. Artefacts in the bathymetry, such as the track lines present in the \textit{O'Hara} dataset (Figure \ref{results:ohara:combined:cluster}), can lead to clusters that are not reflective of the habitat. Finally, the \textit{TSP} method does not have a pathway to continue planning beyond the initial survey, while the other methods can continue planning indefinitely.

\definecolor{LightBlue}{RGB}{230, 242, 255}
\definecolor{LightRed}{RGB}{255, 214, 204}
\definecolor{LightYellow}{RGB}{255, 255, 230}
\definecolor{LightOrange}{RGB}{255, 214, 157}
\definecolor{LightRed}{RGB}{255, 185, 185}

\newcolumntype{i}{>{\columncolor{LightBlue}}c}
\newcolumntype{j}{>{\columncolor{LightYellow}}c}
\newcolumntype{k}{>{\columncolor{LightOrange}}c}
\newcolumntype{r}{>{\columncolor{LightRed}}c}

\begin{table}[!ht]
\centering
\begin{tabular}{|p{0.08\textwidth}|c|| i i i |j|k|}%
\hline%
&&MCTS&RRT&Templates&TSP*&$L_B$\\%
\hline%
\multirow{6}{*}{O'Hara}&$M_{PD}$&$\mathbf{2.04 \pm 0.34}$&$1.75 \pm 0.26$&$1.54 \pm 0.18$&$1.54 \pm 0.17$&$1.23 \pm 0.02$\\%
&$M_K$&$1.20 \pm 0.00$&$\mathbf{1.19 \pm 0.00}$&$1.20 \pm 0.01$&$1.31 \pm 0.00$&$1.53 \pm 0.00$\\%
&$M_C$*&$0.96 \pm 0.10$&$\mathbf{0.98 \pm 0.06}$&$0.97 \pm 0.05$&$0.97 \pm 0.07$&$0.73 \pm 0.01$\\%
&D (m)&$2481 \pm 12.07$&$2418 \pm 150.57$&$2500 \pm 0.00$&$1881 \pm 361.22$&$2500 \pm 0.00$\\%
\hline%
\multirow{6}{*}{Vincentia}&$M_{PD}$&$1.57 \pm 0.46$&$\mathbf{2.25 \pm 0.27}$&$1.55 \pm 0.44$&$1.34 \pm 0.25$&$0.86 \pm 0.04$\\%
&$M_K$&$1.31 \pm 0.01$&$\mathbf{1.31 \pm 0.01}$&$1.34 \pm 0.01$&$1.32 \pm 0.02$&$1.72 \pm 0.00$\\%
&$M_C$*&$0.87 \pm 0.09$&$0.79 \pm 0.07$&$0.79 \pm 0.12$&$\mathbf{0.99 \pm 0.03}$&$0.38 \pm 0.03$\\%
&D (m)&$2485 \pm 13.00$&$2309 \pm 278.60$&$2500 \pm 0.01$&$3150 \pm 730.10$&$2500 \pm 0.00$\\%
\hline%
\multirow{6}{*}{Trimodal}&$M_{PD}$&$1.24 \pm 0.08$&$\mathbf{1.61 \pm 0.26}$&$1.40 \pm 0.15$&$1.40 \pm 0.12$&$1.20 \pm 0.02$\\%
&$M_K$&$\mathbf{0.55 \pm 0.00}$&$0.61 \pm 0.00$&$0.56 \pm 0.00$&$0.62 \pm 0.00$&$0.61 \pm 0.00$\\%
&$M_C$*&$0.90 \pm 0.08$&$0.98 \pm 0.04$&$0.90 \pm 0.05$&$\mathbf{0.99 \pm 0.03}$&$0.85 \pm 0.01$\\%
&D (m)&$59 \pm 0.43$&$48 \pm 9.12$&$60 \pm 0.00$&$51 \pm 7.71$&$60 \pm 0.00$\\%
\hline%
\end{tabular}
\caption{Results for exploring the feature space, where the feature space is composed of the geometric features (Section \ref{sec:feature_extraction:geometric}). Each value is the mean value over ten runs. The blue columns indicate paths planned using the information metric proposed in Section \ref{sec:infoplan:reward}, which are compared to Cluster-TSP in the yellow column (Section \ref{sec:info:cluster_tsp}) and random transects over the area ($L_B$) in the orange column. $M_{PD}$ and $M_C$ should be maximised, while $M_K$ should be maximised. As the clustering process does not produce usable clusters with geometric features, * indicates methods and metrics that use encoded features.}
\label{table:results:geometric}
\end{table}

\begin{table}[!ht]
\centering
\begin{tabular}{|p{0.08\textwidth}|c|| i i i |j|k|}%
\hline%
&&MCTS&RRT&Templates&TSP&$L_B$\\%
\hline%
\multirow{6}{*}{O'Hara}&$M_{PD}$&$13.80 \pm 1.02$&$\mathbf{14.70 \pm 3.15}$&$8.60 \pm 1.46$&$9.56 \pm 1.19$&$3.87 \pm 0.20$\\%
&$M_K$&$\mathbf{8.96 \pm 0.24}$&$9.06 \pm 0.05$&$8.97 \pm 0.03$&$9.26 \pm 0.03$&$9.36 \pm 0.00$\\%
&$M_C$&$0.96 \pm 0.07$&$0.97 \pm 0.05$&$\mathbf{0.99 \pm 0.03}$&$0.97 \pm 0.07$&$0.72 \pm 0.02$\\%
&D (m)&$2492 \pm 11.92$&$2311 \pm 247.45$&$2500 \pm 0.00$&$1881 \pm 361.22$&$2500 \pm 0.00$\\%
\hline%
\multirow{6}{*}{Vincentia}&$M_{PD}$&$16.23 \pm 5.71$&$\mathbf{18.00 \pm 3.20}$&$11.02 \pm 2.53$&$8.49 \pm 2.17$&$2.87 \pm 0.25$\\%
&$M_K$&$9.07 \pm 0.01$&$8.96 \pm 0.09$&$\mathbf{8.68 \pm 0.05}$&$8.78 \pm 0.02$&$9.14 \pm 0.00$\\%
&$M_C$&$0.80 \pm 0.09$&$0.76 \pm 0.08$&$0.82 \pm 0.06$&$\mathbf{0.99 \pm 0.03}$&$0.38 \pm 0.02$\\%
&D (m)&$2493 \pm 11.59$&$2401 \pm 174.95$&$2499 \pm 0.00$&$3150 \pm 730.10$&$2500 \pm 0.00$\\%
\hline%
\multirow{6}{*}{Trimodal}&$M_{PD}$&$5.45 \pm 0.83$&$\mathbf{5.99 \pm 0.96}$&$3.98 \pm 0.81$&$5.14 \pm 0.46$&$4.07 \pm 0.04$\\%
&$M_K$&$2.98 \pm 0.02$&$2.93 \pm 0.03$&$\mathbf{2.90 \pm 0.01}$&$3.09 \pm 0.06$&$3.20 \pm 0.00$\\%
&$M_C$&$0.98 \pm 0.04$&$\mathbf{0.99 \pm 0.03}$&$0.93 \pm 0.05$&$\mathbf{0.99 \pm 0.03}$&$0.86 \pm 0.01$\\%
&D (m)&$59 \pm 0.54$&$54 \pm 4.75$&$60 \pm 0.00$&$50 \pm 7.71$&$60 \pm 0.00$\\%
\hline%
\end{tabular}
\caption{Results for exploring the feature space, where the feature space is latent space of an autoencoder trained on bathymetry patches (Section \ref{sec:feature_extraction:learnt}). $\pm$ refers to plus or minus one standard deviation. Each value is the mean value over ten runs. The blue columns indicate paths planned using the information metric proposed in Section \ref{sec:infoplan:reward}, which are compared to Cluster-TSP in the yellow column (Section \ref{sec:info:cluster_tsp}) and random transects over the area ($L_B$) in the orange column. As the encoded feature space has more dimensions than the geometric feature space (17 vs 5), the values of $M_{PD}$ and $M_K$ are larger than in Table \ref{table:results:geometric}.}
\label{table:results:encoded}
\end{table}

\begin{table}
\centering
\begin{tabular}{|c|c|| i i i |j|k|}%
\hline%
&&MCTS&RRT&Templates&TSP&$L_B$\\%
\hline%
\multirow{1}{*}{Geometric}&$M_H$&$0.97 \pm 0.10$&$\mathbf{1.00 \pm 0.00}$&$0.93 \pm 0.13$&$\mathbf{1.00 \pm 0.00}$&$0.89 \pm 0.02$\\%
\hline%
\multirow{1}{*}{Encoded}&$M_H$&$\mathbf{1.00 \pm 0.00}$&$0.97 \pm 0.10$&$0.93 \pm 0.13$&$\mathbf{1.00 \pm 0.00}$&$0.89 \pm 0.02$\\%
\hline%
\end{tabular}
\caption{The average habitat visits for the densely sampled \textit{Trimodal} dataset over ten runs. $\pm$ refers to plus or minus one standard deviation.  These results highlight how using informed planning leads to observing all the habitats. As this dataset is small compared to the budget, it restricts straight transects to a smaller area making it likely to hit most of the classes.}
\end{table}

Noise or artefacts in the remotely-sensed data can lead to incorrect estimates of the seafloor structure. For acoustic bathymetry, these artefacts are either from sensor noise or errors in the localisation of the sensor. This is evident in the bathymetry for the Fortesque region and is highlighted in the clusters (Figure \ref{results:ohara:combined:cluster}), which shows the track lines that correspond to the vessels path when collecting the bathymetry. These artefacts can make the planner to unnecessarily traverse to these regions, however this was not evident in these experiments.

\FloatBarrier

\subsection{Computation Considerations}

The planning time is not considered an important factor when comparing these approaches, as the planning time is insignificant compared to the time taken to complete a survey. These planners take several minutes to plan a survey, whereas completing the survey will take several hours. Table \ref{table:time} shows the time taken to plan a survey on the O'Hara dataset, with all methods producing an informed survey plan within several minutes. All runs were performed on the O'Hara dataset, running on a 12-core Ryzen 3900X CPU at 3.8GHz coupled with a Nvidia 2080Ti GPU. Parameters, such as the expansion distance and number of reward samples, have an impact on the time taken to complete the survey and hence they are kept the same as those used to complete the results. The \textit{MCTS} and \textit{RRT} methods are capped at 10,000 iterations, whereas the \textit{Templates} method uses 1000 candidate surveys. The time budget for the \textit{TSP} method is set at 180.0 seconds. All these methods are \lq any time' planners and can instead be given a planning time budget. The time is reported for planning with geometric features. The time taken to plan with encoded features is the same, by pre-extracting the features at every point on the raster and creating a new geotiff with these features.

The reward function uses the history of features in its calculation and hence will become slower as the number of samples increases, however this is not significant in testing. For the \textit{MCTS} method, performing a \lq heavy playout' \citep{Browne2012} by  increasing the number of simulations performed during each cycle significantly extends the execution time. Four iterations per cycle were empirically found to balance execution time and quality of the resulting survey. For the \textit{RRT} method, the parameter that most increases execution time is the number of nearest neighbours to evaluate during the \lq Select' process. In these experiments 50 nearest neighbours were used.

\begin{table}[!ht]
\centering
\begin{tabular}{|c | iii|j|}
\hline
     & MCTS & RRT & Templates & TSP \\
\hline
Time (s) & 164  & 116 & 454       & 180 \\
\hline
\end{tabular}
\caption{Time (in seconds) to plan one run for each method.}
\label{table:time}
\end{table}
\FloatBarrier

\subsection{Feature Visualisation}

\begin{figure}[!ht]
    \centering
    \begin{subfigure}[t]{0.48\columnwidth}
        \centering
        \includegraphics[width=\textwidth]{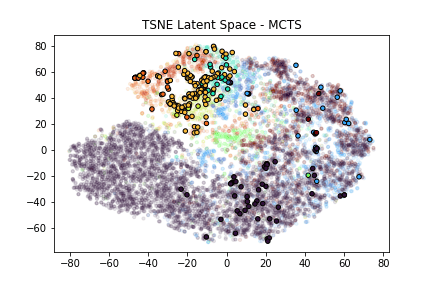}
        \caption{}
        \label{feature_vis:ohara:tsne:mcts}
    \end{subfigure}
    ~
    \begin{subfigure}[t]{0.48\columnwidth}
        \centering
        \includegraphics[width=\textwidth]{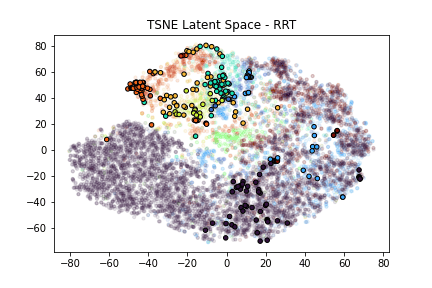}
        \caption{}
        \label{feature_vis:ohara:tsne:rrt}
    \end{subfigure}
    ~
        \begin{subfigure}[t]{0.48\columnwidth}
        \centering
        \includegraphics[width=\textwidth]{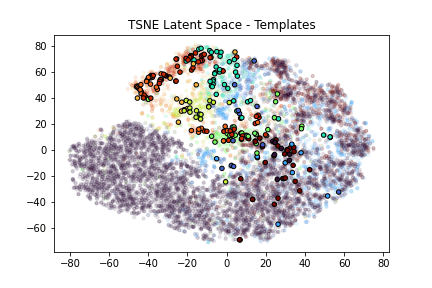}
        \caption{}
        \label{feature_vis:ohara:tsne:templates}
    \end{subfigure}
    ~
    \begin{subfigure}[t]{0.48\columnwidth}
        \centering
        \includegraphics[width=\textwidth]{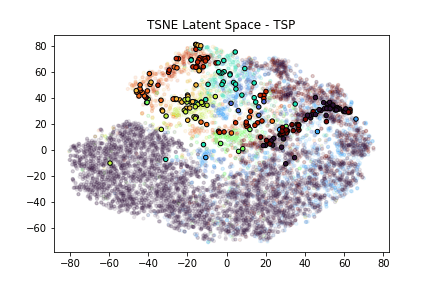}
        \caption{}
        \label{feature_vis:ohara:tsne:tsp}
    \end{subfigure}
    ~
    \caption{2D TSNE projection of the encoded feature space for the O'Hara region, and each path's respective exploration of this feature space. The partially transparent points are features randomly sampled from the area with the colours corresponding to the clusters outlined in Figure \ref{results:ohara:combined:cluster}. The opaque points with the black border represent the features extracted from the respective paths.}
    \label{feature_vis:ohara:tsne}
\end{figure}

To demonstrate the features collected along the paths, in Figure \ref{feature_vis:ohara:tsne} we show a 2D representation of the encoded feature space reduced using the t-SNE algorithm \citep{Maaten2008}. For this example, we use the O'Hara dataset and the example paths present in Figure \ref{results:ohara:combined}. The sampled features consist of 10,000 random samples from the area that are shaded transparently. Features collected on the planned paths are overlaid onto the sampled features. Although this 2D representation does not preserve feature distances, it shows the planned paths collecting features from all regions of the feature space.

\FloatBarrier
\begin{figure}[!ht]
    \centering
    \begin{subfigure}[t]{0.48\columnwidth}
        \centering
        \includegraphics[width=\textwidth]{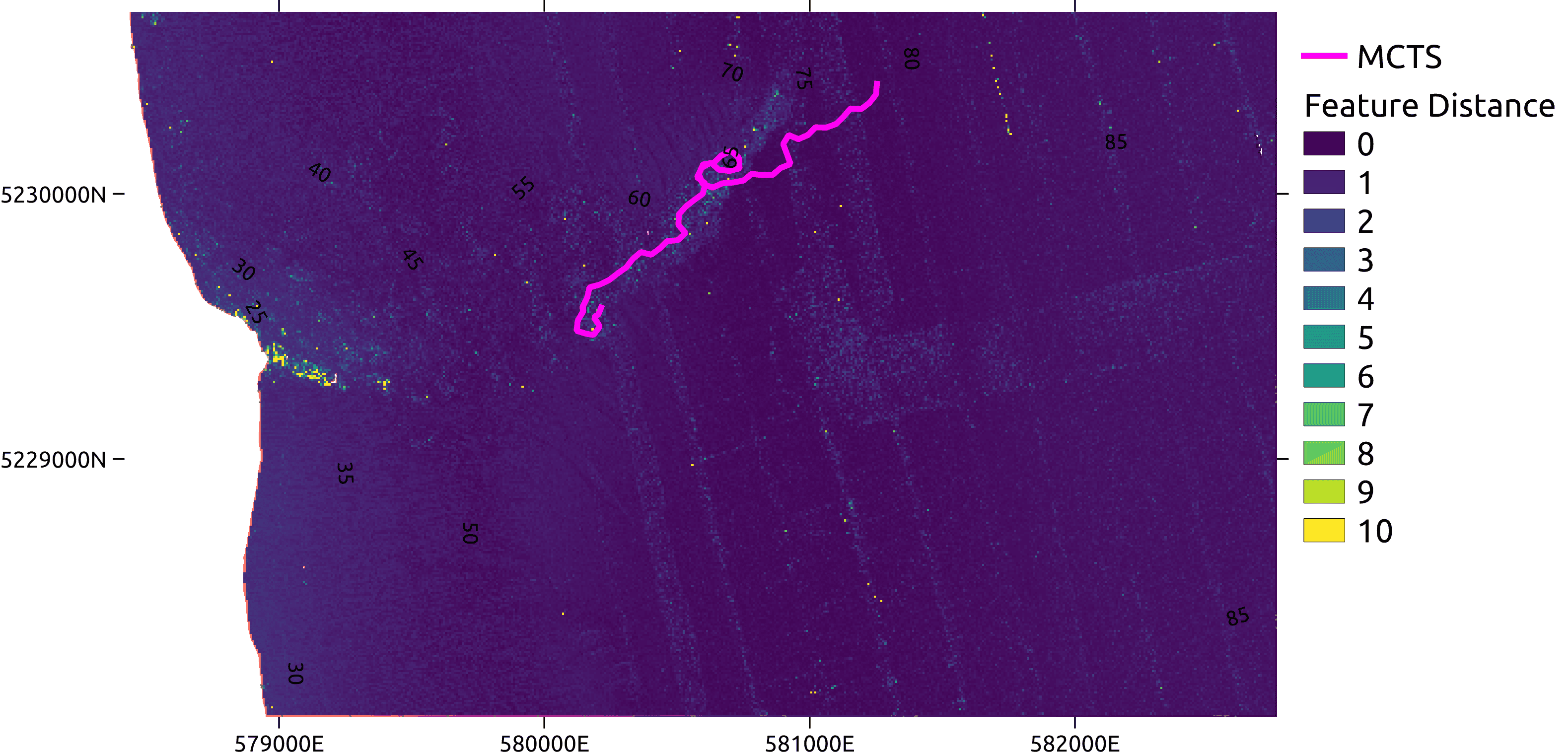}
        \caption{}
        \label{feature_dist:ohara:raw:mcts}
    \end{subfigure}
    ~ 
    \begin{subfigure}[t]{0.48\columnwidth}
        \centering
        \includegraphics[width=\textwidth]{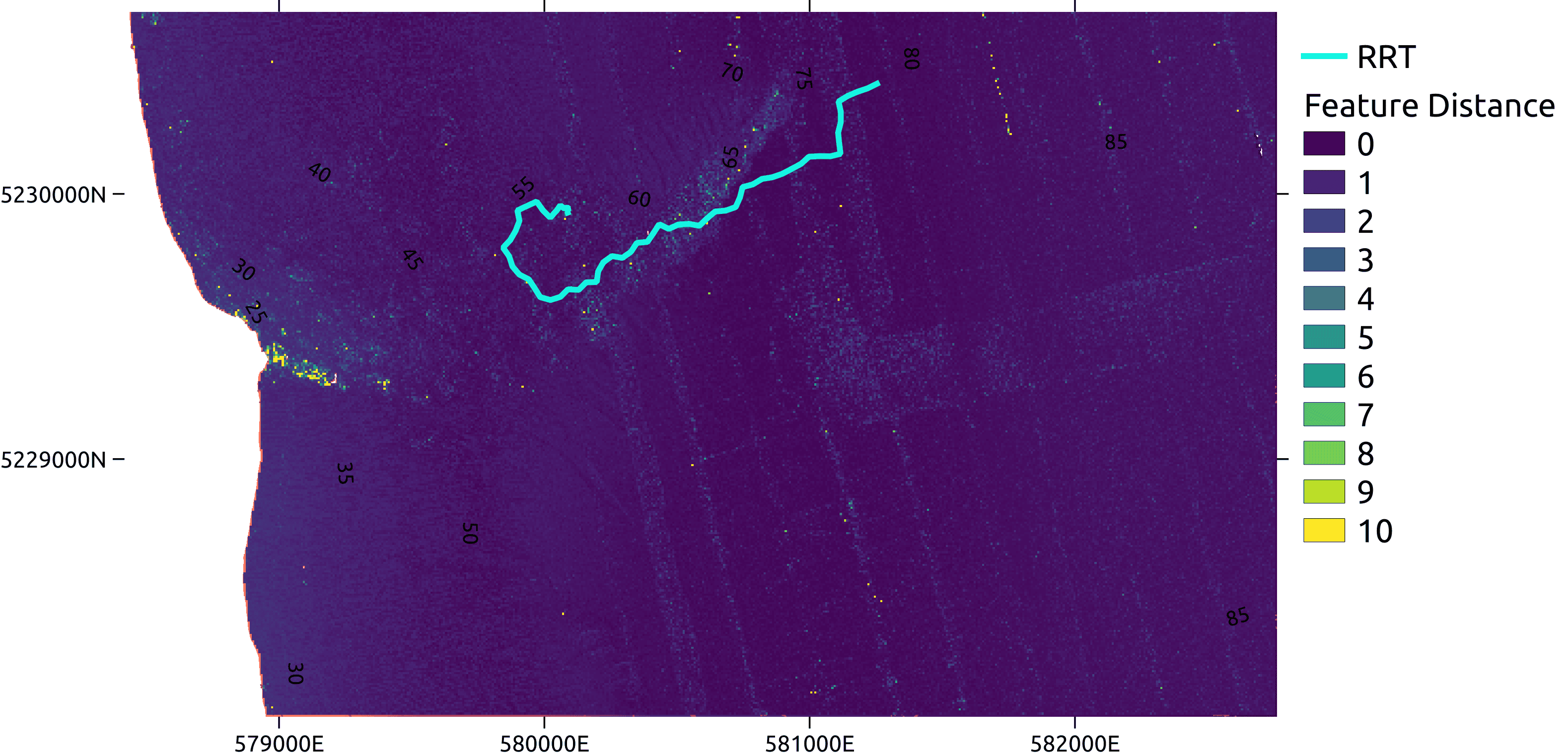}
        \caption{}
        \label{feature_dist:ohara:raw:rrt}
    \end{subfigure}
    ~
    \begin{subfigure}[t]{0.48\columnwidth}
        \centering
        \includegraphics[width=\textwidth]{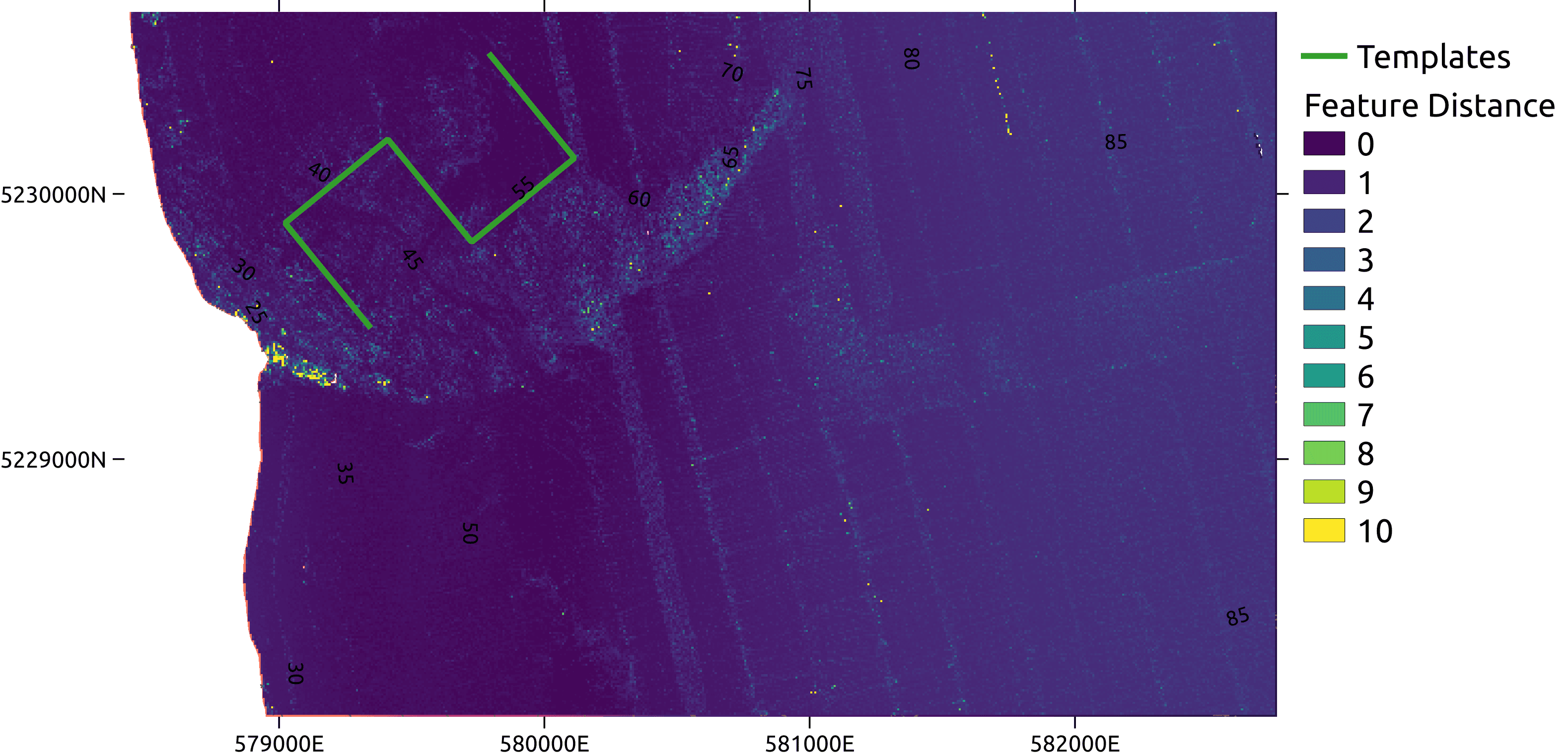}
        \caption{}
        \label{feature_dist:ohara:raw:template}
    \end{subfigure}
    ~
    \begin{subfigure}[t]{0.48\columnwidth}
        \centering
        \includegraphics[width=\textwidth]{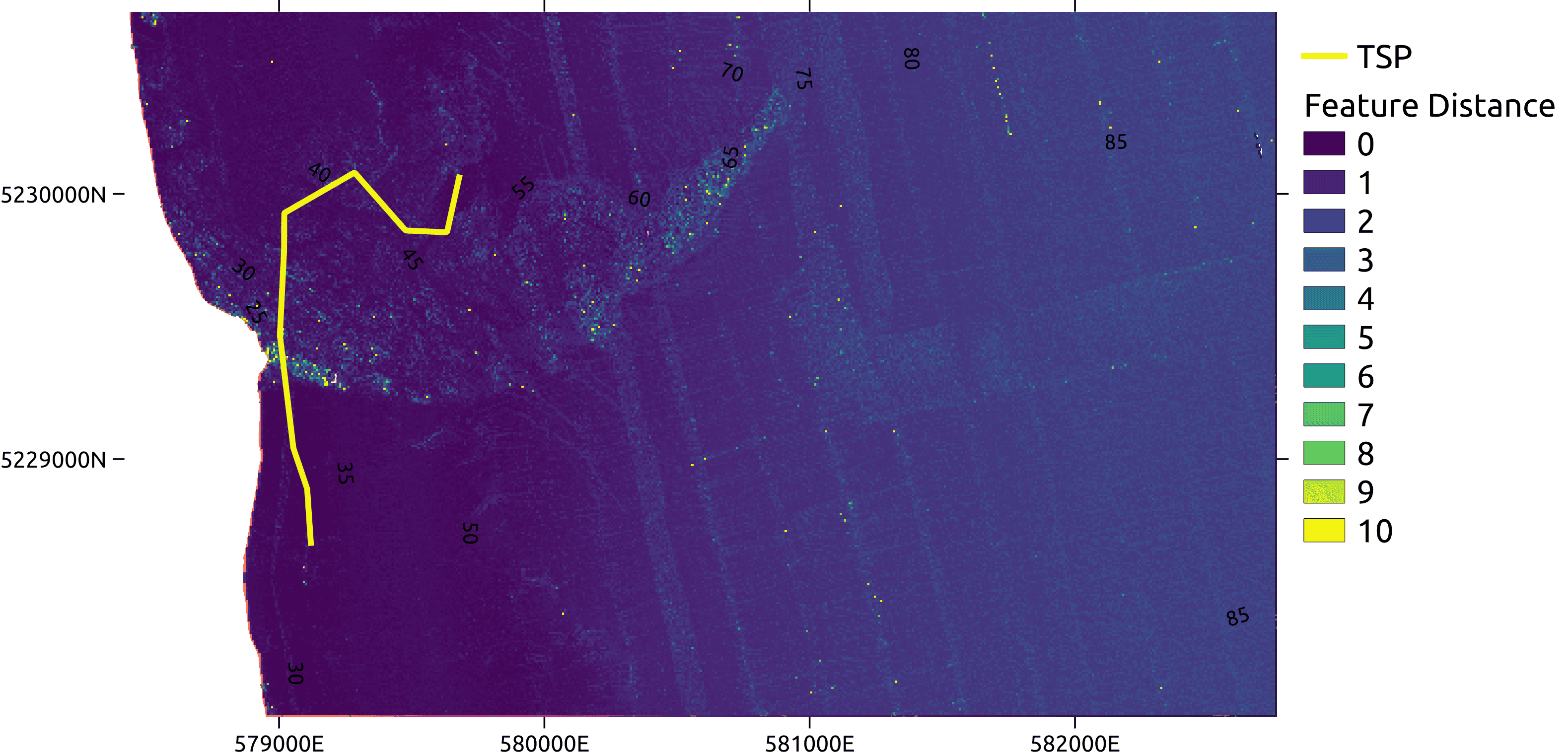}
        \caption{}
        \label{feature_dist:ohara:raw:tsp}
    \end{subfigure}
    ~
    \begin{subfigure}[t]{0.48\columnwidth}
        \centering
        \includegraphics[width=\textwidth]{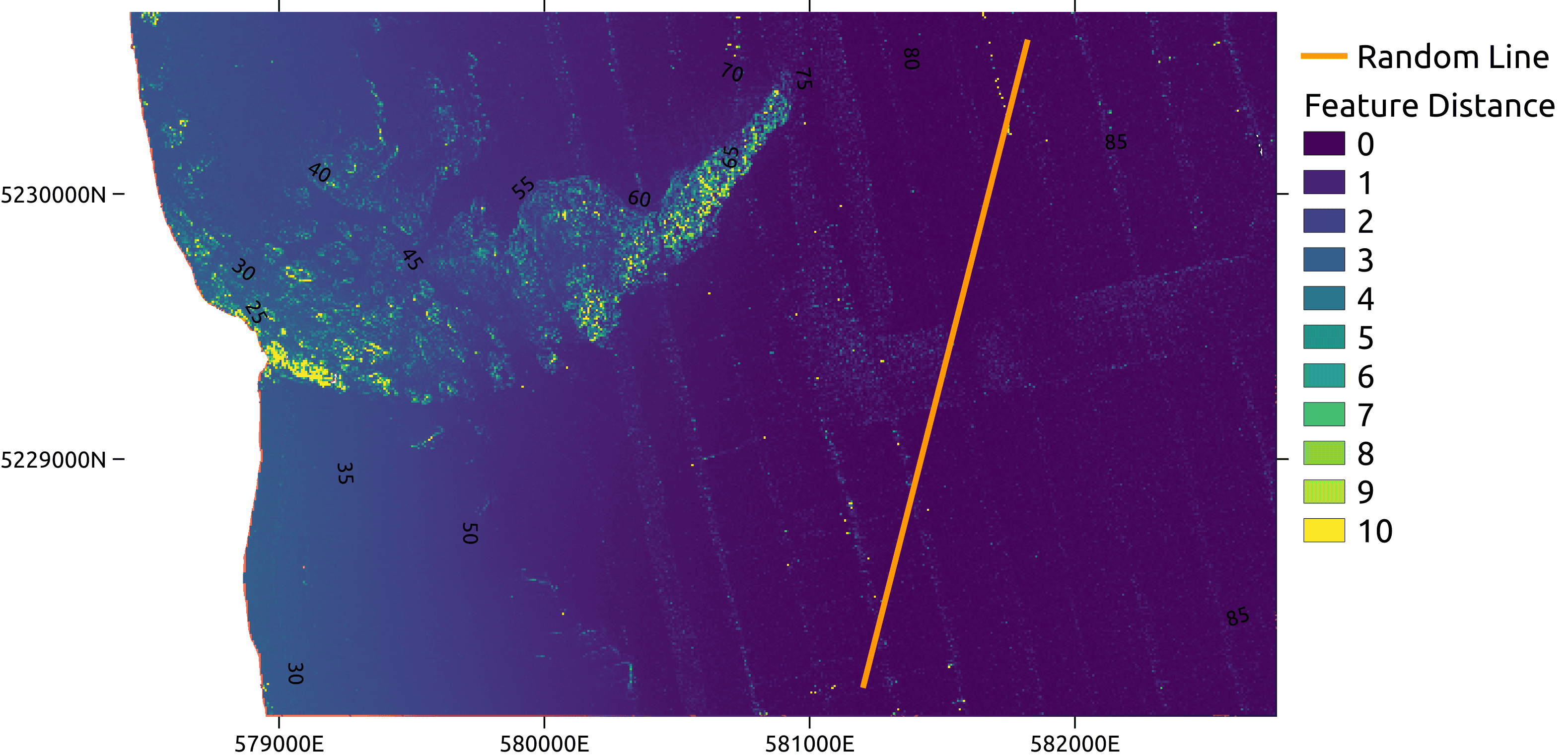}
        \caption{}
        \label{feature_dist:ohara:raw:line}
    \end{subfigure}
    ~
    \caption{Mapping the feature distance from the feature at a given location, to planned initial survey paths for the O'Hara region. (a) \textit{MCTS}, (b) \textit{RRT}, (c) \textit{Templates}, (d) \textit{TSP} and (e) a random line. These figures show the geometric feature distance. Areas that have a small feature distance are well explained by features collected on the respective paths. The \textit{MCTS} and \textit{RRT} surveys exhibit lower feature distances over the survey area.}
    \label{feature_dist:ohara:raw}
\end{figure}

\FloatBarrier
\subsection{Relative Feature Distance}

\begin{figure}[!ht]
\centering
        \includegraphics[width=0.6\columnwidth]{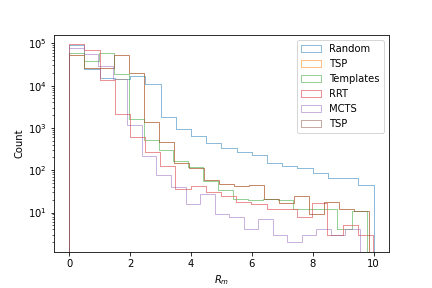}
        \caption{A histogram representation of the feature distance maps in Figure \ref{feature_dist:ohara:raw}. The log-scaled y-axis is the count of pixels corresponding to each feature distance bin. The surveys planned with \textit{MCTS} and \textit{RRT}, have a larger count of smaller feature distances and lesser counts of higher feature distances, indicating a better representation of the feature space. All the informatively planned surveys better characterise the feature space than the random survey.}
        \label{feature_dist:ohara:raw:all:hist}
\end{figure}

The objective of these initial planning surveys is to collect an initial survey that best captures the range of bathymetric terrain present in the area. Providing a set of training examples that comprehensively explore the feature space enables informed habitat models of the area to be created. An effective way to visualise this is to map the feature distance for each point in the survey area, to features collected on the respective initial surveys. This is visualised in Figure \ref{feature_dist:ohara:raw} for geometric features. For the $L_B$ benchmark method, a single random transect is used (Figure \ref{feature_dist:ohara:raw:line}). The dark areas indicate a smaller feature distance, which means that features from that area are similar to features collected on the path. As captured in Figure \ref{feature_dist:ohara:raw} the paths planned with \textit{RRT} and \textit{MCTS}  minimise the feature distance over the entire survey area. The maps in Figure \ref{feature_dist:ohara:raw} highlight the benefit of informed sampling. An uninformed, randomly-placed transect (Figure \ref{feature_dist:ohara:raw:line}) explains the large, sparse area that dominates the survey area, but fails to sample from the reef area that is likely to contain varying benthic habitats. Informed sampling, as demonstrated by the \textit{RRT}, \textit{MCTS}, \textit{Templates} and \textit{TSP} is able to explain the large, sparse area despite only sampling there briefly. These maps are plotted as histograms in Figure \ref{feature_dist:ohara:raw:all:hist}, where the feature distances are binned and counted. The histograms show that informative surveys have a larger number of small feature distances and fewer large distances, highlighting that the informative surveys are a better representation of the entire survey area.
%Each informative survey is compared against the random survey, with the histograms highlighting that the informative surveys have a larger number of small feature distances and fewer large feature distances. Figure \ref{feature_dist:ohara:raw:rrt_vs_tsp:hist} compares the \textit{RRT} survey with the \textit{TSP} survey, and demonstrates the improved feature space coverage of the \textit{RRT} survey. By distributing samples across the feature space, these methods can plan paths that represent the entire survey area. 

\FloatBarrier

\begin{figure}[!ht]
    \centering
    \begin{subfigure}[t]{0.48\columnwidth}
        \centering
        \includegraphics[width=\textwidth]{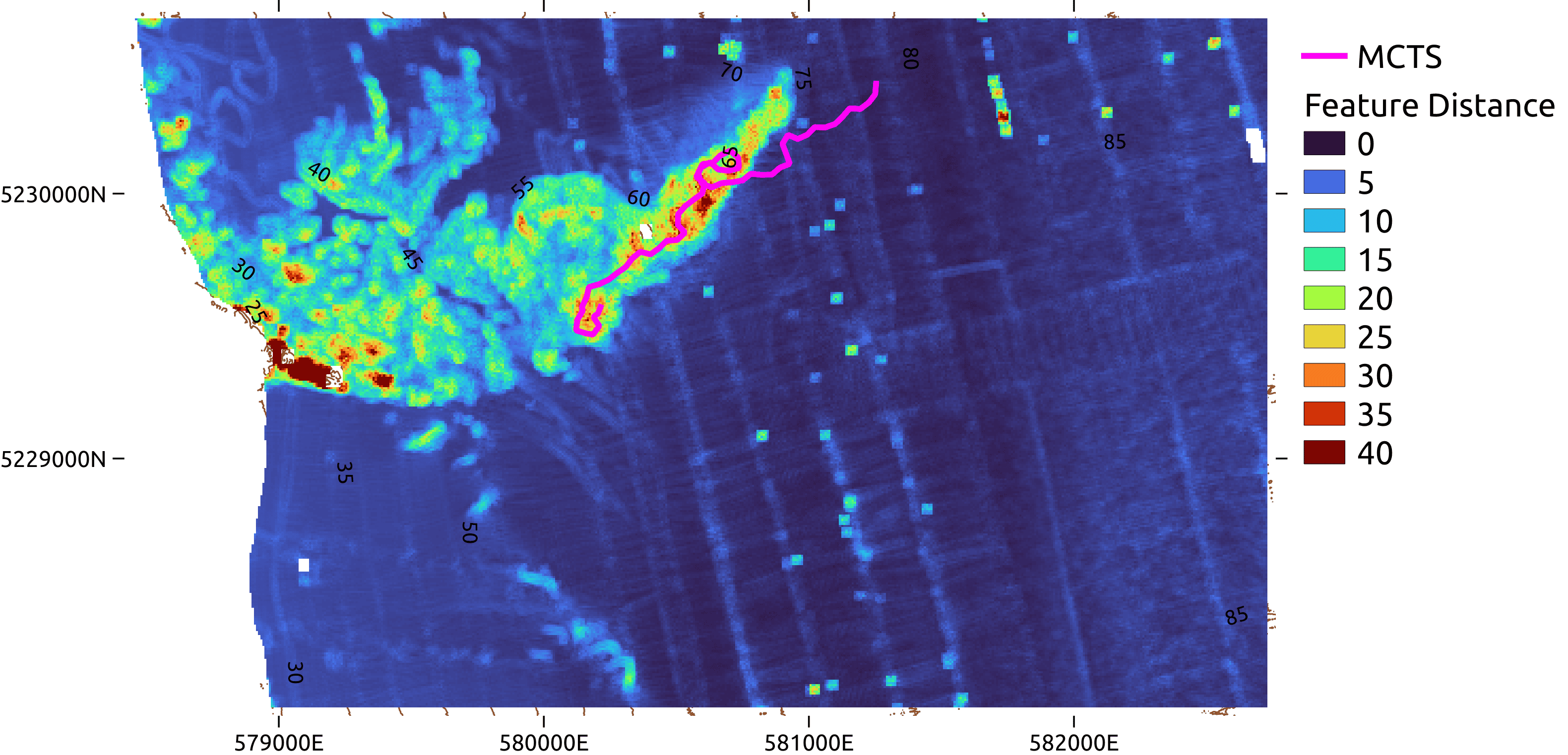}
        \caption{}
        \label{feature_dist:ohara:enc:mcts}
    \end{subfigure}
    ~ 
    \begin{subfigure}[t]{0.48\columnwidth}
        \centering
        \includegraphics[width=\textwidth]{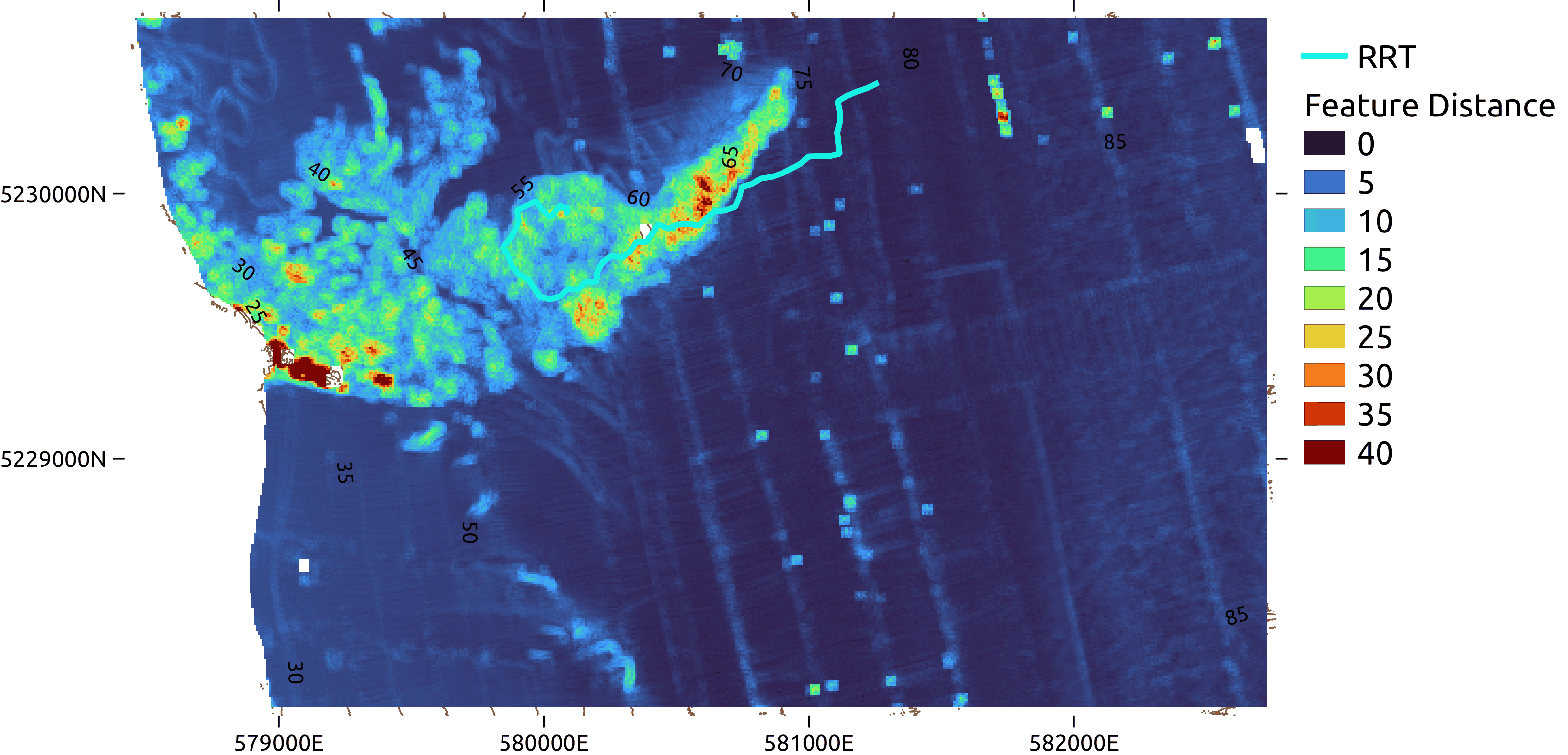}
        \caption{}
        \label{feature_dist:ohara:enc:rrt}
    \end{subfigure}
    ~
    \begin{subfigure}[t]{0.48\columnwidth}
        \centering
        \includegraphics[width=\textwidth]{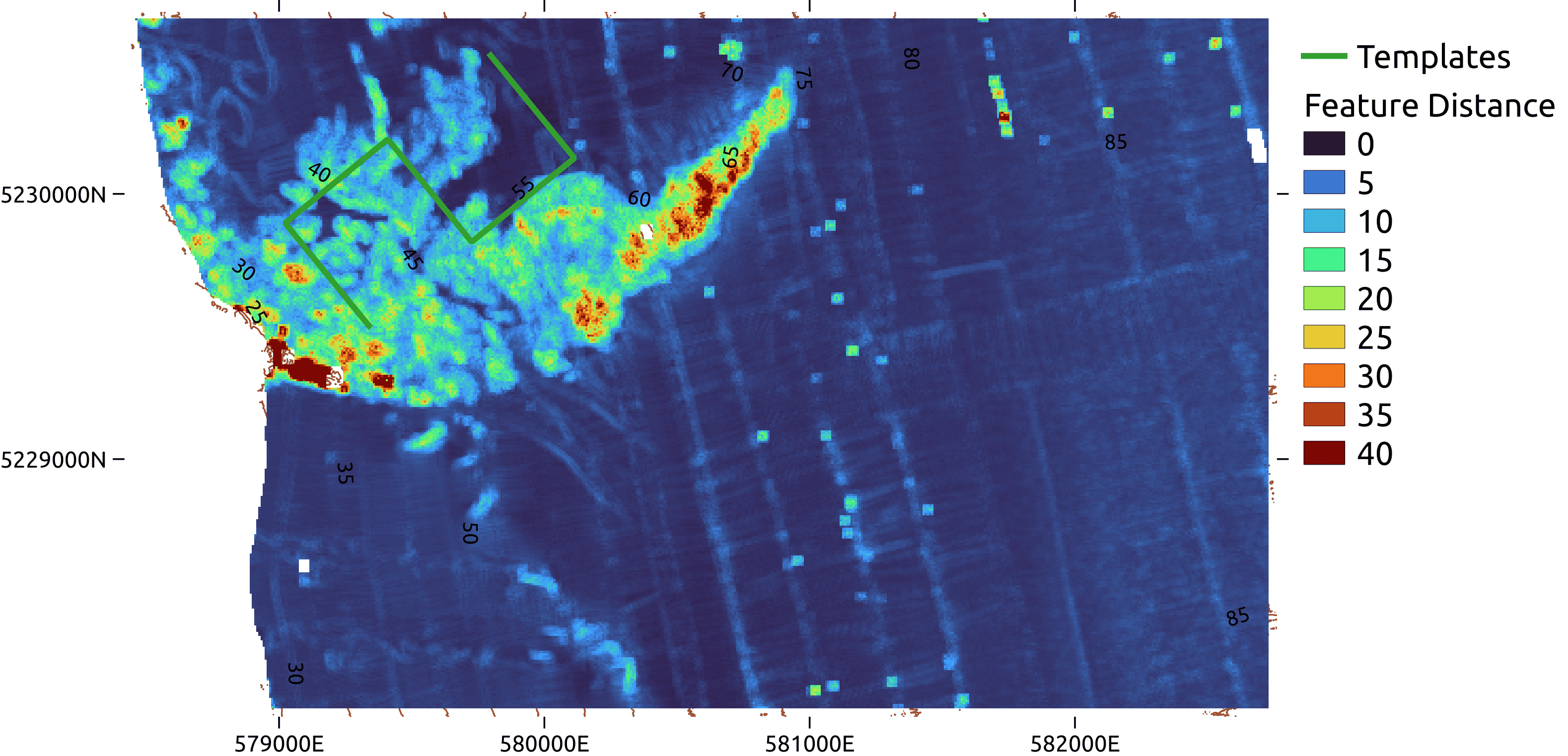}
        \caption{}
        \label{feature_dist:ohara:enc:template}
    \end{subfigure}
    ~
    \begin{subfigure}[t]{0.48\columnwidth}
        \centering
        \includegraphics[width=\textwidth]{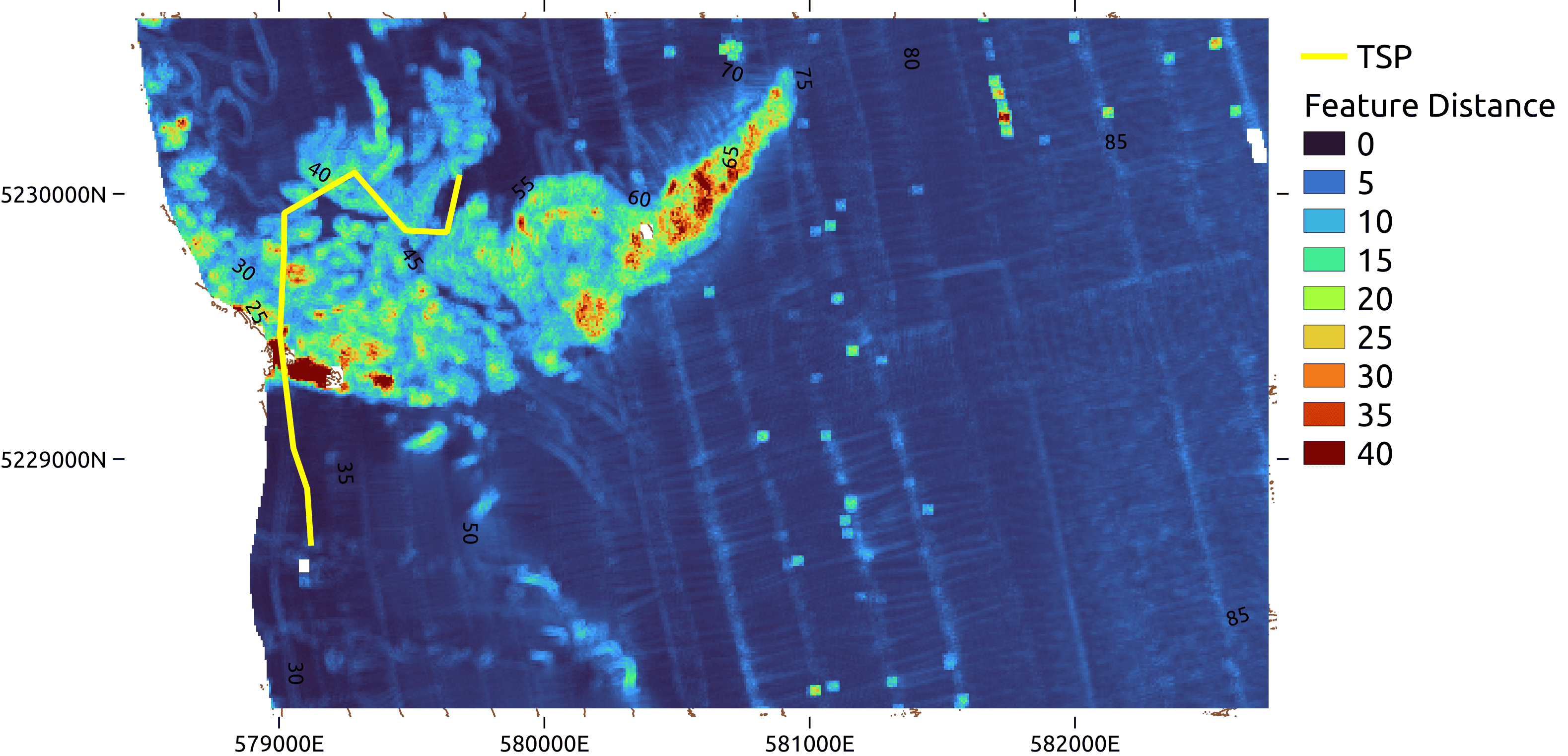}
        \caption{}
        \label{feature_dist:ohara:enc:tsp}
    \end{subfigure}
    ~
    \begin{subfigure}[t]{0.48\columnwidth}
        \centering
        \includegraphics[width=\textwidth]{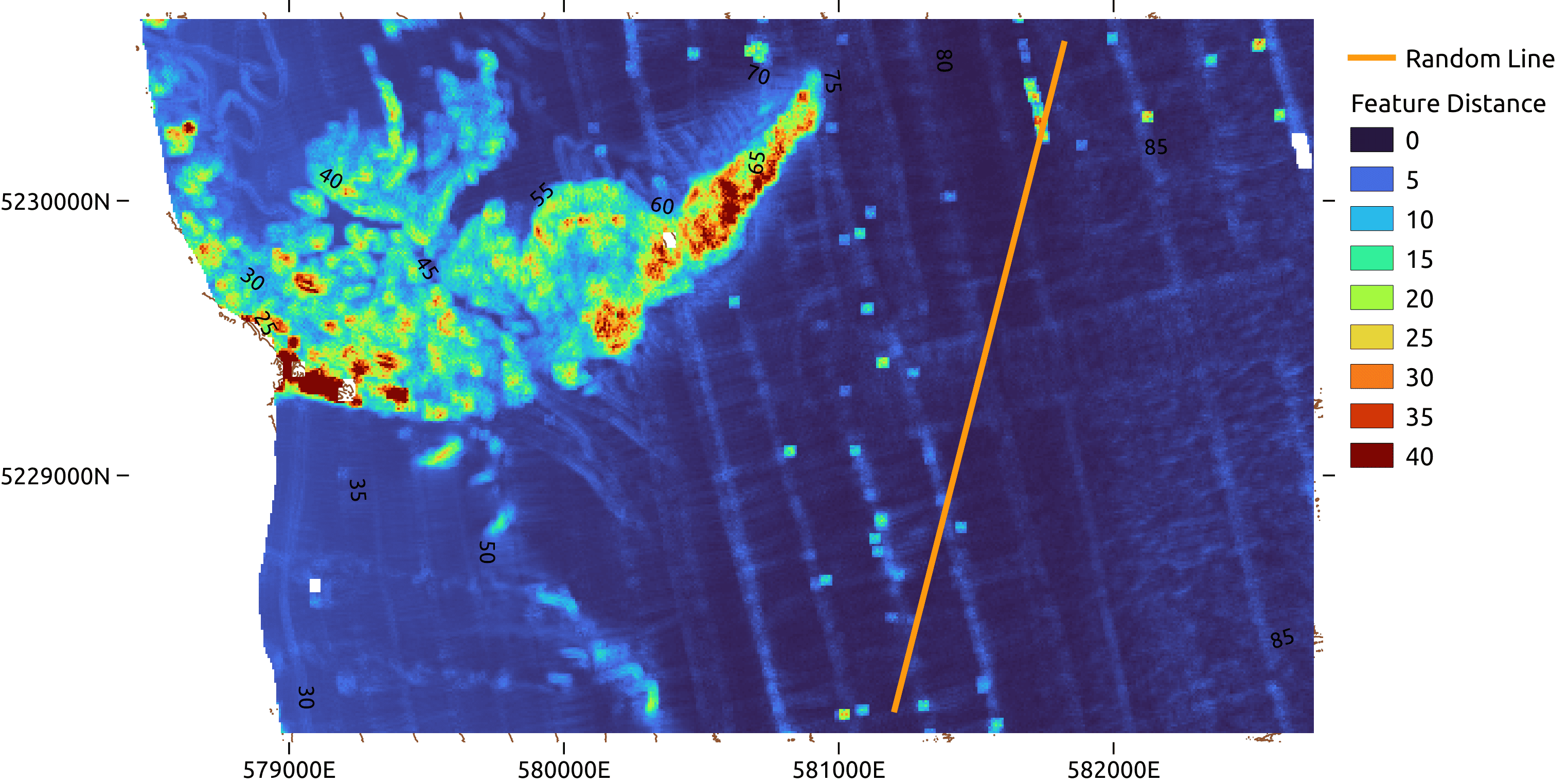}
        \caption{}
        \label{feature_dist:ohara:enc:line}
    \end{subfigure}
    ~
    \caption{Mapping the feature distance from the feature at a given location, to planned initial survey paths for the O'Hara region. These figures show the encoded feature distance. The layout mimics Figure \ref{feature_dist:ohara:raw} but the paths are omitted from the plot as they obscure the view. For reference the plots correspond to (a) MCTS, (b) RRT, (c) Templates, (d) TSP and (e) random transect.
    Due to increased sparsity and distance concentration in higher dimensions, the difference between each map is harder to perceive. The \lq turbo\rq colour map is used to increase contrast.}
    \label{feature_dist:ohara:enc}
\end{figure}

When increasing the dimensions of the feature space, the differences in feature distance can be harder to perceive due to increased sparsity and distance concentration. Features with seemingly small differences between them can have large feature distances. While increasing the feature space dimensions can be advantageous, by prompting the robot to explore interesting areas, it can also lead to the robot over-exploring habitats whose features are diverse, at the expense of sampling from other habitats. Figure \ref{feature_dist:ohara:enc} presents the feature distance for each point in the survey area from the features collected on each respective path. Using this feature representation, there is an advantage of using the \textit{RRT} path (Figure \ref{feature_dist:ohara:enc:rrt}) which is able to best approximate the reef dominated areas of the O'Hara Reef and the vast, bare terrain that exists beyond it.

\FloatBarrier

\subsection{Budget Considerations}

\begin{figure}[!ht]
    \centering
    \begin{subfigure}[t]{0.48\columnwidth}
        \centering
        \includegraphics[width=\textwidth]{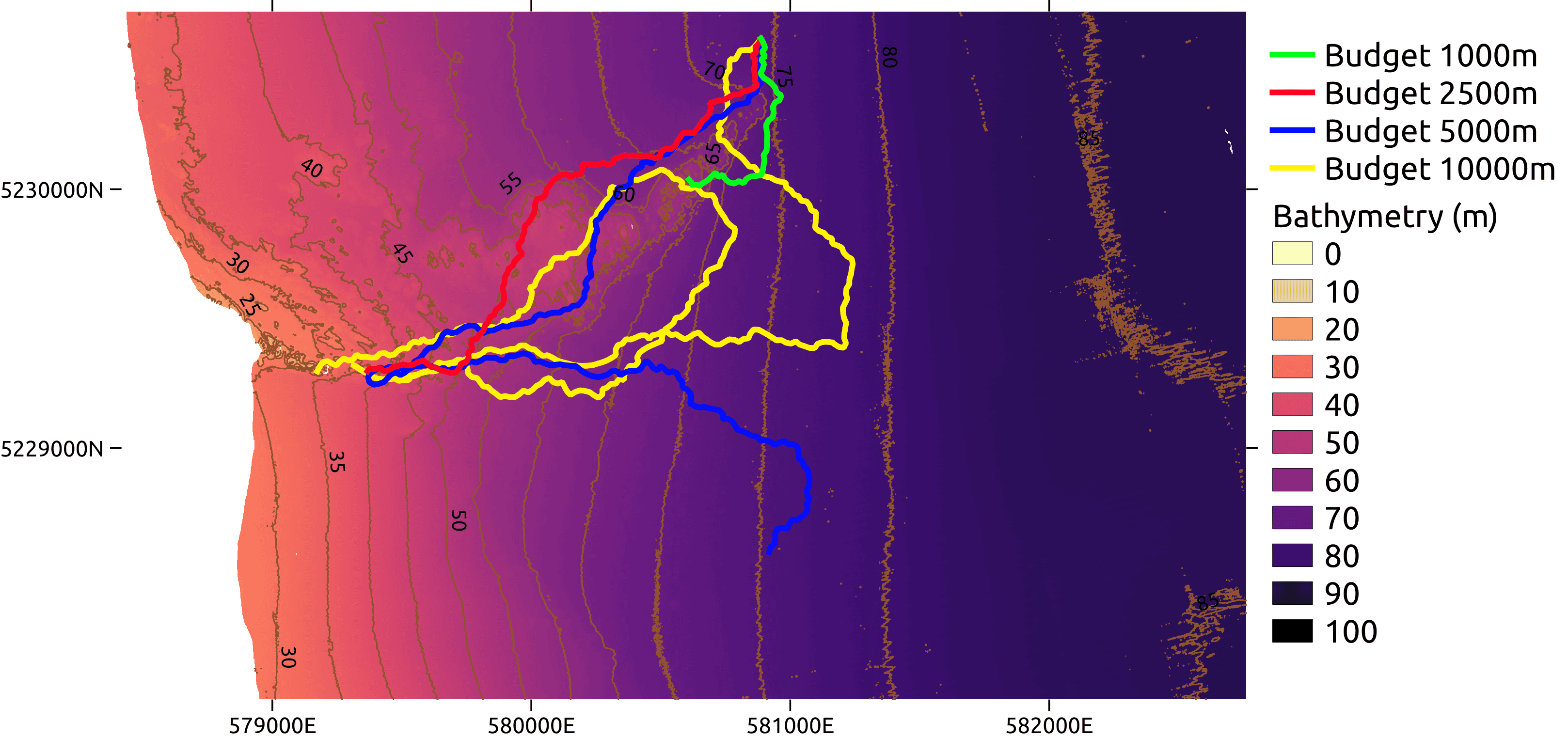}
        \caption{}
        \label{budgets:ohara:rrt:bathy}
    \end{subfigure}
    ~ 
    \begin{subfigure}[t]{0.48\columnwidth}
        \centering
        \includegraphics[width=\textwidth]{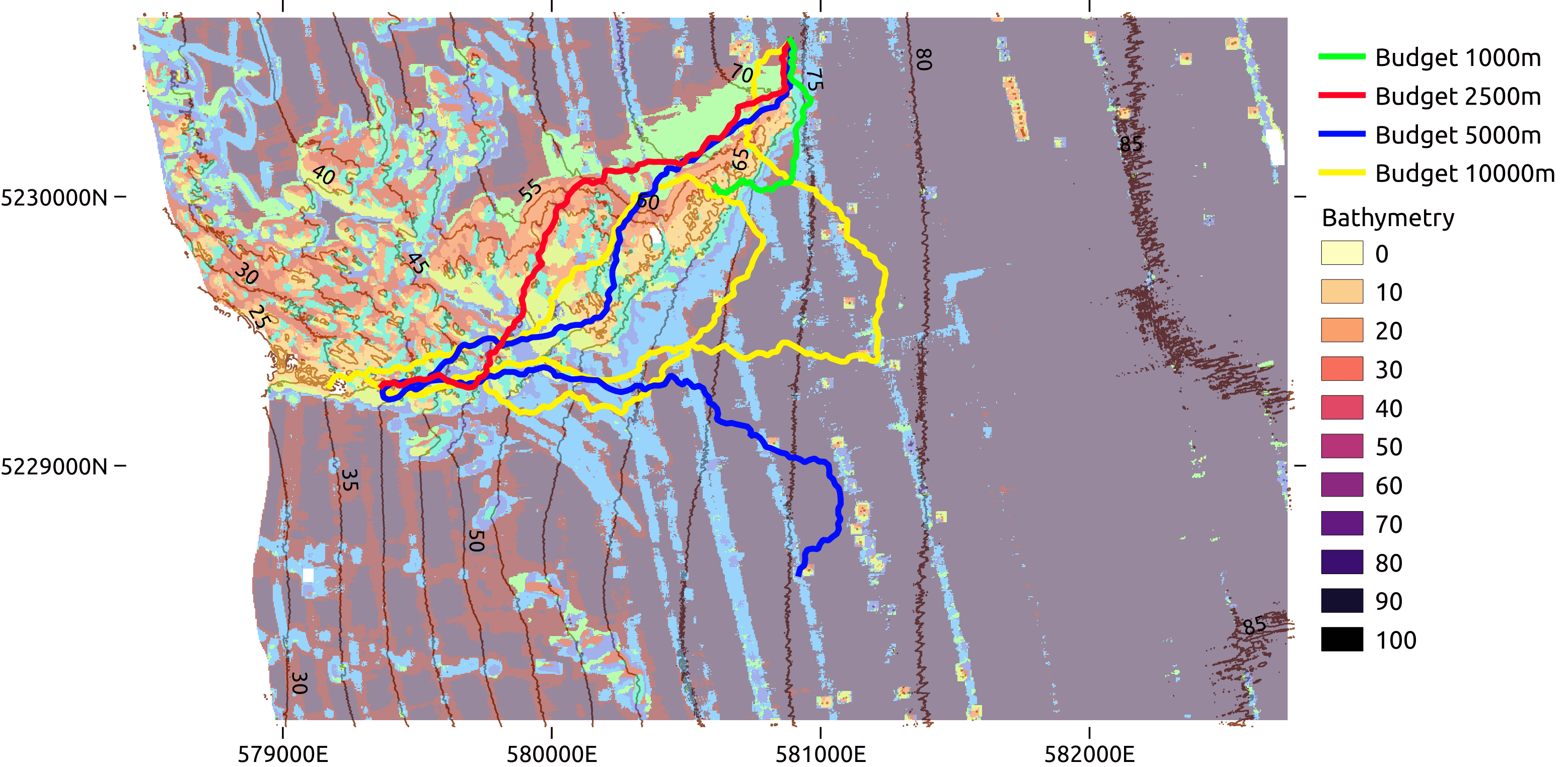}
        \caption{}
        \label{budgets:ohara:rrt:clusters}
    \end{subfigure}
    ~
    \begin{subfigure}[t]{0.48\columnwidth}
        \centering
        \includegraphics[width=\textwidth]{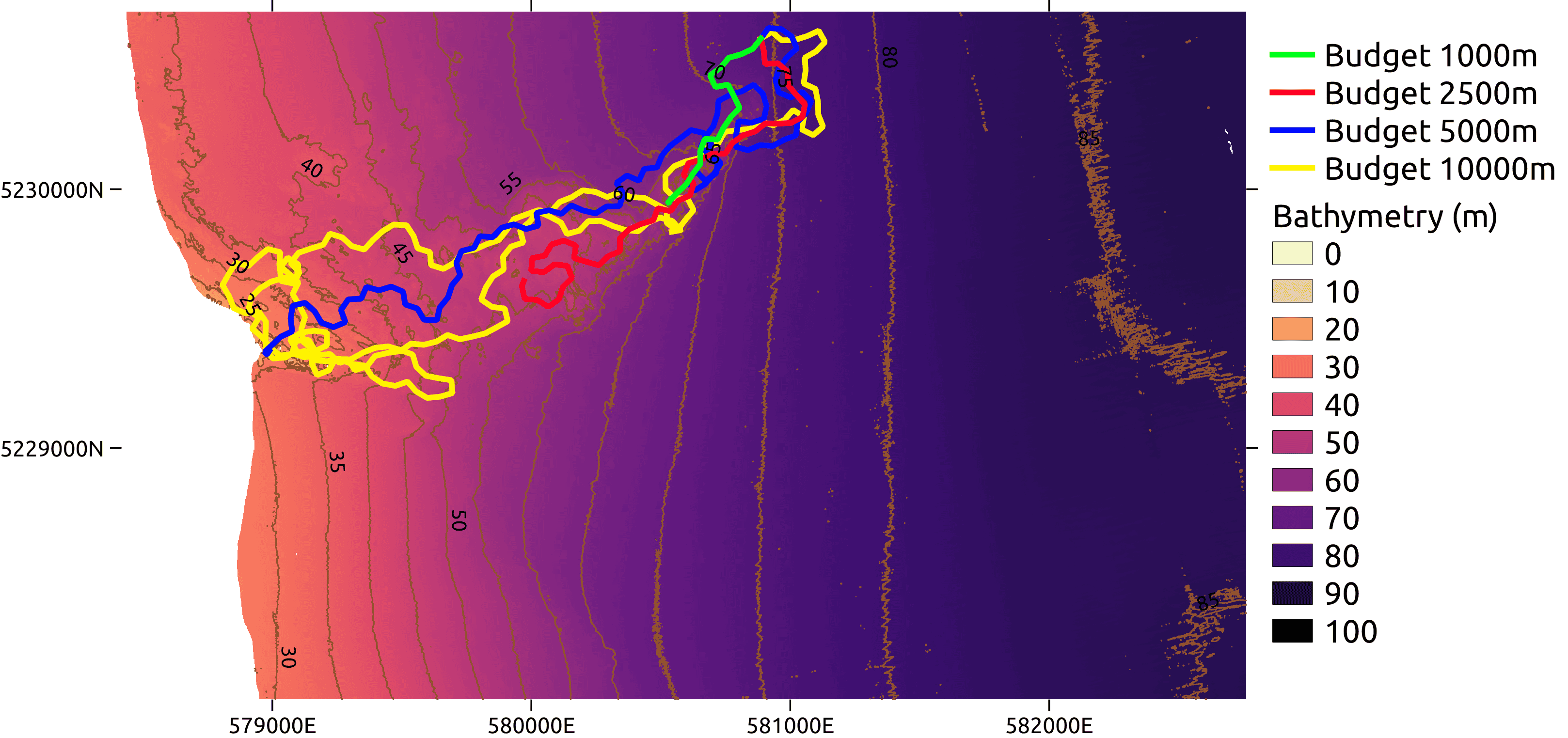}
        \caption{}
        \label{budgets:ohara:mcts:bathy}
    \end{subfigure}
    ~ 
    \begin{subfigure}[t]{0.48\columnwidth}
        \centering
        \includegraphics[width=\textwidth]{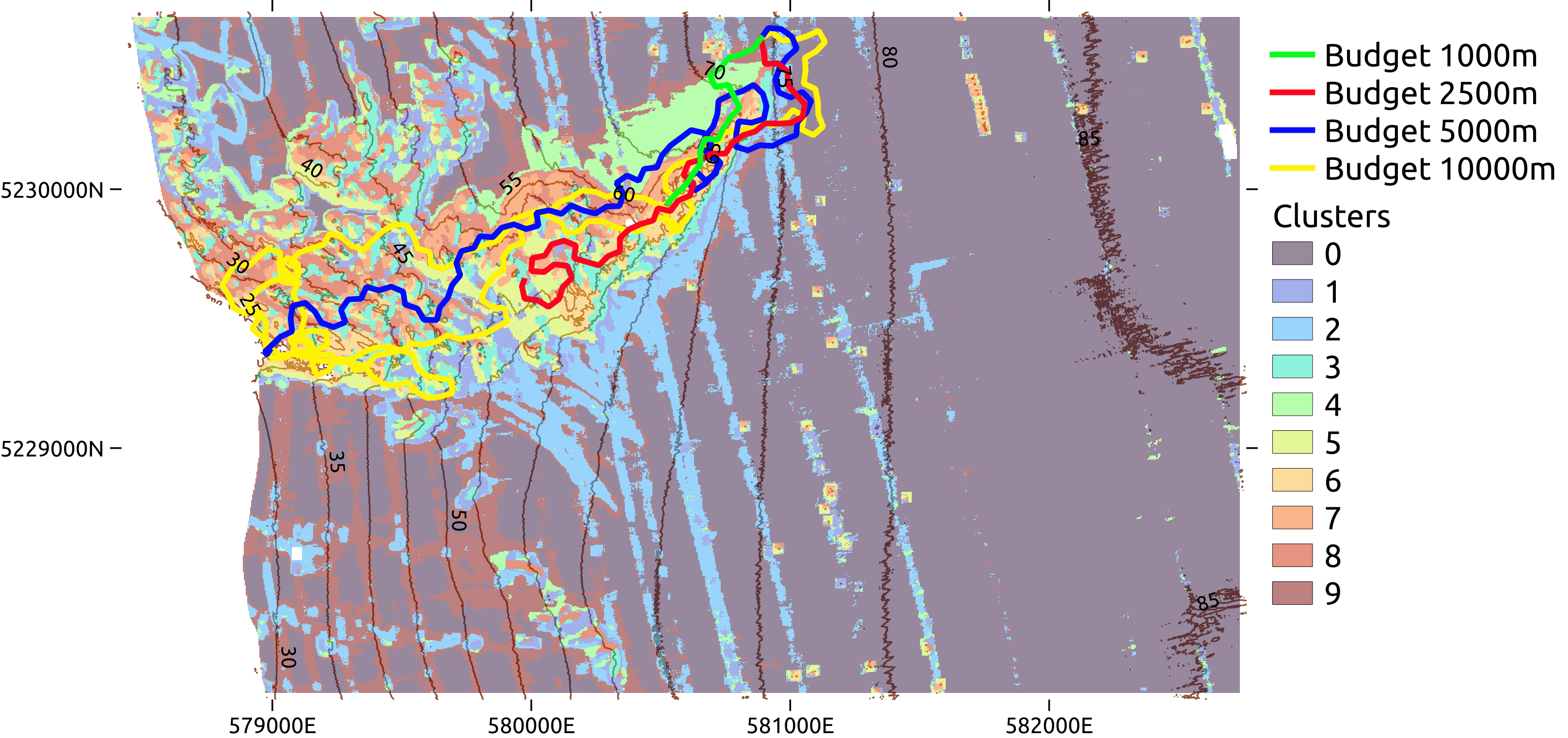}
        \caption{}
        \label{budgets:ohara:mcts:clusters}
    \end{subfigure}
    ~
    \begin{subfigure}[t]{0.48\columnwidth}
        \centering
        \includegraphics[width=\textwidth]{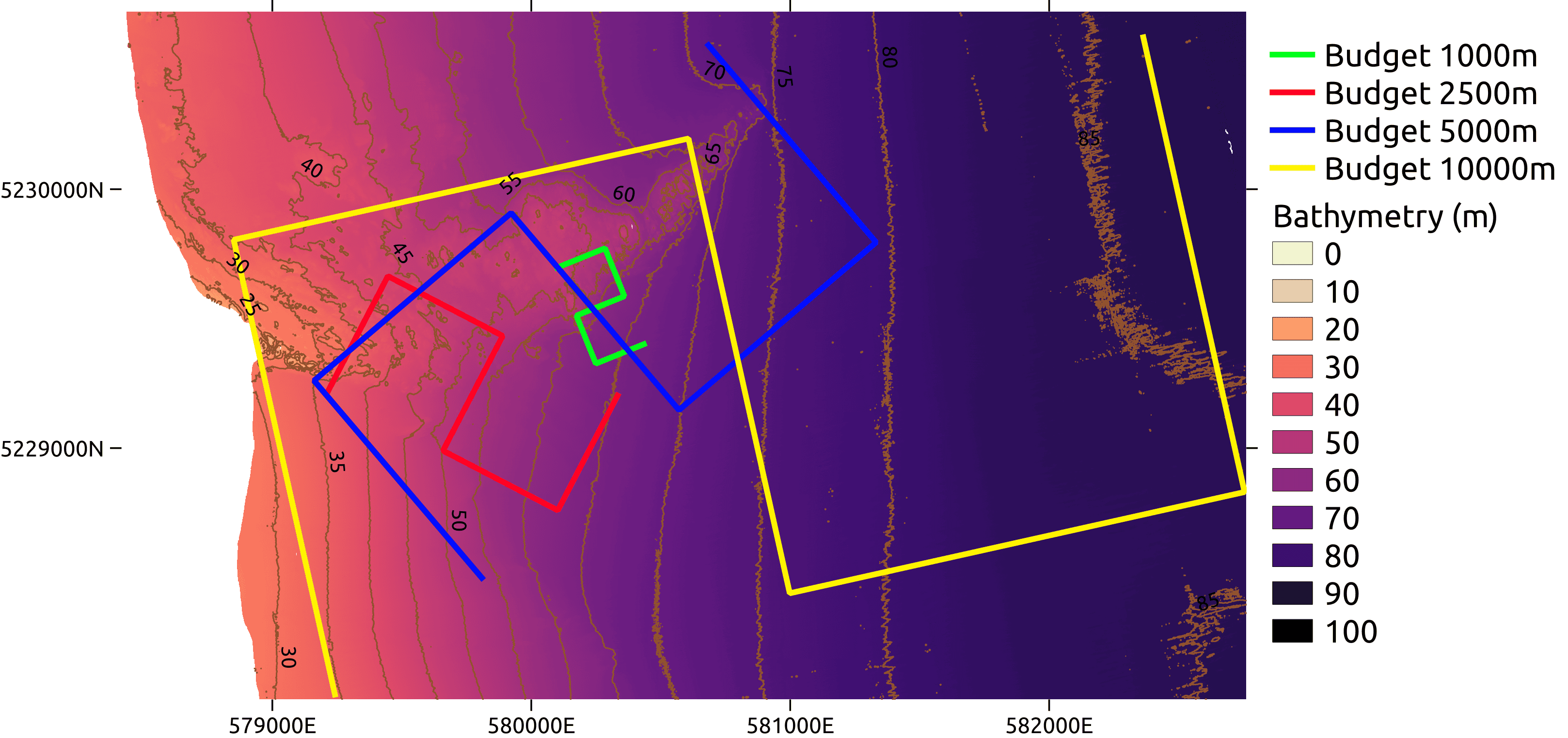}
        \caption{}
        \label{budgets:ohara:template:bathy}
    \end{subfigure}
    ~ 
    \begin{subfigure}[t]{0.48\columnwidth}
        \centering
        \includegraphics[width=\textwidth]{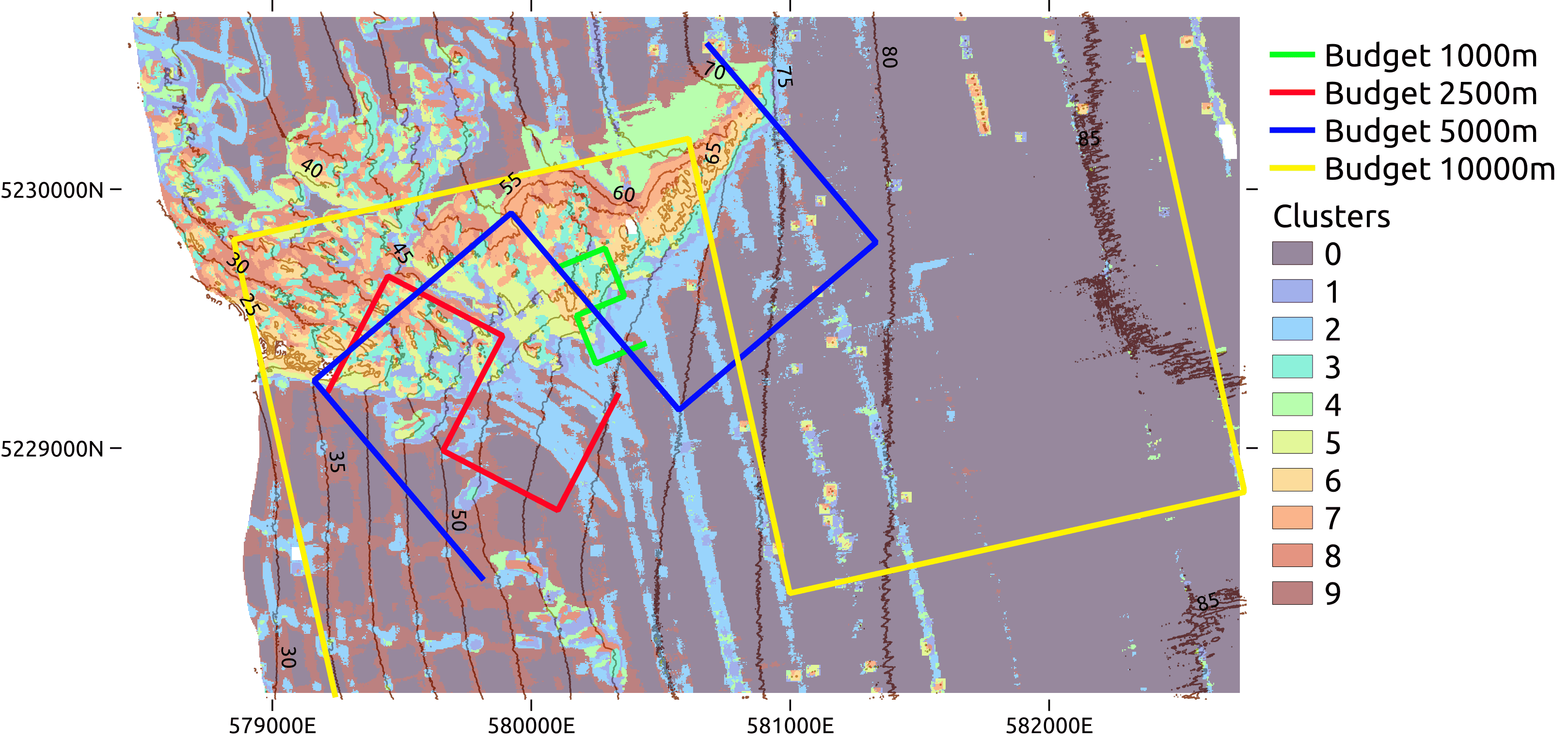}
        \caption{}
        \label{budgets:ohara:template:clusters}
    \end{subfigure}
    ~
    \caption{Paths planned with several distance budgets, using \textit{RRT} (a,b), \textit{MCTS} (c,d) and \textit{Templates} (e,f).}
    \label{budgets:ohara}
\end{figure}

For the results presented in Section \ref{sec:results}, the distance budget is set using the \textit{TSP} method to provide a reasonable distance in which to explore the area. For the deployments (in Section \ref{sec:field_trials}) the budget was set to allow for time for two deployments in a single day. Without this operating constraint, the optimal operating distance budget should be determined that sufficiently characterises the environment. As the paths consider an entire trajectory together, simply using sections of a longer path is not sufficient, as the path may be traversing a low value area to get to a high value area. This experiment looks at planning with several different budgets, to identify at what point further sampling is not effective. The information reward methods (\textit{RRT}, \textit{MCTS} and \textit{Templates}) methods are compared. While a budget limit can be set to \textit{TSP}, it will aim to find the shortest path and so will produce a path under the budget if possible.  The budgets selected were  $\{1000,2500,5000,10000\}$ metres, which offer a range of path lengths than can potentially visit all the different areas of terrain. Paths shorter than 1000m are likely to be ineffective as this distance is too short to visit all the different areas of terrain. Paths longer than 10000m were not evaluated as the set survey template of the \textit{Templates} method cannot fit within the survey area without altering it. It is not possible to compare the  benchmark method $L_B$ as the survey area is too small to fit a 5000m or a 10000m line.The planned paths for each method and distance budget are displayed in Figure \ref{budgets:ohara}. The \textit{RRT} and \textit{MCTS} follow similar trajectories for the smaller budgets, moving from the deeper flatter areas to explore the reef outcrop, making their way towards the shallower rugose areas.

\begin{figure}[!ht]
    \centering
    \begin{subfigure}[t]{0.48\columnwidth}
        \centering
        \includegraphics[width=\textwidth]{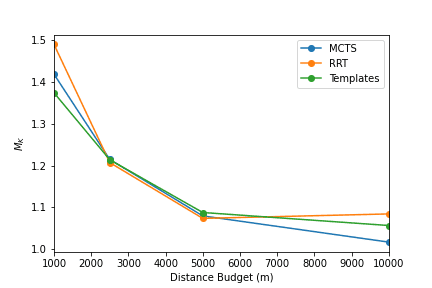}
        \caption{}
        \label{budgets:ohara:budget_vs_mk}
    \end{subfigure}
    ~ 
    \begin{subfigure}[t]{0.48\columnwidth}
        \centering
        \includegraphics[width=\textwidth]{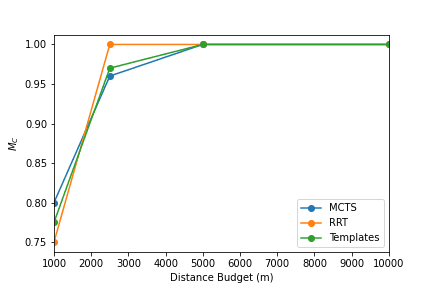}
        \caption{}
        \label{budgets:ohara:budget_vs_mc}
    \end{subfigure}
    ~
    \caption{Plots show the $M_K$ metric (a) and the $M_C$ metric (b) for each planning method with several different budgets. With a 1000m budget, the paths do not sufficiently explore the feature space. The $M_C$ metric shows that after after 5000m all paths have visited all the clusters. The \textit{MCTS} method shows continual improvement on the $M_K$ metric, while the \textit{RRT} and \textit{Templates} exhibit little performance improvement with a budget of 10000m.}
    \label{budgets:ohara:budget_vs}
\end{figure}

Figure \ref{budgets:ohara:budget_vs_mc} shows the proportion of clusters visited for each path at each distance budget. With a budget of 1000m, the paths fail to visit all the clusters, suggesting this path length is too short. For a budget of 2500m, the \textit{RRT} visits all the clusters while the \textit{MCTS} and \textit{Templates} methods visit on average 96\% and 97\% of clusters respectively. At 5000m, all the paths visit all of the clusters and there is a significant elbow point in the $M_K$ metric for all the methods. For the \textit{Templates} method, the larger survey template of the 10000m path could not effectively align with the terrain, while the \textit{RRT} method offered no improvement with the 10000m survey. However, the 10000m surveys result in a further reduction in the $M_K$ metric for the \textit{MCTS} method, highlighting the benefit of further exploration. This shows that further exploration can be beneficial, even when the path has visited all the clusters. When trying to identify an effective budget, proposing a value of $M_K$ at which the environment is completely explored would be arbitrary and misleading, however evaluating a range of budgets can identify elbow points, where further exploration is less beneficial. Selecting the budget so that the path can visit all the clusters is a more effective way to determine an efficient budget.

%However, just visiting all of the clusters complete as there can be diverse terrain contained within a cluster. In Section \ref{sec:results}, the \textit{TSP} method visited all of the clusters but did not perform as well on the $M_K$ metric, highlighting the need to more comprehensively explore within a cluster. For a budget of 1000m, the \textit{Templates} method performs the best, suggesting the starting position of the \textit{RRT} and \textit{Templates} method is not optimal for the budget. The \textit{Templates} method has a lower $M_K$ at 5000m than 10000m, as the 10000m path can barely fit within the survey area and cannot align efficiently with the terrain. The \textit{MCTS} and \textit{RRT} paths continuously reduce the $M_K$ metric as distance increases, highlighting that further exploration can be beneficial. Proposing a value of $M_K$ at which the environment is completely explored would be arbitrary and misleading. For an initial pass, visiting all the clusters would be a better measure of this. Future work will investigate the utility of a longer survey versus recovering the AUV and moving to a different starting position. 

\FloatBarrier

\section{Field Deployments}\label{sec:field_trials}

Experimental field trials of the informative path planning were carried out by the Universitat de Girona with the Sparus II AUV \citep{Carreras2018} exploring a canyon offshore of Blanes, Spain. The objective was to explore the bathymetric feature space with a limited budget. The shorter paths are compared against a longer broad grid. The broad grid was generated automatically from a large bounding box covering the area of interest. It was designed to cover a large area while also exploring the depth range of the map. The original broad grid was trimmed to allow for the survey to be conducted in a single day. A budget of 2500m was set for the shorter paths, allowing two surveys to be conducted in a single day. An encoded feature space was used but with a smaller patch size of $11\times11$ cells, allowing for a feature dimension of 16 to reduce the negative impacts of the high dimensionality present in Section \ref{sec:results}.

When deploying the AUVs in the field, the plans generated have to ensure safe operation of the AUV. For these field trials, the position of the altitude sensor meant the AUV could not go up steep inclines so the paths were restricted to go predominantly downhill and only allow a rise of 15\%. This restricts the type of terrain the AUV can explore. This constraint was integrated into the planners. For the incremental planners (\textit{MCTS} and \textit{RRT}), constraints can be integrated naturally, where if the path expansion violates the constraint, it is disallowed. For the \textit{Templates} method, the constraints have to be checked for the entire path and if any section of the path breaks the constraint that survey template position is disallowed. This is much less effective and leads to a large number of disallowed surveys, whereas the incremental planners can potentially plan around the area violating the constraint.

\begin{figure}[!ht]
    \centering
    \begin{subfigure}[t]{0.48\columnwidth}
        \centering
        \includegraphics[width=\textwidth]{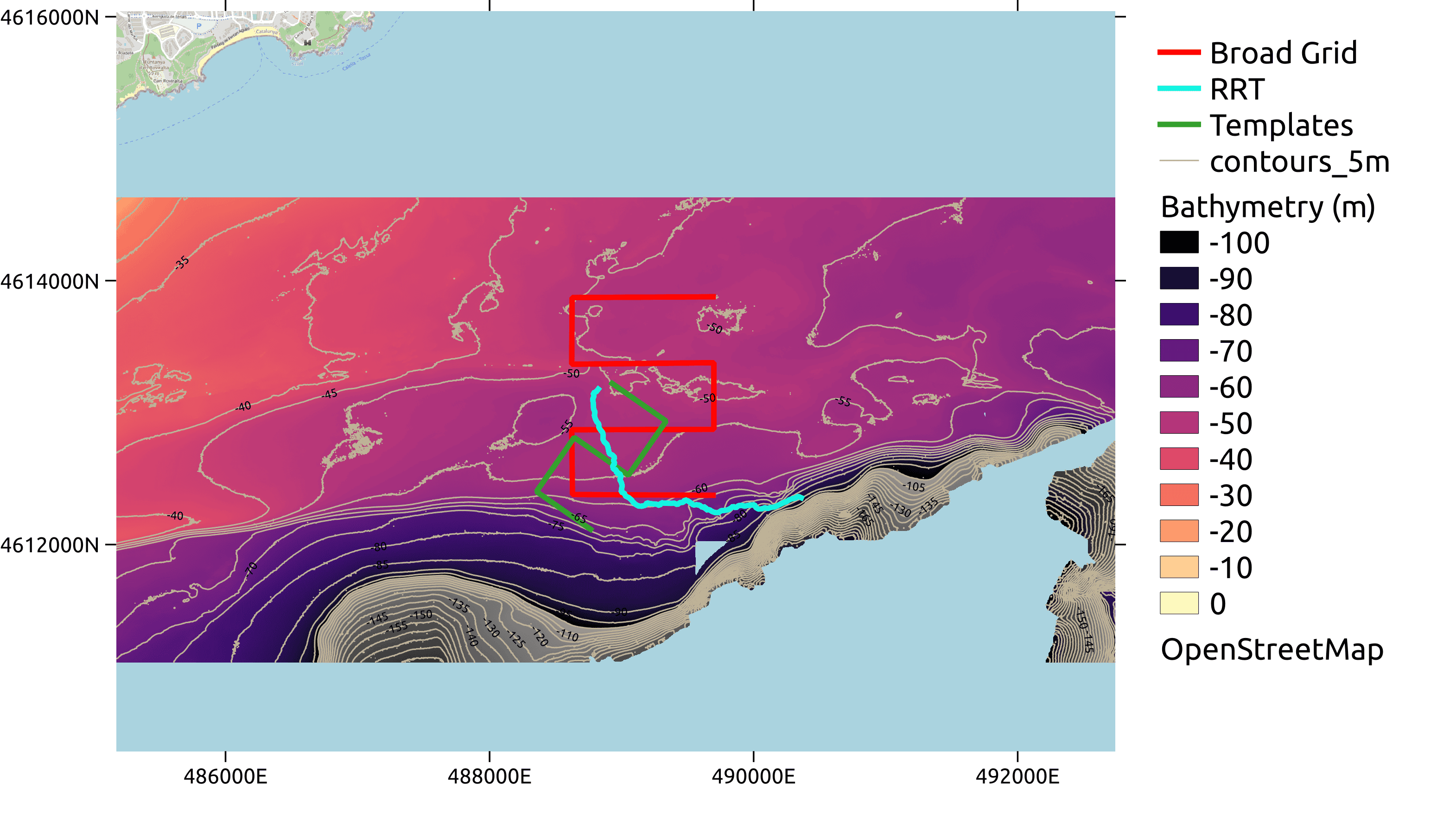}
        \caption{}
        \label{fig:field:plans:bathy}
    \end{subfigure}
    ~ 
    \begin{subfigure}[t]{0.48\columnwidth}
        \centering
        \includegraphics[width=\textwidth]{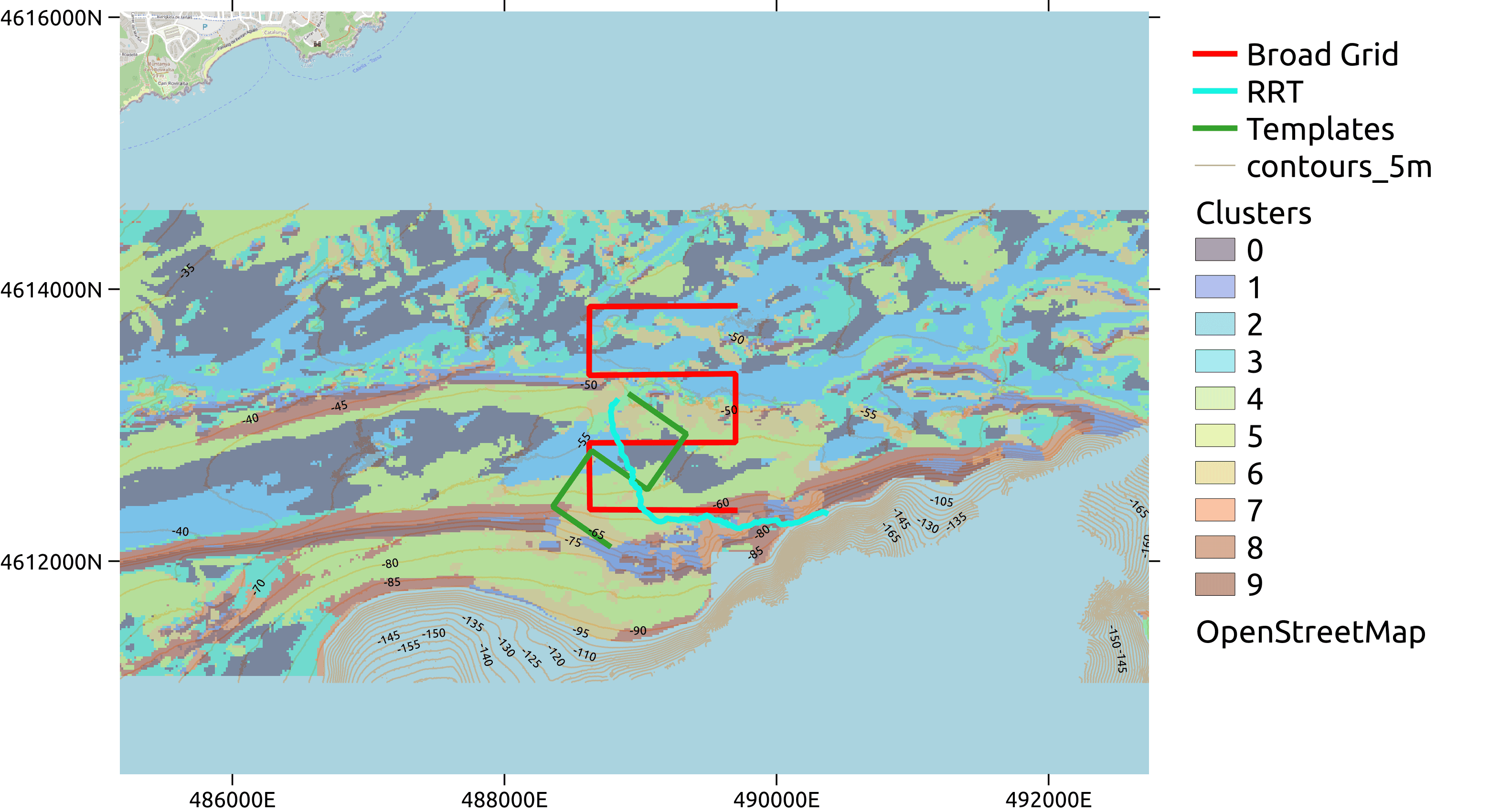}
        \caption{}
        \label{fig:field:plans:clusters}
    \end{subfigure}
    ~
    \begin{subfigure}[t]{0.48\columnwidth}
        \centering
        \includegraphics[width=\textwidth]{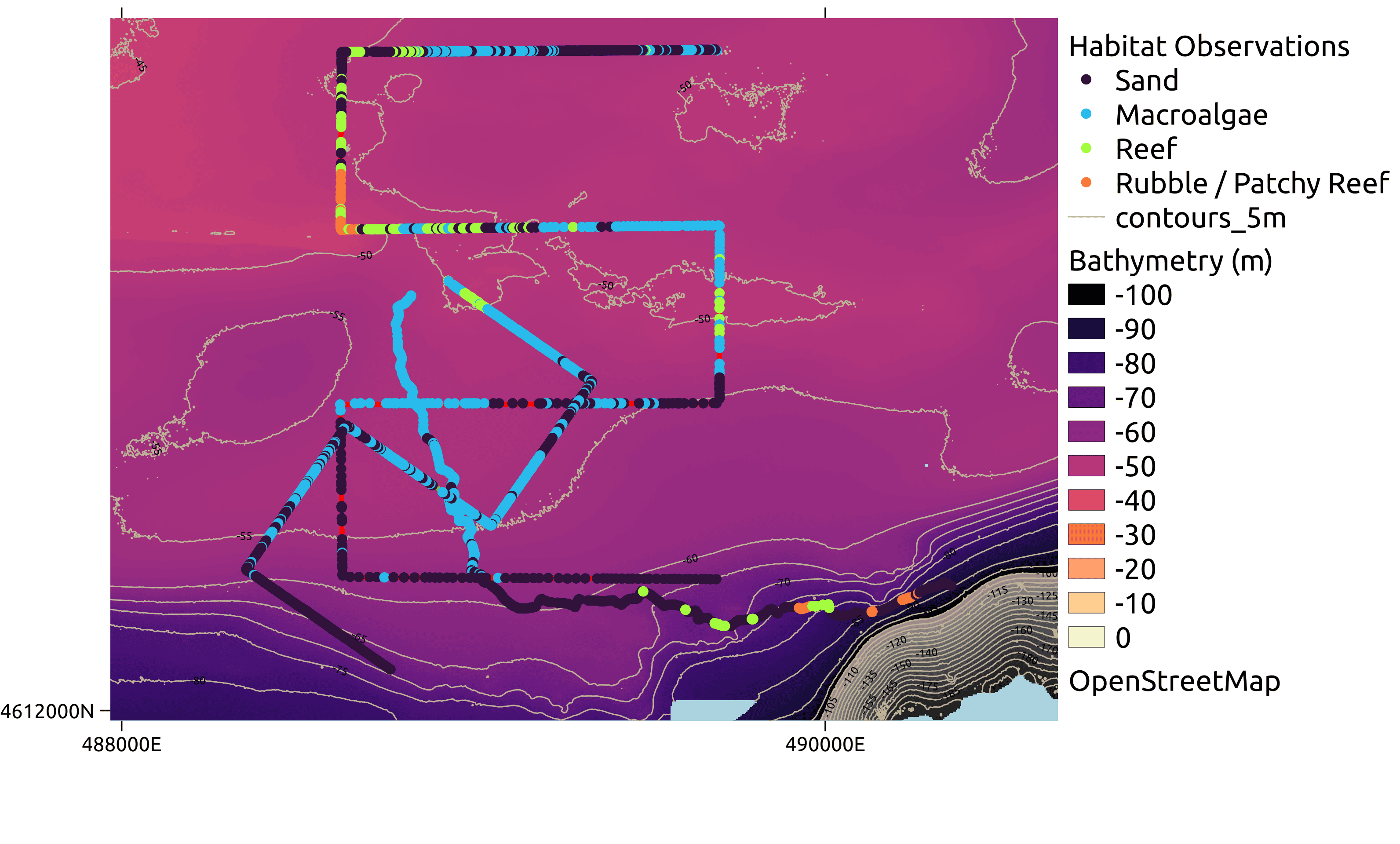}
        \caption{}
        \label{fig:field:hablabels:bathy}
    \end{subfigure}
    ~ 
    \begin{subfigure}[t]{0.48\columnwidth}
        \centering
        \includegraphics[width=\textwidth]{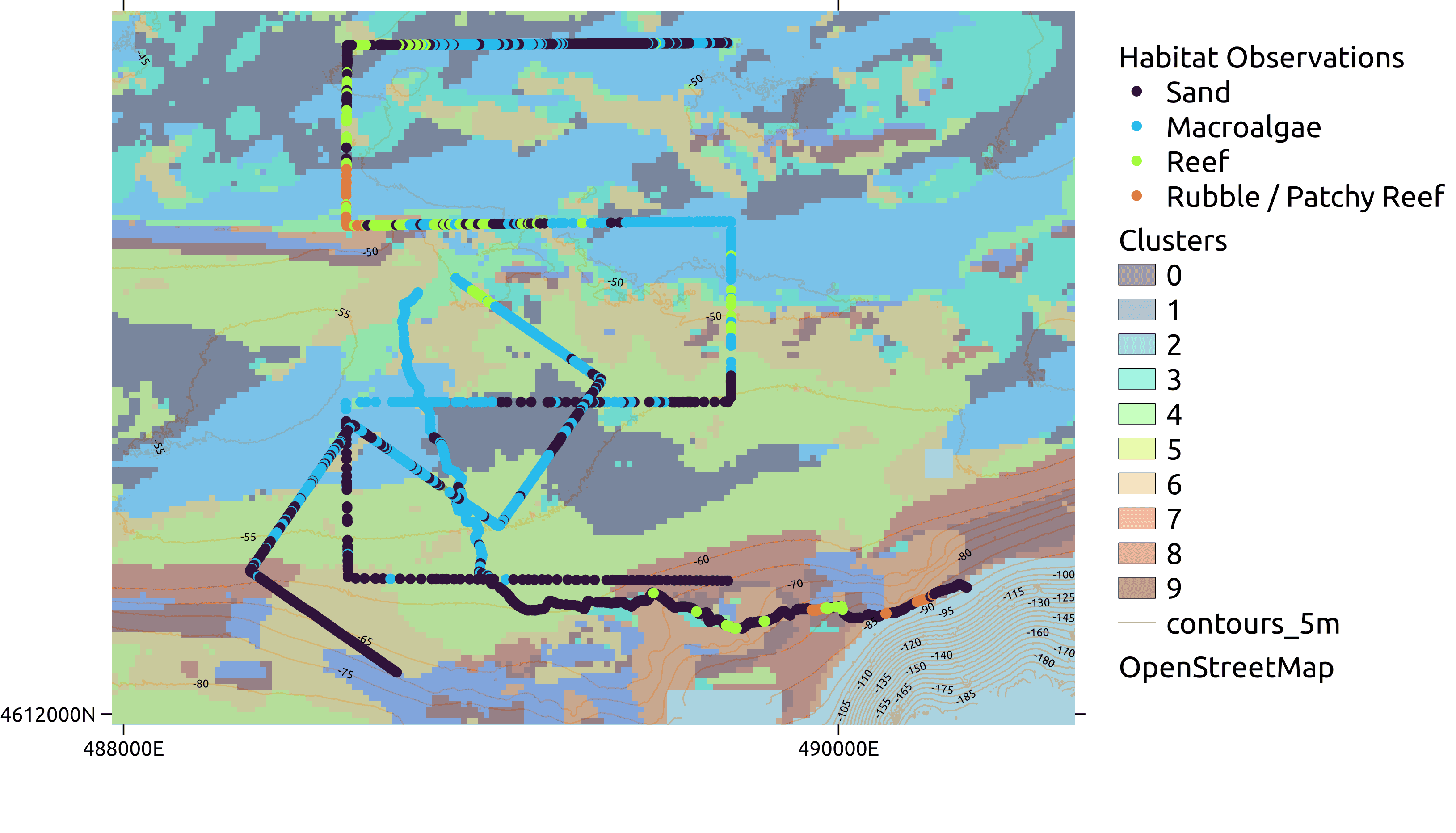}
        \caption{}
        \label{fig:field:hablabels:clusters}
    \end{subfigure}
    ~
    \caption{Paths planned and performed by the Sparus II AUV exploring a canyon offshore of Blanes, Spain. The left column shows the paths over the bathymetry, while the right shows the path over the bathymetric clusters. The top row highlights each of the respective paths, while the bottom row shows the habitat observations, obtained from the Sparus II imagery. The RRT and Benchmark paths observed all the the habitats.}
    \label{fig:field:plans}
\end{figure}

\begin{table}[!ht]
\centering
\begin{tabular}{|c|i|i|r|}
\hline
             & RRT  & Templates & Broad Grid \\ \hline
$M_{PD}$          & \textbf{3.24} & 1.97      & 2.11       \\
$M_K$           & 2.39 & 2.50      & \textbf{2.34}       \\
$M_C$           & \textbf{1.0}  & 0.8       & \textbf{1.0}        \\
D(m) & \textbf{2499} & 2500      & 5792      \\ \hline
\end{tabular}
\caption{Evaluation metrics for the field deployments near Blanes, Spain.}
\label{tab:field:results}
\end{table}

The paths planned by each method are displayed in Figures \ref{fig:field:plans:bathy} and \ref{fig:field:plans:clusters}. The respective evaluation metrics are displayed in Table \ref{tab:field:results}. The \textit{Broad Grid} displays slightly better feature space exploration metrics (a lower $M_K$) but also travels significantly further, 5792m vs 2500m for the informative trajectories.  The \textit{RRT} method has a slightly worse $M_K$ metric, but it still visits all of the clusters. The template method performs relatively poorly compared to the other methods, indicating that the chosen template type is not well suited to this area.

\begin{figure}[!ht]  % old col width = 0.57
    \centering
    \includegraphics[width=0.8\columnwidth]{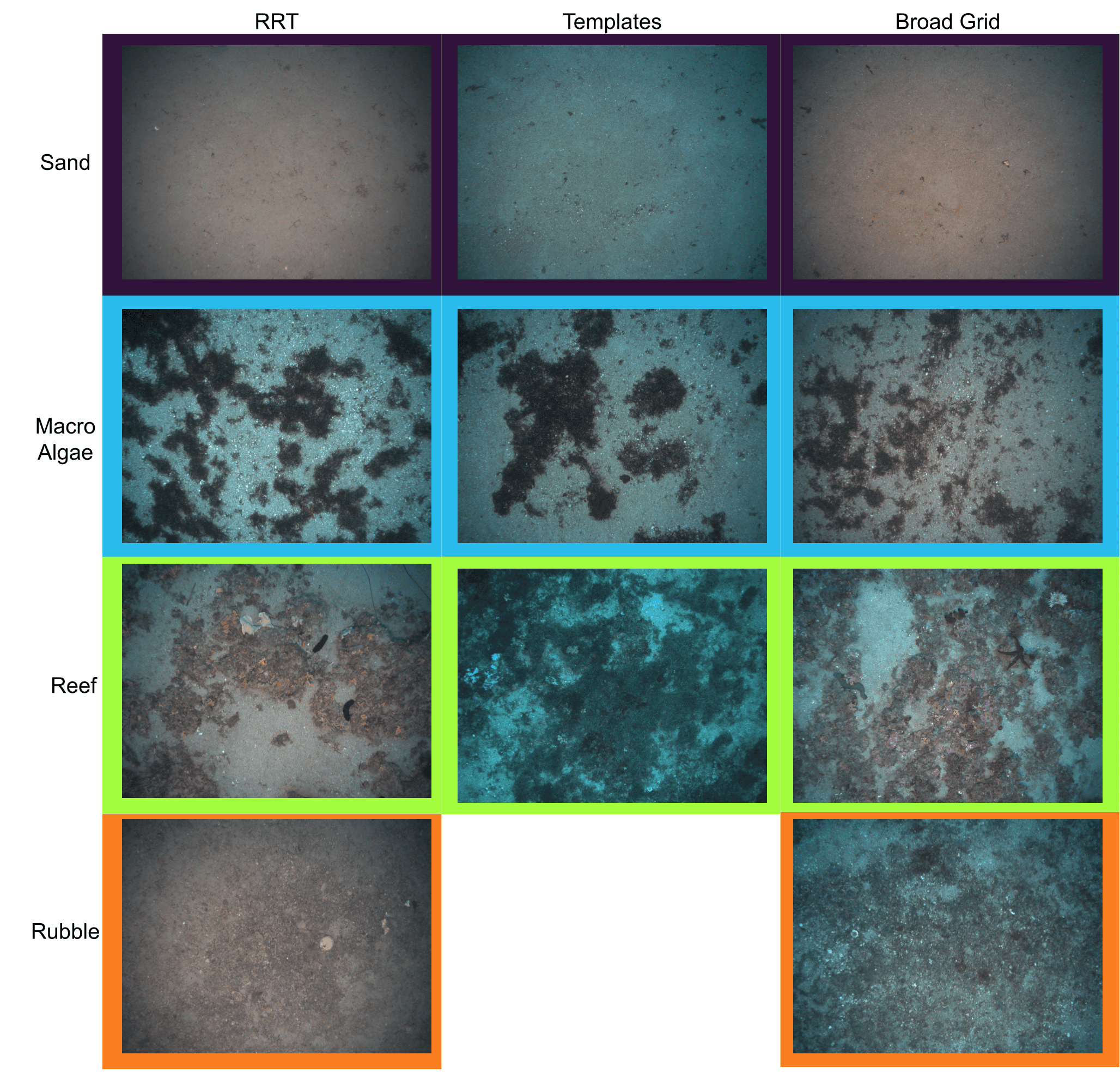}
    \caption{Images captured by the Sparus II AUV during surveys of the underwater canyon near Blanes, Spain. The border colour corresponds to the habitat class of the image, which is displayed spatially in Figure \ref{fig:field:hablabels:bathy}. The \textit{RRT} and \textit{Broad Grid} paths observed all the habitat classes, while the \textit{Templates} path did not observe the Rubble class.}
    \label{fig:field:example_images}
\end{figure}

The imagery collected by the Sparus II AUV was labelled into one of four broad habitat classes; sand, macroalgae, reef and rubble. Examples of these images and the corresponding habitat labels are displayed in Figure \ref{fig:field:example_images}. The \textit{RRT} and \textit{Broad Grid} both observe all the habitat categories. This highlights that the shorter informative path is able to explore the feature space with a limited budget. The \textit{Template} method does not observe the rubble habitat, due to either a poor fit between the template and the area, or the rubble habitat not being identifiable from the bathymetry. When considering the entire \textit{Broad Grid} path, it visits more reef areas on the first half, however the second half (2896m) of the survey does not visit any reef areas, highlighting the risk when not using informative planning.

The starting position of \textit{RRT} was selected using the method for finding an informative starting region (outlined in Section \ref{sec:info:rrt}), which generally picks successful starting positions on the border between multiple areas of bathymetric terrain. The starting position selected for the \textit{RRT} method is in an informative area, however it is deeper than a gathering of clusters just beyond it. This gathering of clusters is where the \textit{Broad Grid} observed reef habitats. Before the constraint was added, the \textit{RRT} path starting from this same starting position would visit the gathering of clusters before descending on a similar trajectory to the current \textit{RRT} path. However, when the decline constraint was added, the \textit{RRT} path could not visit this area and was forced to move deeper. Although the \textit{RRT} path still observed the reef habitats, a more efficient path could have been possible with a different starting position. This highlights the impact of the starting position and the need to adapt the process to integrate the constraints.

The field trials demonstrate how a shorter informative path can be used to explore the feature space of an area.  They highlight the adaptability of these methods to real-world constraints, however a tighter integration of the constraints into the process finding the starting position can further improve these surveys.

\section{Conclusions and Future Work}\label{sec:conclusion}

The large spatial extents of survey areas combined with the small sensor footprint necessitate efficient sampling of the survey area to maximise the information collected with limited resources. Bathymetry of an area can be collected efficiently and has a tangible connection to the benthic habitats present and can therefore be used as a prior to plan initial surveys. We present a principled approach that links feature based active learning to information gathering planning, to plan initial benthic surveys that comprehensively explore the feature space. By developing a reward function that prioritises samples that are distinct from those already on the path, the informativeness of the entire trajectory is maximised. 

A set of informative planners that utilise the reward function is proposed. These planners either plan freeform surveys (\textit{MCTS}, \textit{RRT}) or place survey templates (\textit{Templates}). All these planners exhibit significant performance improvements over random transects, highlighting the benefit of an informed survey. Furthermore, these paths more thoroughly explore the feature space than the \textit{Cluster-TSP} method, where each of the bathymetric cluster types are visited. When planning freeform surveys, the \textit{RRT} method performs marginally better on most datasets than the \textit{MCTS} method. When the survey template being evaluated is matched well to the informative areas of the survey area, informatively placing survey templates is an effective method for planning.

We demonstrate planning using two different feature extraction methods for the bathymetry; geometric features and encoded features. For both feature representations the informed survey paths comprehensively explore the feature space. However, as the number of dimensions increases, sparsity and distance concentration can reduce interpretability of the results. The relatively large distances between seemingly similar terrain can lead the planners to over-sample in some areas. This can be seen in some of the surveys based on large feature spaces, with the planners preferring to focus on rugose reef sections rather than visiting other areas of the survey area. Therefore the dimensionality of the feature space should be carefully considered as there is a tradeoff between expressibility of the feature space and the ability to plan over it.

The field trials conducted showed a shorter informative survey could still explore the feature space, but indicated further work is necessary to improve the planners to satisfy real-world constraints and objectives. Further development could also focus on making the planner more risk-averse, including ensuring the trajectory is safe when there are localisation errors.

Future work will focus on ongoing surveys of the same area, where a habitat model will be used to guide the AUV to areas that will most improve the model. Another avenue of further research will look at adapting this survey planning method to different vehicle modalities, such as drifting AUVs.

\section{Acknowledgements}

We would like to acknowledge and thank Miquel Massot Campos and Blair Thornton from the University of Southampton and Guillem Vallicrosa from the Universitat de Girona for conducting the fieldwork component of this research.

\bibliographystyle{apalike}
\bibliography{references_JFR}
\end{document}